\documentclass[10pt,a4paper,logo]{paper}

\usepackage[authoryear,round]{natbib}
\usepackage{mathtools}
\usepackage{tikz}
\usepackage{centernot}
\usepackage{algorithm}
\usepackage{algpseudocode}
\usepackage{etoolbox}
\AtBeginEnvironment{algorithmic}{\small}
\usepackage{wrapfig}
\usepackage{multirow}
\usepackage{threeparttable}
\usepackage{listings}
\usepackage{tcolorbox}
\definecolor{title_blue}{HTML}{204899}
\definecolor{cite_blue}{HTML}{044DC1}
\definecolor{cite_purple}{HTML}{7406A7}
\hypersetup{
    colorlinks=true,
    citecolor=cite_blue,
    linkcolor=cite_purple,
    urlcolor=cite_purple,
    pdftitle={Proper Scoring Rule-based Diffusion for Probabilistic Weather Forecasting},
    pdfauthor={Joonhyeong Park, Giung Nam, Hyungi Lee, Kyunghyun Cho, Byoungwoo Park, Juho Lee}
}

\usepackage{cleveref}
\usepackage{xcolor,soul}
\usepackage{colortbl}

\def\[#1\]{\begin{align}#1\end{align}}

\theoremstyle{plain}

\def\[#1\]{\begin{align}#1\end{align}}

\definecolor{myellow}{RGB}{194, 125, 47}
\definecolor{mgreen}{RGB}{48, 160, 111}

\definecolor{mteal}{RGB}{221, 254, 242}
\definecolor{bgteal}{RGB}{236, 245, 245}
\definecolor{mpurple}{RGB}{120, 111, 177}
\definecolor{citationcolor}{RGB}{80, 90, 180}

\definecolor{mygray}{gray}{0.95}
\newcommand{\graybox}[1]{%
\par\noindent
\begingroup
\setlength{\fboxsep}{0pt}%
\colorbox{mygray} {
\begin{minipage}{\linewidth}
\vspace{-0.5em}%
{#1}%
\end{minipage}%
}
\endgroup
}

\definecolor{mmgreen}{RGB}{243, 247, 243}
\definecolor{mmpurple}{RGB}{246, 242, 247}
\definecolor{mmccolor}{RGB}{235, 242, 250}

\definecolor{citationcolor}{RGB}{80, 90, 180}
\definecolor{background}{RGB}{240, 240, 250}
\definecolor{bggreen}{RGB}{229,242,229}
\newcommand{\cellbg}{\cellcolor{background}}

\usepackage{amsmath,amsfonts,bm}

\def\eqref#1{(\ref{#1})}
\def\Eqref#1{Equation~\ref{#1}}

\def\1{\bm{1}}

\DeclareMathAlphabet{\mathsfit}{\encodingdefault}{\sfdefault}{m}{sl}
\SetMathAlphabet{\mathsfit}{bold}{\encodingdefault}{\sfdefault}{bx}{n}

\definecolor{codemodel}{HTML}{1263FA}
\definecolor{codefunc}{HTML}{1263FA}
\definecolor{codezeta}{HTML}{C58A24}
\definecolor{codeloss}{HTML}{F23E43}
\definecolor{ddmadd}{HTML}{7C3FB2}
\newcommand{\ddmhl}[1]{\textcolor{ddmadd}{\textbf{#1}}}
\definecolor{codecomment}{HTML}{39868A}
\definecolor{codebg}{HTML}{FFFCF8}

\lstdefinestyle{ddmpython}{
    language=Python,
    backgroundcolor=\color{codebg},    
    basicstyle=\fontfamily{pcr}\fontsize{7}{8.2}\selectfont,
    keywordstyle=\color{black}\bfseries,
    commentstyle=\color{codecomment},
    stringstyle=\color{black},
    emph={[1]range,sample,append,update},
    emphstyle={[1]\color{codefunc}},
    emph={[2]model},
    emphstyle={[2]\color{codemodel}\mdseries},
    emph={[3]loss,fair_crps},
    emphstyle={[3]\color{codeloss}},
    emph={[4]zeta,p_zeta},
    emphstyle={[4]\color{codezeta}},
    columns=fullflexible,
    keepspaces=true,
    showstringspaces=false,
    numbers=none,
    frame=none,
    xleftmargin=0pt,
    aboveskip=3pt,
    belowskip=3pt,
    escapeinside={(*@}{@*)}
}

\definecolor{crpslossbg}{HTML}{EEF4FF}
\definecolor{ddmlossbg}{HTML}{FFF1EB}
\definecolor{ddmaccent}{HTML}{D35400}
\definecolor{crpsblue}{HTML}{1263FA}
\definecolor{ddmorange}{HTML}{FF672C}

\definecolor{step12h}{HTML}{C44E52} 
\definecolor{step1d}{HTML}{E69F00}  
\definecolor{step3d}{HTML}{228833}  
\definecolor{step5d}{HTML}{00A6A6}  

\title{Proper Scoring Rule-based Diffusion\\for Probabilistic Weather Forecasting}
\reportnumber{}

\author[1]{Joonhyeong Park}
\author[1]{Giung Nam}
\author[2]{Hyungi Lee}
\author[3]{Kyunghyun Cho}
\author[1,$\dagger$]{Byoungwoo Park}
\author[1,$\dagger$]{Juho Lee}
\affil[1]{KAIST}
\affil[2]{Kookmin University}
\affil[3]{New York University}
\renewcommand{\authornotes}{\footnotesize\textsuperscript{$\dagger$}Equal advising}
\correspondingauthor{\email{clearclouds@kaist.ac.kr}, \email{bw.park@kaist.ac.kr}, \email{juholee@kaist.ac.kr}}

\begin{abstract}
Recent probabilistic weather forecasters train stochastic predictors with the continuous ranked probability score (CRPS) to generate each ensemble member in a single forward pass. These models learn the predictive distribution from the forecast context alone, which becomes difficult at longer forecast horizons where uncertainty is high. To learn the predictive distribution more effectively, we introduce auxiliary conditional denoising tasks that predict the same future state from the context and its corrupted version, which provides partial future information that can reduce prediction ambiguity. Building on distributional diffusion models, we learn the conditional distributions of these tasks with a single stochastic predictor by minimizing a proper scoring rule across noise levels. At inference, the predictor can still generate each ensemble member in a single forward pass at the fully corrupted endpoint. Standard CRPS training is recovered as the endpoint-only special case of our formulation, so our framework extends existing CRPS-based forecasters with only additional conditioning inputs. Controlled experiments show that the auxiliary tasks improve one-step forecasting across architectures, with larger gains at longer forecast horizons. The gains extend to high-dimensional global weather forecasting under both training from scratch and fine-tuning, along with improved calibration and potential benefits for generalization under distribution shift.
\end{abstract}

\begin{document}
\maketitle

\section{Introduction}
\label{intro}

Numerical weather prediction (NWP) has long underpinned weather forecasting by simulating atmospheric evolution. Ensembles of these simulations represent uncertainty from imperfect initial conditions and model approximations~\citep{leutbecher2008ensemble,slingo2011uncertainty}, but are computationally demanding at high resolution~\citep{bauer2015quiet}. Recently, deterministic machine learning (ML) models have achieved competitive forecast skill at substantially lower inference cost than NWP~\citep{bi2023accurate,lam2023learning}, with ensemble forecasts generated from perturbed initial conditions~\citep{kurth2023fourcastnet,chen2023fuxi}. However, deterministic models trained to predict the conditional mean can produce overly smooth ensemble members~\citep{price2025probabilistic}. This limitation has motivated growing interest in probabilistic ML models for weather forecasting.

In probabilistic weather forecasting, models learn predictive distributions over future atmospheric states conditioned on the forecast context~\citep{gneiting2014probabilistic}. Diffusion-based forecasters generate spatially detailed ensemble members from noise through iterative conditional denoising~\citep{price2025probabilistic,couairon2026archesweathergen}. Yet this iterative sampling requires repeated model evaluations for each member at every forecast step, making ensemble forecasting costly. Recent CRPS-based forecasters offer an efficient alternative, training stochastic predictors with objectives based on the continuous ranked probability score (CRPS) to generate each ensemble member in a single forward pass per forecast step~\citep{alet2025skillful,bonev2025fourcastnet,lang2026aifs}. Such one-step ensemble generation is therefore promising for probabilistic weather forecasting.

However, these forecasters learn the predictive distribution from the forecast context alone, and such context-only training can become harder as forecast uncertainty grows. When forecasting several days ahead in one step, for example, we observe that CRPS-trained models tend to overfit (\Cref{sec:observation}). To learn the predictive distribution more effectively, we draw on evidence that jointly learning related tasks can enhance performance on the target task~\citep{caruana1997multitask,maurer2016benefit} and consider auxiliary prediction tasks while retaining one-step forecasting. This leads to the following question.
\begin{figure}[!t]
    \centering
    \includegraphics[width=0.85\linewidth]{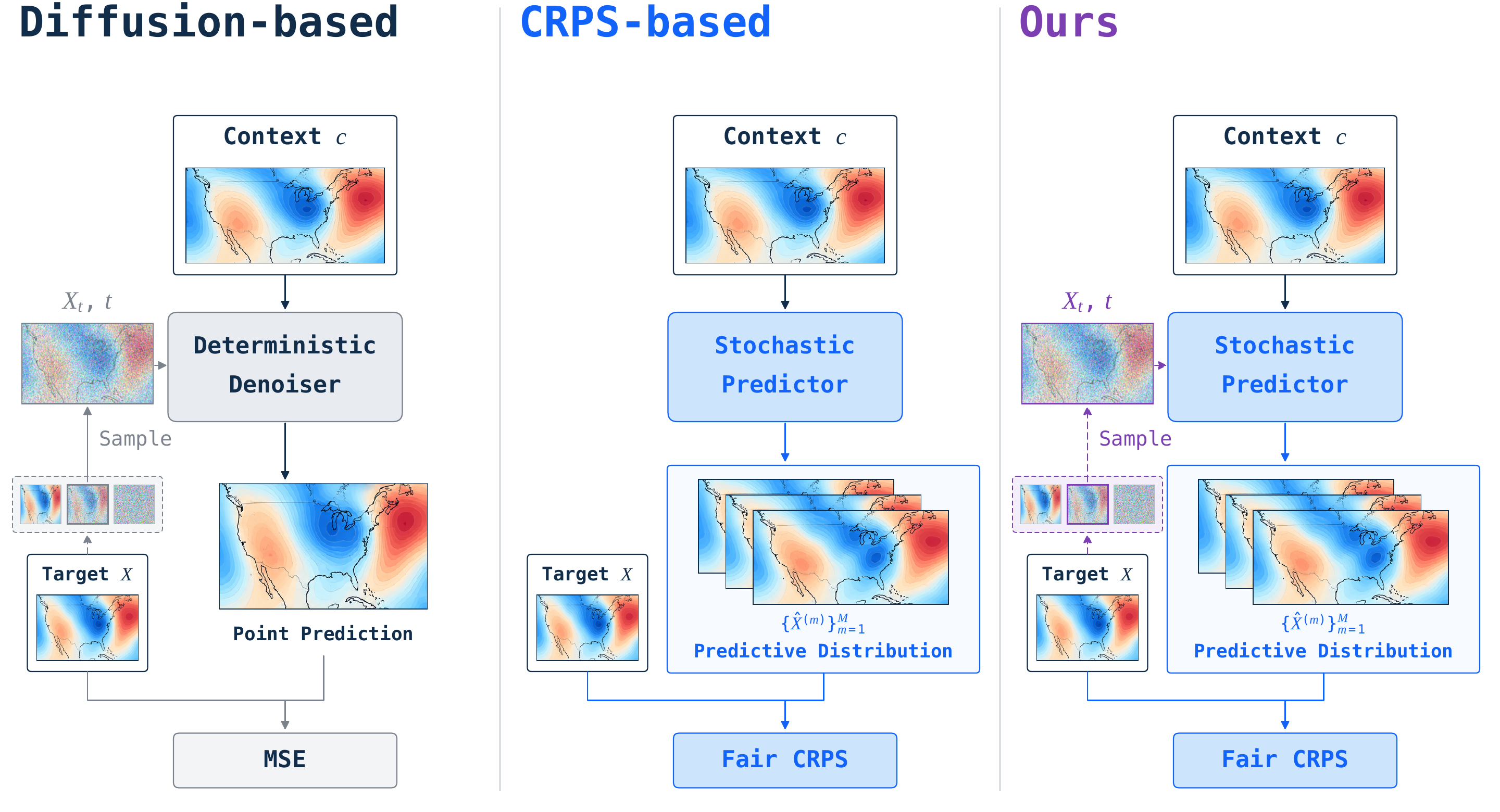}
    \caption{
    \textbf{Comparison of probabilistic weather forecasting methods}. Diffusion-based models (left) train a deterministic denoiser, while CRPS-based models (middle) train a stochastic predictor to generate ensembles. Our method (right) retains the stochastic predictor and adds a sampled corrupted future and its noise level as conditioning inputs (\textit{\ddmhl{purple}}). These inputs introduce auxiliary conditional denoising tasks learned through shared parameters, while preserving one-step ensemble generation.
    }
    \label{fig:main_conceptual_figure}
    \vspace{-1em}
\end{figure}

\graybox{%
\
\begin{center}
\textit{How can auxiliary tasks help one-step forecasters learn \\ the predictive distribution under high uncertainty?}
\end{center}
}

We address this question with auxiliary tasks that predict the same future from the context and an auxiliary variable carrying partial information about the future. We hypothesize that learning these tasks can help a single forecaster better capture the context-future relationship, as this information reduces prediction ambiguity. To include the original forecasting problem, the auxiliary variable should admit an uninformative case and, ideally, allow control over how much future information it provides. Diffusion forward process~\citep{sohl2015deep} naturally satisfies these requirements: the corrupted future becomes uninformative at the fully corrupted endpoint, and the noise level controls the amount of future information~\citep{kingma2021variational}. We therefore construct auxiliary conditional denoising tasks that predict the same clean future from its corrupted version and the forecast context.

To learn such conditional denoising tasks while retaining one-step forecasting, we introduce a proper scoring rule-based diffusion framework built on distributional diffusion models~\citep[DDMs;][]{pmlr-v267-de-bortoli25b}. In our framework, a stochastic predictor learns the conditional distributions of these tasks and generates ensemble members directly at the fully corrupted endpoint. With component-wise CRPS as the scoring rule, we show that CRPS training is an endpoint-only special case of our formulation. As illustrated in~\Cref{fig:main_conceptual_figure}, our framework can therefore extend existing CRPS-based weather forecasters by adding only the corrupted future and its noise level as conditioning inputs.

Through extensive experiments across diverse architectures and forecast horizons, we show that our formulation improves one-step forecasting over standard CRPS training, with larger gains under high uncertainty. These gains extend to high-dimensional global weather forecasting under both training from scratch and fine-tuning. We also observe improved generalization under distribution shift and in tropical cyclone cases, consistent with a possible regularizing effect of auxiliary tasks.

Our contributions are summarized below.
\vspace{-1em}
\begin{itemize}[leftmargin=25pt]
\item We introduce a proper scoring rule-based diffusion framework that uses auxiliary conditional denoising tasks to learn predictive distributions more effectively under high uncertainty.

\item We revisit standard CRPS training as the endpoint-only special case of our formulation and generalize it with the auxiliary conditional denoising tasks.

\item Building on this connection, we analyze how auxiliary conditional denoising tasks can improve one-step forecasting and support our interpretation through extensive controlled experiments.

\item We find that these benefits extend to high-dimensional weather forecasting across architectures and training regimes, with only minor conditioning changes to existing CRPS-based forecasters.
\end{itemize}

\section{Preliminaries}
\label{sec:prelim}

In this section, we briefly review probabilistic weather forecasting, focusing on scoring rule-based and diffusion-based approaches. Further background including related work is provided in~\Cref{sec:app:related_work}.

\paragraph{Probabilistic Weather Forecasting}
Let $X^{\tau}$ denote the atmospheric state at time step $\tau$, with consecutive steps separated by a fixed forecast interval. Probabilistic weather forecasting commonly adopts an autoregressive formulation, factorizing the distribution over forecast trajectories into one-step conditional distributions~\citep{price2025probabilistic,alet2025skillful},
\begin{equation}
    p\left(X^{1:T}\mid c^{0}\right)
    =
    \textstyle{\prod_{\tau=0}^{T-1}}
    p\left(X^{\tau+1}\mid c^{\tau}\right),
    \label{eq:ar_factorization}
\end{equation}
where the forecast context $c^{\tau}$ consists of the current state $X^{\tau}$ and may additionally include previous states and static and forcing variables. For notational simplicity, we write $X=X^{\tau+1}$ and $c=c^{\tau}$, so the goal is to model the predictive distribution $p(X\mid c)$~\citep{gneiting2014probabilistic}. In practice, a learned model $p_\theta(X\mid c)$ generates an ensemble of $M$ future states $\smash{\{\hat{X}^{(m)}\}_{m=1}^{M}}$ approximating this distribution, and each member is rolled out by updating the context with its predicted state.

\paragraph{Scoring Rules for Weather Forecasting}
Scoring rules assign a scalar loss to a predictive distribution and an observed outcome. Throughout this work, we adopt the convention that lower scores are better. They are \emph{proper} if expected loss is minimized when the predictive distribution matches the true distribution, and \emph{strictly proper} if the minimizer is unique~\citep{gneiting2007strictly}. 

In weather forecasting, a common choice is CRPS~\citep{matheson1976scoring}, which is given by
\begin{equation}
S_{\texttt{CRPS}}\left(p_\theta(\cdot\mid c),X\right)
=
\mathbb{E}\left[\|\hat{X}-X\|_1\right]
-
\frac{1}{2}
\mathbb{E}\left[\|\hat{X}-\hat{X}'\|_1\right],
\end{equation}
where $\hat{X},\hat{X}'\overset{\mathrm{i.i.d.}}{\sim}p_\theta(\cdot\mid c)$ and $\|\cdot\|_1$ sums absolute differences over forecast channels and grid points. With $M$ ensemble members, we estimate $S_{\texttt{CRPS}}$ with the unbiased \textit{fair CRPS}~\citep{ferro2014fair},
\begin{equation}
\hat{S}_{\texttt{CRPS}}\left(\{\hat{X}^{(m)}\}_{m=1}^{M},X\right)
=
\frac{1}{M}\sum_{m=1}^M\|\hat{X}^{(m)}-X\|_1
-
\frac{1}{2M(M-1)}\sum_{m\neq n}\|\hat{X}^{(m)}-\hat{X}^{(n)}\|_1.
\label{eq:fair_crps}
\end{equation}
Beyond evaluation, CRPS can also serve as a training objective for learning predictive distributions. Weather forecasting models can directly learn $p_\theta(\cdot\mid c)$ by minimizing a CRPS-based ensemble loss and generate each ensemble member in a single forward pass~\citep{alet2025skillful, lang2026aifs}.

\paragraph{Diffusion-based Weather Forecasting}
Diffusion-based forecasters model the predictive distribution $p(X\mid c)$ by iteratively denoising random noise $\varepsilon$ into future states conditioned on the context.

Following~\citet{lipman2023flow}, we consider the forward corruption using a linear noise schedule,
\begin{equation}
q_t(X_t\mid X)
=
\mathcal{N}\bigl(X_t\mid(1-t)X,t^2I\bigr),
\qquad
X_t=(1-t)X+t\varepsilon,
\quad
\varepsilon\sim\mathcal{N}(0,I),
\label{eq:diffusion_forward}
\end{equation}
where $t\in[0,1]$ is the noise level. Reverse sampling uses a denoiser $D_\theta(c,X_t,t)$ trained to minimize
\begin{equation}
\mathcal{L}_{\mathrm{Diffusion}}(\theta)
=
\int_0^1
w_t\,
\mathbb{E}_{\substack{
    c,X,X_t\sim q_t(\cdot\mid X)
}}
\left[
    \left\|X-D_\theta(c,X_t,t)\right\|^2
\right]
\,\mathrm{d}t,
\end{equation}
where $w_t\geq0$ weights noise levels. The optimal denoiser outputs the posterior mean, ${D^*(c,X_t,t)=\mathbb{E}[X\mid c,X_t,t]}$. Within DDIM sampling~\citep{song2021denoising}, the denoiser output serves as a point-estimate approximation of the clean-state posterior, $p(\cdot\mid c,X_t,t)\approx\delta_{D_\theta(c,X_t,t)}$, and this approximation can degrade sample quality under coarse time discretization~\citep{pmlr-v267-de-bortoli25b}. Generating high-quality ensemble members therefore typically requires fine time discretization, with repeated denoiser evaluations for each member incurring substantial inference cost~\citep{price2025probabilistic}.

\section{Proper Scoring Rule-based Diffusion Forecaster}
\label{sec:method}

We first examine challenges in context-only forecasting, which learns the predictive distribution $p(X\mid c)$ of the future state $X$ from the context $c$ alone. Motivated by these challenges, we introduce auxiliary conditional denoising tasks and evaluate their benefits. We then build on distributional diffusion models~\citep[DDMs;][]{pmlr-v267-de-bortoli25b} to formalize a proper scoring rule-based diffusion forecaster that retains one-step ensemble generation. Finally, we show that standard CRPS training can be viewed as an endpoint-only special case of our formulation and present the resulting algorithm.

\subsection{Motivating Observations: Challenges of Context-Only Forecasting}
\label{sec:observation}

\begin{wrapfigure}[11]{r}{0.39\textwidth}
    \centering

    \vspace{-2em}
    \includegraphics[width=\linewidth]{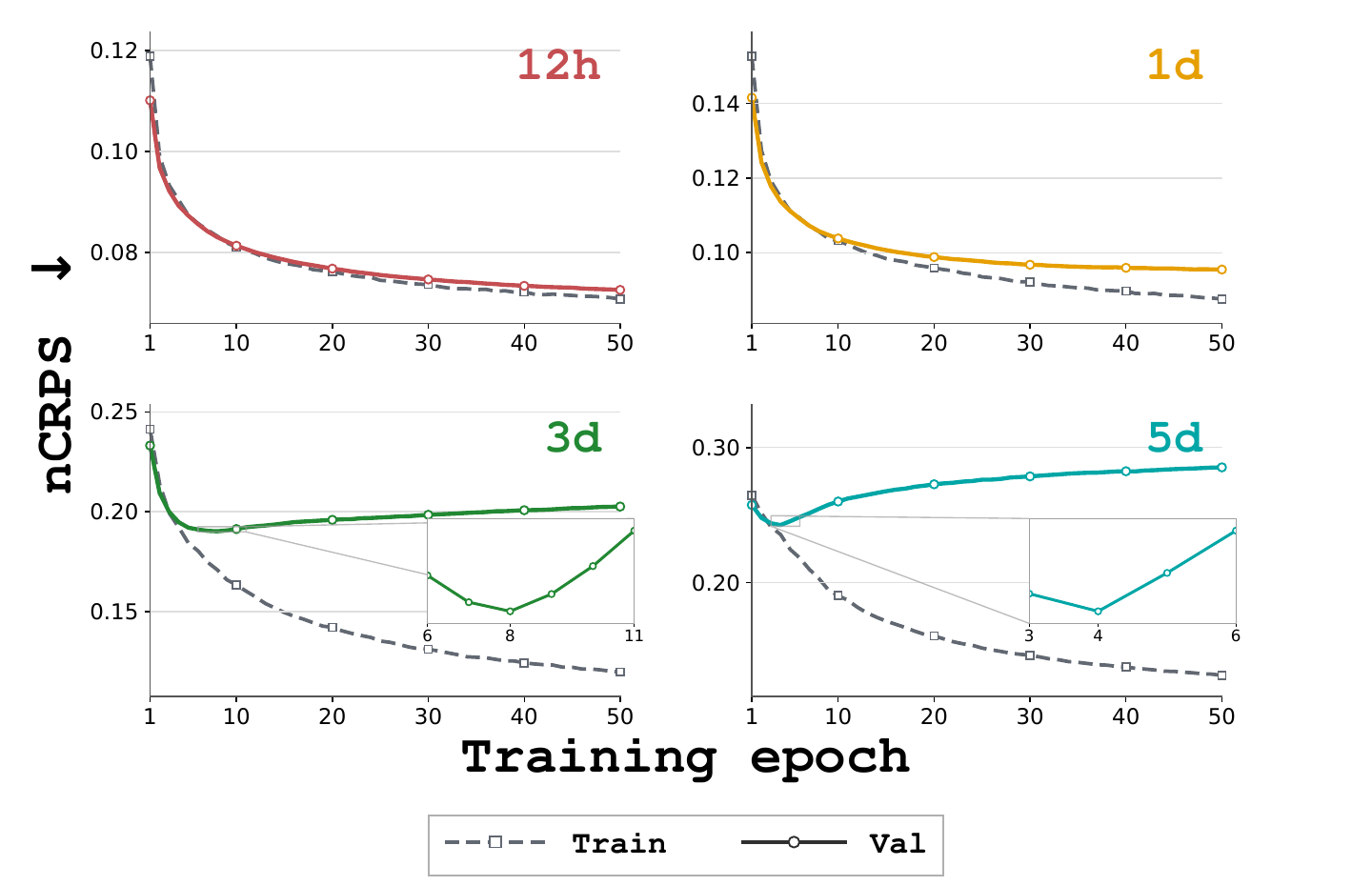}

    \vspace{-0.5em}
    \caption{FCN3~\citep{bonev2025fourcastnet} train and validation normalized CRPS (nCRPS) across forecasting horizons.}
    
    \label{fig:crps_overfitting}
\end{wrapfigure}

To examine how context-only forecasting behaves as uncertainty increases, we directly forecast the weather 12 hours and 1, 3, and 5 days ahead without autoregressive rollout. As shown in~\Cref{fig:crps_overfitting}, the validation loss of the CRPS-trained models generally improves at 12 hours and 1 day. In contrast, at 3 and 5 days, it initially improves and then deteriorates while the training loss keeps decreasing, indicating overfitting. Similar patterns appear across diverse architectures, as shown in~\Cref{fig:crps_training}. These observations suggest that context-only CRPS training can struggle at longer forecast horizons, prompting us to seek more effective ways to learn the same predictive distribution ${p(X\mid c)}$.

\paragraph{Intuition of Auxiliary Denoising Tasks}

\begin{wrapfigure}[13]{r}{0.39\textwidth}
    \centering

    \vspace{-1em}
    \includegraphics[width=\linewidth]{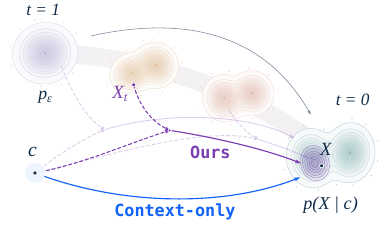}
    \caption{Schematic comparison of the context-only forecasting task and auxiliary conditional denoising tasks.}
    \label{fig:ddm_geometry}
\end{wrapfigure}

\begin{figure*}[t]
    \centering
    \includegraphics[width=1.0\textwidth]{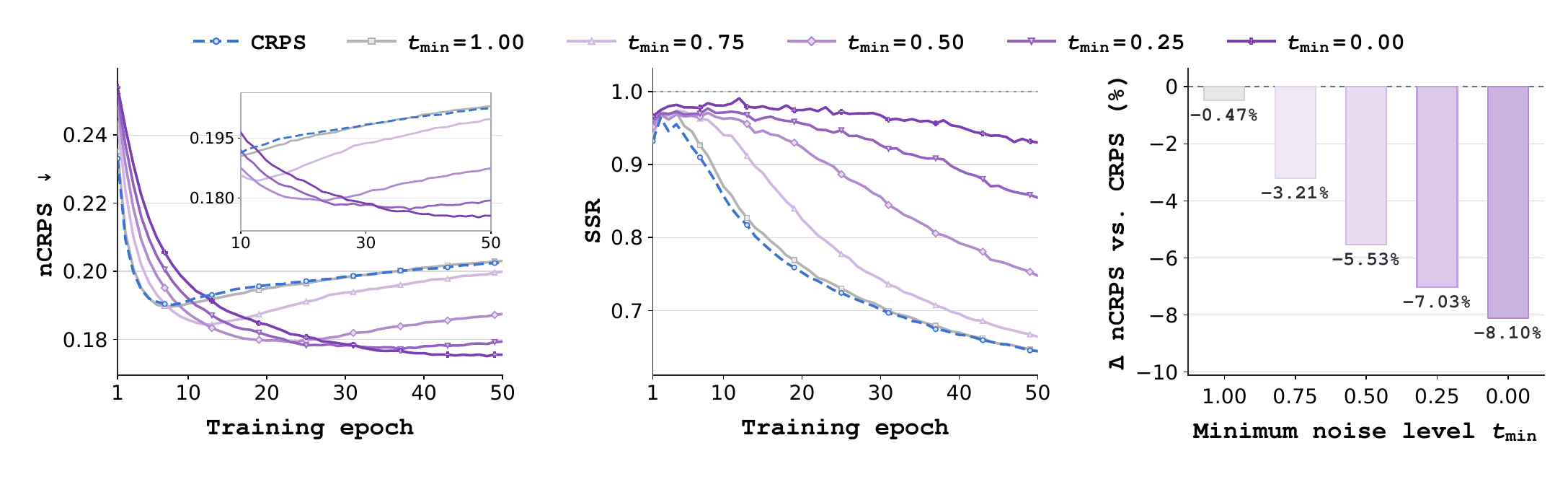}
    
    \caption{\textbf{Effect of conditional denoising tasks on one-step forecasting}. For 3-day-ahead weather forecasting, corrupted futures with noise levels $t\in[t_{\min},1]$ serve as additional training inputs. Panels show validation nCRPS (left), spread–skill ratio (SSR, middle), and test nCRPS change (\%) relative to the CRPS-trained baseline (right). Endpoint-only training ($t_{\min}=1$) closely matches the CRPS baseline. Lowering $t_{\min}$ broadens the range of auxiliary tasks, mitigating performance degradation.
    }
    
    \label{fig:tfloor_figure}
\end{figure*}

Motivated by evidence that joint training on related tasks can improve predictive performance on a target task~\citep{caruana1997multitask}, we hypothesize that jointly learning to predict the same future $X$ with access to partial information can improve context-only forecasting through shared parameters. As illustrated in~\Cref{fig:ddm_geometry}, partial future information provides clues about the future to be predicted and can reduce the ambiguity of the auxiliary prediction. Learning the same context-future relationship under these conditions may provide useful supervision for the shared model, helping address the difficulties observed in context-only forecasting. Shared training may also act as regularization and improve generalization in the original context-only forecasting task.

Under this intuition, we seek a family of auxiliary prediction tasks that includes the original forecasting problem $p(X\mid c)$ while allowing control over the amount of additional future information supplied to each task. The diffusion process in~\eqref{eq:diffusion_forward} naturally satisfies these requirements. Varying the noise level controls the information of $X_t$, and the independent-noise endpoint recovers context-only forecasting. This yields a family of conditional denoising tasks with target distributions $\{p(X\mid c,X_t,t)\}_{t\in[0,1]}$.

\paragraph{Effect of Auxiliary Denoising Tasks}

To assess the hypothesis, we jointly train a shared stochastic predictor on such conditional denoising tasks with the objective formalized in~\Cref{sec:ddm_introduction}, building on DDMs~\citep{pmlr-v267-de-bortoli25b}. We revisit the 3-day-ahead weather forecasting setting and vary the training noise range $t\in[t_{\min},1]$, while inference always uses a single forward pass at $t=1$ ($\mathrm{NFE}=1$). As shown in~\Cref{fig:tfloor_figure}, endpoint-only training ($t_{\min}=1$) closely mirrors the validation trajectories of the CRPS-trained, context-only baseline. Lowering $t_{\min}$ provides the model with richer auxiliary supervision, improving one-step forecasting and mitigating overfitting in line with our hypothesis. Additional experimental details and results are provided in~\Cref{sec:app:poc_experiment}.

\subsection{Proper Scoring Rule-based Diffusion for Auxiliary Denoising Tasks}
\label{sec:ddm_introduction}

We now formalize the joint learning of the conditional predictive distributions introduced in~\Cref{sec:observation}. Building on DDMs~\citep{pmlr-v267-de-bortoli25b}, we adopt a proper scoring rule-based objective for a shared stochastic predictor and present its empirical estimator based on finite ensembles.

\paragraph{Proper Scoring Rule-based Diffusion}
To learn the family of the conditional predictive distributions ${\{p(X \mid c, X_t, t)\}_{t\in[0,1]}}$ indexed by the noise level $t$, we minimize the expected scoring loss across noise levels using a shared stochastic predictor. Specifically, with the auxiliary variable ${X_t=(1-t)X+t\varepsilon}$ and $\varepsilon\sim\mathcal{N}(0,I)$ independent of $(c,X)$ in~\eqref{eq:diffusion_forward}, the training objective is
\begin{equation}
    \mathcal{L}_{\mathrm{DDM}}^{\lambda,\beta}(\theta)
    =
    \int_0^1
    w_t\,
    \mathbb{E}_{c,X,
        X_t\sim q_t(\cdot\mid X)
    }
    \left[
        S_{\lambda,\beta}
        \left(
            p_\theta(\cdot\mid c,X_t,t),X
        \right)
    \right]
    \,\mathrm{d}t,
    \label{eq:ddm_objective}
\end{equation}
where $p_\theta(\cdot\mid c,X_t,t)$ is the trainable predictive distribution. The nonnegative weight $w_t$ controls the contribution of each noise level, and we use uniform weighting ($w_t\equiv 1$) throughout this work. 

Here, the scoring rule $S_{\lambda,\beta}$ is a \textit{generalized energy score}~\citep{pmlr-v267-de-bortoli25b} that provides a sample-based criterion for learning the conditional predictive distribution from observed targets,
\begin{equation}
    S_{\lambda,\beta}(P,X)
    =
    \mathbb{E}_{\hat X\sim P}
    \left[\|\hat X-X\|^\beta\right]
    -
    \frac{\lambda}{2}
    \mathbb{E}_{(\hat X,\hat X')\sim P\otimes P}
    \left[\|\hat X-\hat X'\|^\beta\right],
    \label{eq:generalized_energy_score}
\end{equation}
where $\lambda\in[0,1]$ and $\beta\in(0,2]$ control dispersion and distance sensitivity, respectively. With a proper scoring rule ($\lambda=1$), a predictor satisfying $p_\theta(\cdot\mid c,X_t,t)=p(\cdot\mid c,X_t,t)$ for all $t$ attains a population minimum of~\eqref{eq:ddm_objective}. At this solution, $p_\theta(\cdot\mid c,X_1,1)=p(\cdot\mid c)$, so each ensemble member can be sampled from the target predictive distribution in a single forward pass, and multi-step sampling also recovers the same target distribution. In practice, learning an accurate one-step predictor can be challenging under high uncertainty, and multi-step sampling may improve forecasting in such cases.

\paragraph{Empirical Computation} We estimate this objective using $N$ training pairs $(c_n,X_n)$ and $M\geq 2$ predicted samples $\{\hat{X}_n^{(m)}\}_{m=1}^M$ for each corrupted future state $X_{t_n}$ with $t_n\sim\mathcal{U}(0,1)$ as
\begin{equation}
    \widehat{\mathcal{L}}_{\mathrm{DDM}}^{\lambda,\beta}(\theta)
    =
    \frac{1}{N}
    \sum_{n=1}^{N}
    \left[
        \frac{1}{M}
        \sum_{m=1}^{M}
        \left\|
            \hat X_n^{(m)}-X_n
        \right\|^\beta
        -
        \frac{\lambda}{2M(M-1)}
        \sum_{m\neq m'}
        \left\|
            \hat X_n^{(m)}-\hat X_n^{(m')}
        \right\|^\beta
    \right],
    \label{eq:empirical_ddm_objective}
\end{equation}
\begin{equation}
    \zeta_n^{(m)}\overset{\mathrm{i.i.d.}}{\sim}p_\zeta
    \;\xrightarrow{\text{introducing stochasticity}}\;
    \hat X_\theta(c_n,X_{t_n},t_n,\zeta_n^{(m)})
    =
    \hat X_n^{(m)}
    \sim p_\theta(\cdot\mid c_n,X_{t_n},t_n).
    \label{eq:predictive_samples}
\end{equation}
Here, an additional stochastic source $\zeta$, distinct from the corruption noise $\varepsilon$, allows the predictor to represent the predictive distribution by generating multiple clean futures under the same conditioning.

\subsection{Revisiting CRPS Training from a Diffusion Perspective}
\label{sec:method:crps_is_ddm}

We first establish standard context-only CRPS training as an endpoint-only special case of our formulation. We then present the resulting training algorithm, which extends standard CRPS training by incorporating auxiliary conditional denoising tasks while retaining one-step ensemble generation.

\paragraph{Generalization of CRPS Training}

From the distributional diffusion view in~\Cref{sec:ddm_introduction}, standard CRPS training can be understood as an endpoint-only special case. If we set $\lambda=\beta=1$ and apply the score component-wise, we obtain $S_{1,1}=S_{\texttt{CRPS}}$. With training restricted to $t=1$ and the endpoint predictive distribution independent of corruption noise, \Eqref{eq:ddm_objective} reduces to the CRPS objective\footnote{See~\Cref{sec:app:crps_endpoint} for the mathematical details.}:

\graybox{
\vspace{0.5em}
\begin{equation}
\begin{aligned}
    \mathcal{L}_{\mathrm{DDM}}^{1,1}(\theta)
    &=
    \int_0^1
    \mathbb{E}_{c,X,\,X_t\sim q_t(\cdot\mid X)}
    \left[
        S_{\texttt{CRPS}}
        \left(p_\theta(\cdot\mid c,X_t,t),X\right)
    \right]
    \,\mathrm{d}t
    \\[0.4em]
    &\xrightarrow{\text{restrict training to }t=1}
    \mathbb{E}_{c,X}
    \left[
        S_{\texttt{CRPS}}
        \left(p_\theta(\cdot\mid c),X\right)
    \right].\vphantom{\int}
\end{aligned}
\label{eq:endpoint_crps}
\end{equation}

}

From this view, standard CRPS training can be interpreted as learning only the endpoint of the task family in~\Cref{sec:observation}, and our formulation as generalizing it with auxiliary denoising tasks at $t<1$.
\paragraph{Training Algorithm }
\begin{wrapfigure}[14]{r}{0.40\textwidth}
    \vspace{-1em}
    \refstepcounter{algorithm}
    \label{alg:ddm_training}

    \hrule height 0.8pt
    \vspace{2pt}
    {\small\textbf{Algorithm \thealgorithm}\ Training\par}
    \vspace{2pt}
    \hrule height 0.4pt

\begin{lstlisting}[
    style=ddmpython,
    basicstyle=\fontfamily{pcr}\fontsize{7}{8.2}\selectfont
]
# model - stochastic predictor
# M - ensemble size (M >= 2)
for c, x in dataloader:
  # Corrupt the future
  (*@\ddmhl{t = uniform(0, 1)}@*)
  (*@\ddmhl{eps = randn\_like(x)}@*)
  (*@\ddmhl{x\_t = (1-t)*x + t*eps}@*)

  # Stochastic Ensemble Generation
  x_hat = []
  for m in range(M):
    zeta = sample(p_zeta)
    pred = model(c,zeta,(*@\ddmhl{x\_t=x\_t,t=t}@*))
    x_hat.append(pred)

  # CRPS update
  loss = fair_crps(x_hat, x)
  update(model, loss)
\end{lstlisting}
    \hrule height 0.4pt
\end{wrapfigure}
In this work, we set ${\lambda=\beta=1}$ and apply the score component-wise, yielding a CRPS-based DDM loss that naturally extends standard CRPS training with auxiliary denoising tasks across noise levels.

The training procedure is summarized in~\Cref{alg:ddm_training}, with \textit{the only additional operations} relative to standard CRPS training highlighted in \textit{\ddmhl{purple}}. These operations sample a noise level $t$, construct a corrupted future $X_t$, and provide both as additional conditioning inputs. For each context--future pair $(c, X)$, we generate $M$ ensemble members $\{\hat{X}^{(m)}\}_{m=1}^{M}$ conditioned on the same corrupted future $X_t$, using independent draws of $\zeta^{(m)}$ to introduce stochasticity. At each sampled noise level, the predictor is then updated using a loss with the fair CRPS form.

The resulting training objective averages this fair CRPS over $N$ context--future pairs and noise levels:
\begin{equation}
    \hat{L}(\theta)
    =
    \frac{1}{N}
    \sum_{n=1}^{N}
    \left[
        \frac{1}{M}
        \sum_{m=1}^{M}
        \left\|
            \hat{X}_n^{(m)}-X_n
        \right\|_1
        -
        \frac{1}{2M(M-1)}
        \sum_{m\neq m'}
        \left\|
            \hat{X}_n^{(m)}-\hat{X}_n^{(m')}
        \right\|_1
    \right],
    \label{eq:empirical_crps_ddm_objective}
\end{equation}
where $\hat{X}_n^{(m)}=\hat{X}_\theta(c_n,X_{t_n},t_n,\zeta_n^{(m)})$ and $\|\cdot\|_1$ sums absolute differences across forecast channels and spatial locations. We apply area and channel weighting as detailed in Appendix~\ref{sec:app:exp}, but omit them here for simplicity. One-step and multi-step sampling procedures are provided in Appendix~\ref{sec:app:sampling}.

\section{Experiments}
\label{sec:experiments}

We evaluate our formulation against standard CRPS training in two settings. In the controlled pilot setting, we vary the forecast horizon to assess the contribution of auxiliary conditional denoising tasks under different levels of uncertainty. In the high-dimensional global setting, we test whether the gains transfer across training regimes, including training from scratch and fine-tuning, and examine how the auxiliary tasks affect generalization under distribution shift and in tropical cyclone cases. Both settings include multiple architectures, allowing us to assess the consistency of the benefits.

\paragraph{Experimental Setup} We use ERA5 reanalysis with \texttt{1979--2017}, \texttt{2018--2019}, and \texttt{2020} for training, validation, and testing, respectively~\citep{hersbach2020era5}. The controlled experiments use a $2.8125^\circ$ ($64\times128$) global latitude--longitude grid with 9 representative channels, while the large-scale experiments use a $1.5^\circ$ ($121\times240$) grid with 84 channels. We refer to models trained with the DDM objective as DDM and compare them against standard CRPS training under matched architectures and training configurations. All results in the main text use $\mathrm{NFE}=1$ unless stated otherwise. Further details including data preprocessing and additional results are described in Appendices~\ref{sec:app:exp} and~\ref{sec:app:additional}.

\subsection{Pilot Experiments: Analysis of Auxiliary Conditional Denoising Tasks}

In this $2.8125^\circ$ setting, we consider two CRPS-based model architectures, FGN~\citep{alet2025skillful} and FCN3~\citep{bonev2025fourcastnet}, with direct forecast horizons of \textcolor{step12h}{\textbf{\texttt{12h}}}, \textcolor{step1d}{\textbf{\texttt{1d}}}, \textcolor{step3d}{\textbf{\texttt{3d}}}, and \textcolor{step5d}{\textbf{\texttt{5d}}}. Each direct forecast horizon serves as the \textit{rollout interval} (RI) during autoregressive rollouts. Detailed experimental configurations and additional results including other models are provided in Appendices~\ref{sec:app:2p8} and~\ref{sec:app:additional_2.8}.

\newcommand{\reldelta}[1]{%
    \smash{\raisebox{0.55ex}{%
        \scriptsize\textcolor{gray}{\,#1\%}%
    }}%
}

\begin{figure*}[t]
    \centering
    \begin{minipage}[t]{0.69\textwidth}
        \centering
        \captionof{table}{Direct Forecasting Performance}
        
        \resizebox{\linewidth}{!}{%
        \setlength{\tabcolsep}{0pt}
        \setlength{\aboverulesep}{0pt}
        \setlength{\belowrulesep}{0pt}
        \renewcommand{\arraystretch}{1.5}
        \begin{tabular}{>{\centering\arraybackslash}p{1.3cm}>{\centering\arraybackslash}p{2.2cm}*{8}{>{\centering\arraybackslash}p{1.2cm}}}
            \toprule
        
            & &
            \multicolumn{4}{c}{\textbf{\texttt{nCRPS}} $\downarrow$}
            & \multicolumn{4}{c}{\textbf{\texttt{nRMSE}} $\downarrow$}
            \\
        
            \cmidrule(lr){3-6}
            \cmidrule(lr){7-10}
        
            Model & Objective
            & \textcolor{step12h}{\textbf{\texttt{12h}}}
            & \textcolor{step1d}{\textbf{\texttt{1d}}}
            & \textcolor{step3d}{\textbf{\texttt{3d}}}
            & \textcolor{step5d}{\textbf{\texttt{5d}}}
            & \textcolor{step12h}{\textbf{\texttt{12h}}}
            & \textcolor{step1d}{\textbf{\texttt{1d}}}
            & \textcolor{step3d}{\textbf{\texttt{3d}}}
            & \textcolor{step5d}{\textbf{\texttt{5d}}}
            \\
        
            \midrule
        
            \multirow{3}{*}{\textbf{\texttt{FGN}}}
            & \textbf{\texttt{CRPS}}
            & 0.0581 & 0.0792 & 0.1868 & 0.2463
            & 0.1138 & 0.1586 & 0.3847 & 0.5034
            \\
        
            & \textbf{\texttt{DDM}}
            & \cellbg 0.0577
            & \cellbg 0.0778
            & \cellbg 0.1692
            & \cellbg 0.2290
            & \cellbg 0.1130
            & \cellbg 0.1559
            & \cellbg 0.3542
            & \cellbg 0.4760
            \\
        
            \cmidrule(l){2-10}
            & \textbf{\texttt{$\Delta$(\%)}}
            & -0.8 & -1.8 & -9.4 & -7.0
            & -0.7 & -1.7 & -7.9 & -5.4
            \\
        
            \midrule
        
            \multirow{3}{*}{\textbf{\texttt{FCN3}}}
            & \textbf{\texttt{CRPS}}
            & 0.0736 & 0.0967 & 0.1915 & 0.2452
            & 0.1461 & 0.1937 & 0.3920 & 0.5029
            \\
        
            & \textbf{\texttt{DDM}}
            & \cellbg 0.0735
            & \cellbg 0.0942
            & \cellbg 0.1759
            & \cellbg 0.2283
            & \cellbg 0.1461
            & \cellbg 0.1894
            & \cellbg 0.3664
            & \cellbg 0.4755
            \\
        
            \cmidrule(l){2-10}
            & \textbf{\texttt{$\Delta$(\%)}}
            & -0.1 & -2.6 & -8.1 & -6.9
            & -0.0 & -2.2 & -6.6 & -5.5
            \\
        
            \bottomrule
        \end{tabular}%
        }
    \label{table:pilot_direct}
    \end{minipage}
    \hfill
    \begin{minipage}[t]{0.29\textwidth}
        \centering
        \captionof{figure}{Effect of NFE}
        
        \includegraphics[width=\linewidth]{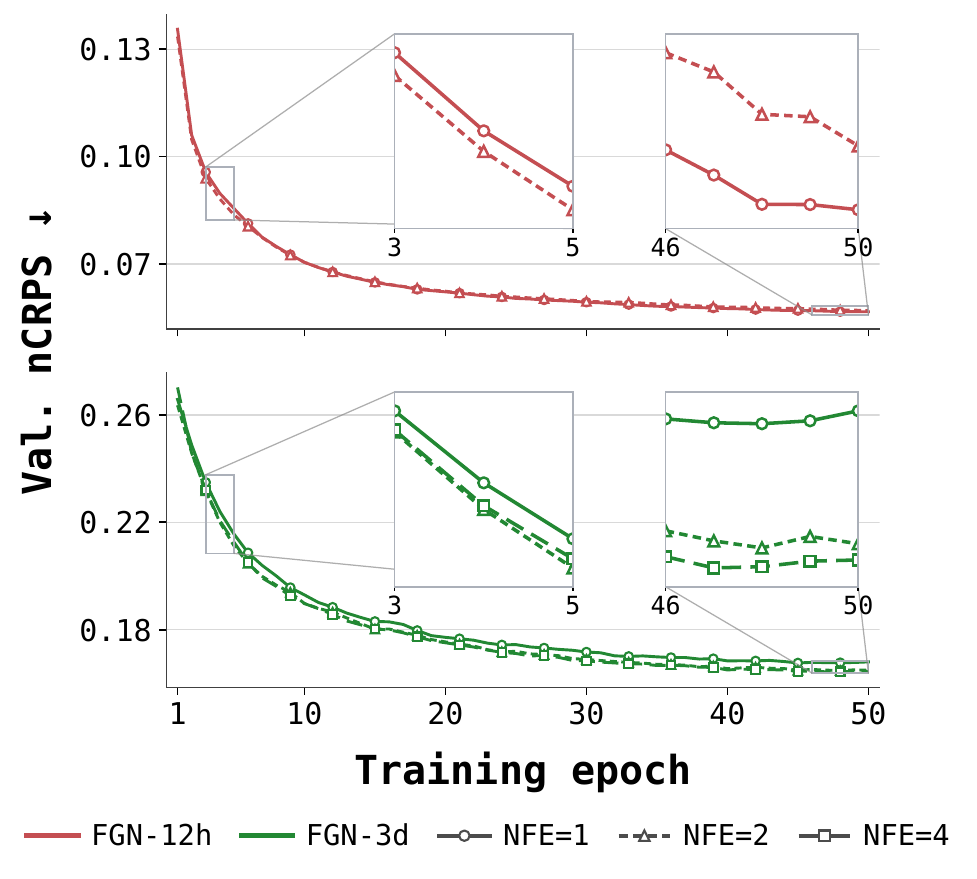}
    \label{fig:pilot_nfe}
    \end{minipage}

\end{figure*}
\newcommand{\scoregain}[2]{%
    \shortstack[c]{#1\\[-1pt]{\scriptsize $#2\%$}}%
}

\begin{figure*}[t]
    \centering
    
    \begin{minipage}[t]{0.69\textwidth}
        \vspace{0pt}
        \centering

        \captionof{table}{Autoregressive Rollout Performance}

        \resizebox{\linewidth}{!}{%
        \setlength{\tabcolsep}{0pt}
        \setlength{\aboverulesep}{0pt}
        \setlength{\belowrulesep}{0pt}
        \renewcommand{\arraystretch}{1.45}
        \begin{tabular}{>{\centering\arraybackslash}p{1.3cm}>{\centering\arraybackslash}p{1.1cm}>{\centering\arraybackslash}p{1.1cm}*{6}{>{\centering\arraybackslash}p{1.6cm}}}

            \toprule
        
            & & &
            \multicolumn{3}{c}{\textbf{\texttt{nCRPS}} $\downarrow$}
            & \multicolumn{3}{c}{\textbf{\texttt{nRMSE}} $\downarrow$}
            \\
            
            \cmidrule(lr){4-6}
            \cmidrule(lr){7-9}
        
            Model
            & RI
            & Obj.
            & {{\textbf{\texttt{12h}}}}
            & {{\textbf{\texttt{72h}}}}
            & {{\textbf{\texttt{216h}}}}
            & {{\textbf{\texttt{12h}}}}
            & {{\textbf{\texttt{72h}}}}
            & {{\textbf{\texttt{216h}}}}
            \\
        
            \midrule
        
            \multirow{6}{*}{\textbf{\texttt{FGN}}}
        
            & \multirow{3}{*}{
                \textcolor{step12h}{\textbf{\texttt{12h}}}
            }
            & \textbf{\texttt{CRPS}}
            & 0.0581
            & 0.1420
            & 0.2391
            & 0.1138
            & 0.2986
            & 0.4981
            \\
        
            &
            & \textbf{\texttt{DDM}}
            & \cellbg 0.0577
            & \cellbg 0.1419
            & \cellbg 0.2391
            & \cellbg 0.1130
            & \cellbg 0.2985
            & 0.4985
            \\
        
            \cmidrule{3-9}
            &
            & \textbf{\texttt{$\Delta$(\%)}}
            & -0.8
            & -0.1
            & -0.0
            & -0.7
            & -0.0
            & +0.1
            \\
        
            \cmidrule(l){2-9}
        
            & \multirow{3}{*}{
                \textcolor{step3d}{\textbf{\texttt{3d}}}
            }
            & \textbf{\texttt{CRPS}}
            & $-$
            & 0.1868
            & 0.2557
            & $-$
            & 0.3847
            & 0.5239
            \\
        
            &
            & \textbf{\texttt{DDM}}
            & $-$
            & \cellbg 0.1692
            & \cellbg 0.2451
            & $-$
            & \cellbg 0.3542
            & \cellbg 0.5088
            \\
        
            \cmidrule{3-9}
            &
            & \textbf{\texttt{$\Delta$(\%)}}
            & $-$
            & -9.4
            & -4.1
            & $-$
            & -7.9
            & -2.9
            \\
        
            \midrule
        
            \multirow{6}{*}{\textbf{\texttt{FCN3}}}
        
            & \multirow{3}{*}{
                \textcolor{step12h}{\textbf{\texttt{12h}}}
            }
            & \textbf{\texttt{CRPS}}
            & 0.0736
            & 0.1634
            & 0.2474
            & 0.1461
            & 0.3399
            & 0.5133
            \\
        
            &
            & \textbf{\texttt{DDM}}
            & \cellbg 0.0735
            & \cellbg 0.1632
            & \cellbg 0.2468
            & \cellbg 0.1461
            & \cellbg 0.3394
            & \cellbg 0.5119
            \\
        
            \cmidrule{3-9}
            &
            & \textbf{\texttt{$\Delta$(\%)}}
            & -0.1
            & -0.1
            & -0.2
            & -0.0
            & -0.1
            & -0.3
            \\
        
            \cmidrule(l){2-9}
        
            & \multirow{3}{*}{
                \textcolor{step3d}{\textbf{\texttt{3d}}}
            }
            & \textbf{\texttt{CRPS}}
            & $-$
            & 0.1915
            & 0.2558
            & $-$
            & 0.3920
            & 0.5246
            \\
        
            &
            & \textbf{\texttt{DDM}}
            & $-$
            & \cellbg 0.1759
            & \cellbg 0.2461
            & $-$
            & \cellbg 0.3664
            & \cellbg 0.5118
            \\
        
            \cmidrule{3-9}
            &
            & \textbf{\texttt{$\Delta$(\%)}}
            & $-$
            & -8.1
            & -3.8
            & $-$
            & -6.6
            & -2.4
            \\
        
            \bottomrule
        \end{tabular}      
        }
    \label{table:pilot_ar}
    \end{minipage}
    \hfill
    \begin{minipage}[t]{0.29\textwidth}
        \centering

        \captionof{figure}{SSR Analysis}
        
        \includegraphics[
            width=\linewidth
        ]{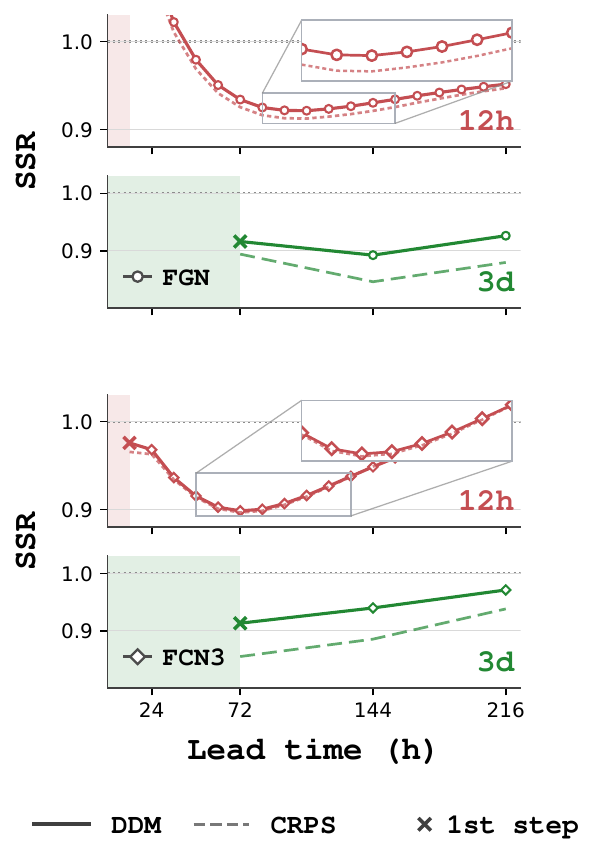}
    \label{fig:pilot_ssr}
    \end{minipage}

\end{figure*}

\paragraph{Direct Forecasting Performance}

We first assess the contribution of auxiliary conditional denoising tasks to direct forecasting at $\text{NFE}=1$ under different levels of uncertainty.  As shown in~\Cref{table:pilot_direct}, training with the DDM objective consistently yields better performance than standard CRPS training. 

Specifically, the gains are larger in the \textcolor{step3d}{\texttt{\textbf{3d}}} and \textcolor{step5d}{\texttt{\textbf{5d}}} tasks, where predicting future states directly from context is more challenging, than for shorter forecast horizons. Under high uncertainty, richer supervision from auxiliary conditional denoising tasks may help the shared model better approximate the target predictive distribution. The noise-range ablation in~\Cref{fig:tfloor_figure} also supports this interpretation, showing that broadening the range of auxiliary tasks improves $\text{NFE}=1$ forecasting in the \textcolor{step3d}{\texttt{\textbf{3d}}} task.

In contrast, gains are smaller in \textcolor{step12h}{\texttt{\textbf{12h}}} and \textcolor{step1d}{\texttt{\textbf{1d}}}. As noted earlier, both formulations admit population-optimal solutions that recover the true predictive distribution $p(X\mid c)$. In these easier tasks, CRPS training may already approximate the forecast marginals well, limiting potential gains from auxiliary conditional denoising. This provides one possible explanation for their similar performance.

\paragraph{Effect of Additional NFEs} The benefits of additional NFEs can also be interpreted from the optimality perspective. At an ideal population optimum where the joint conditional denoising distributions are recovered exactly at all noise levels, DDM sampling recovers the true predictive distribution $p(X\mid c)$ regardless of NFE (see~\Cref{sec:app:sampling}). When one-step predictions already approximate this distribution well, additional NFEs may offer limited gains. Consistent with this expectation, \Cref{fig:pilot_nfe} shows that NFE gains in the \textcolor{step12h}{\texttt{\textbf{12h}}} task are present only in early training. In contrast, gains persist in the \textcolor{step3d}{\texttt{\textbf{3d}}} task, where learning an accurate one-step predictor may be more difficult and iterative refinement may improve the approximation of the target predictive distribution.

\paragraph{Autoregressive Rollout Performance} We next evaluate autoregressive rollouts at $\text{NFE}=1$ per rollout step. Across both architectures, DDM generally yields modest gains with a \textcolor{step12h}{\textbf{\texttt{12h}}} rollout interval (RI) and maintains a clear performance advantage with a \textcolor{step3d}{\textbf{\texttt{3d}}} interval (\Cref{table:pilot_ar}). At longer RIs, DDM also alleviates the underdispersion observed with CRPS training, maintaining SSRs closer to $1$ and suggesting better ensemble calibration (\Cref{fig:pilot_ssr}). Together, the pilot results support our expectation that auxiliary conditional denoising tasks improve direct forecasting at $\text{NFE}=1$, with gains that persist through autoregressive rollouts and are more pronounced at longer RIs.

\subsection{Scaling to High-Dimensional Global Weather Forecasting}
\label{sec:res:realworld}

Towards realistic global weather forecasting, we scale our evaluation to $1.5^\circ$ ERA5 reanalysis with 84 forecast channels. We consider two recently proposed forecasting models, MOSAIC~\citep{zhdanov2026sparse} and U-Cast~\citep{cachay2026u}, adapting their original training recipes. The two models provide complementary settings. MOSAIC is trained probabilistically from scratch and predicts the atmospheric state directly, whereas U-Cast is probabilistically fine-tuned from a model pretrained with mean absolute error (MAE) and predicts residuals. They further differ in how stochasticity is introduced, allowing us to evaluate DDM across distinct training regimes and parameterizations. Detailed descriptions of the model designs and training configurations are provided in Appendix~\ref{sec:app:1p5}.

\begin{table}[!t]
\centering
\caption{
CRPS and ensemble-mean RMSE across representative variables and forecast lead times\protect\footnotemark.
}
\label{tab:1p5deg_analysis}
\resizebox{\linewidth}{!}{%
\setlength{\tabcolsep}{0pt}
\setlength{\aboverulesep}{0pt}
\setlength{\belowrulesep}{0pt}
\renewcommand{\arraystretch}{1.35}
\begin{tabular}{p{2.1cm}*{18}{>{\centering\arraybackslash}p{1.2cm}}}
\toprule

&
\multicolumn{9}{c}{\texttt{\textbf{CRPS}} $\downarrow$}
&
\multicolumn{9}{c}{\texttt{\textbf{RMSE}} $\downarrow$}
\\

\cmidrule(lr){2-10}
\cmidrule(lr){11-19}

&
\multicolumn{3}{c}{\textbf{\texttt{1d}}}
&
\multicolumn{3}{c}{\textbf{\texttt{3d}}}
&
\multicolumn{3}{c}{\textbf{\texttt{10d}}}
&
\multicolumn{3}{c}{\textbf{\texttt{1d}}}
&
\multicolumn{3}{c}{\textbf{\texttt{3d}}}
&
\multicolumn{3}{c}{\textbf{\texttt{10d}}}
\\

\cmidrule(lr){2-4}
\cmidrule(lr){5-7}
\cmidrule(lr){8-10}
\cmidrule(lr){11-13}
\cmidrule(lr){14-16}
\cmidrule(lr){17-19}

Method
& \textbf{\texttt{Z500}}
& \textbf{\texttt{T850}}
& \textbf{\texttt{Q700}}
& \textbf{\texttt{Z500}}
& \textbf{\texttt{T850}}
& \textbf{\texttt{Q700}}
& \textbf{\texttt{Z500}}
& \textbf{\texttt{T850}}
& \textbf{\texttt{Q700}}
& \textbf{\texttt{Z500}}
& \textbf{\texttt{T850}}
& \textbf{\texttt{Q700}}
& \textbf{\texttt{Z500}}
& \textbf{\texttt{T850}}
& \textbf{\texttt{Q700}}
& \textbf{\texttt{Z500}}
& \textbf{\texttt{T850}}
& \textbf{\texttt{Q700}}
\\

\midrule

\textbf{\texttt{IFS-ENS}$\dagger$}
& 24.1 & 0.370 & 0.283
& 59.3 & 0.550 & 0.427
& 263.8 & 1.343 & 0.725
& 45.8 & 0.718 & 0.611
& 133.4 & 1.105 & 0.902
& 623.1 & 2.819 & 1.440
\\

\midrule

\textbf{\texttt{GenCast}$\dagger$}
& 20.2 & 0.269 & 0.222
& 54.3 & 0.459 & 0.367
& 253.6 & 1.281 & 0.676
& 39.3 & 0.542 & 0.498
& 123.5 & 0.955 & 0.807
& 606.5 & 2.737 & 1.375
\\

\textbf{\texttt{NeuralGCM}$\dagger$}
& 22.9 & 0.333 & 0.263
& 54.9 & 0.497 & 0.380
& 253.4 & 1.293 & 0.679
& 44.0 & 0.658 & 0.543
& 126.2 & 1.027 & 0.812
& 606.8 & 2.756 & 1.374
\\

\midrule

\textbf{\texttt{MOSAIC}}
& 22.7 & 0.300 & 0.239
& 58.3 & 0.480 & 0.372
& 261.0 & 1.303 & 0.680
& 44.2 & 0.601 & 0.525
& 133.2 & 1.001 & 0.814
& 619.9 & 2.781 & 1.383
\\

$\textbf{\texttt{MOSAIC}}_{\texttt{DDM}}$
& \cellbg 21.9 & \cellbg 0.290 & \cellbg 0.232
& \cellbg 56.7 & \cellbg 0.467 & \cellbg 0.365
& \cellbg 257.6 & \cellbg 1.290 & \cellbg 0.675
& \cellbg 42.4 & \cellbg 0.581 & \cellbg 0.513
& \cellbg 128.6 & \cellbg 0.973 & \cellbg 0.800
& \cellbg 610.1 & \cellbg 2.755 & \cellbg 1.374
\\

\midrule

\textbf{\texttt{U-Cast}}
& 20.4 & 0.281 & 0.229
& 56.1 & 0.480 & 0.373
& 260.1 & 1.326 & 0.704
& 39.8 & 0.563 & 0.507
& 126.9 & 0.985 & 0.817
& 615.7 & 2.792 & 1.427
\\

$\textbf{\texttt{U-Cast}}_{\texttt{DDM}}$
& \cellbg 20.4 & \cellbg 0.280 & \cellbg 0.229
& \cellbg 56.1 & \cellbg 0.475 & \cellbg 0.372
& \cellbg 259.5 & \cellbg 1.314 & \cellbg 0.694
& \cellbg 39.7 & \cellbg 0.562 & \cellbg 0.507
& \cellbg 126.6 & \cellbg 0.979 & \cellbg 0.814
& \cellbg 613.2 & \cellbg 2.779 & \cellbg 1.409
\\

\bottomrule
\end{tabular}
}
\end{table}
\footnotetext{Results marked with $^\dagger$ are provided for reference, as their configurations differ from our controlled experimental setup. Our primary comparisons evaluate standard CRPS and DDM training under matched settings.}

\paragraph{Forecast Performance}

\Cref{tab:1p5deg_analysis} compares CRPS and ensemble-mean RMSE on three variables at 1-, 3-, and 10-day lead times. At $\text{NFE}=1$ per rollout step, DDM improves both metrics in all cases for MOSAIC and improves or matches them for U-Cast.

\begin{wrapfigure}[13]{r}{0.45\textwidth}
    \centering

    \vspace{-3em}
    \includegraphics[width=\linewidth]{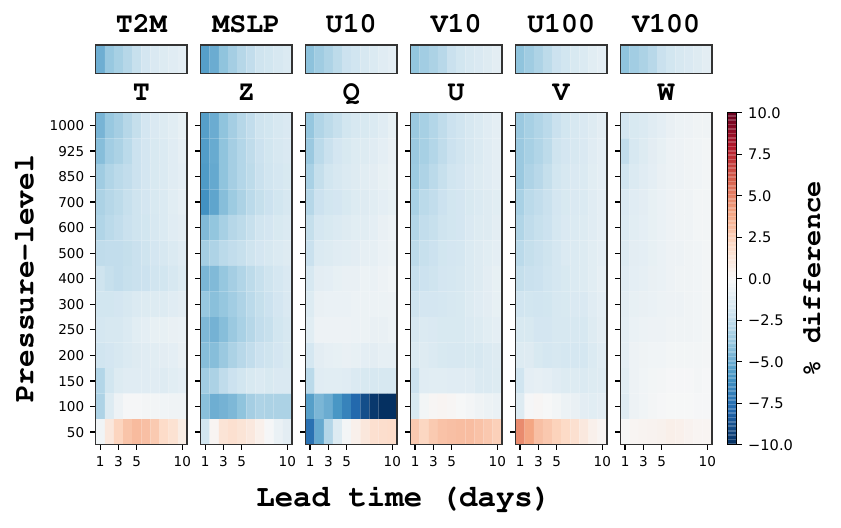}
    \caption{Relative CRPS difference between DDM- and CRPS-trained MOSAIC across all 84 forecast channels over 10-day rollouts. Blue indicates better performance of DDM.}
    \label{fig:1p5deg_heatmap}
    
\end{wrapfigure}

These improvements extend beyond the representative variables. Across all 84 forecast channels, DDM achieves lower CRPS in 787 of 840 channel--lead pairs (93.7\%) for MOSAIC, as shown in~\Cref{fig:1p5deg_heatmap}. For U-Cast, where the objectives differ only during brief fine-tuning, the gains are modest but extend across most channel--lead pairs (82.0\%, see \Cref{fig:scorecard_2x2}). Together, these results indicate that auxiliary conditional denoising tasks benefit high-dimensional forecasting in both training from scratch and fine-tuning.

\begin{figure*}[t]
    \centering
    \includegraphics[width=0.9\textwidth]{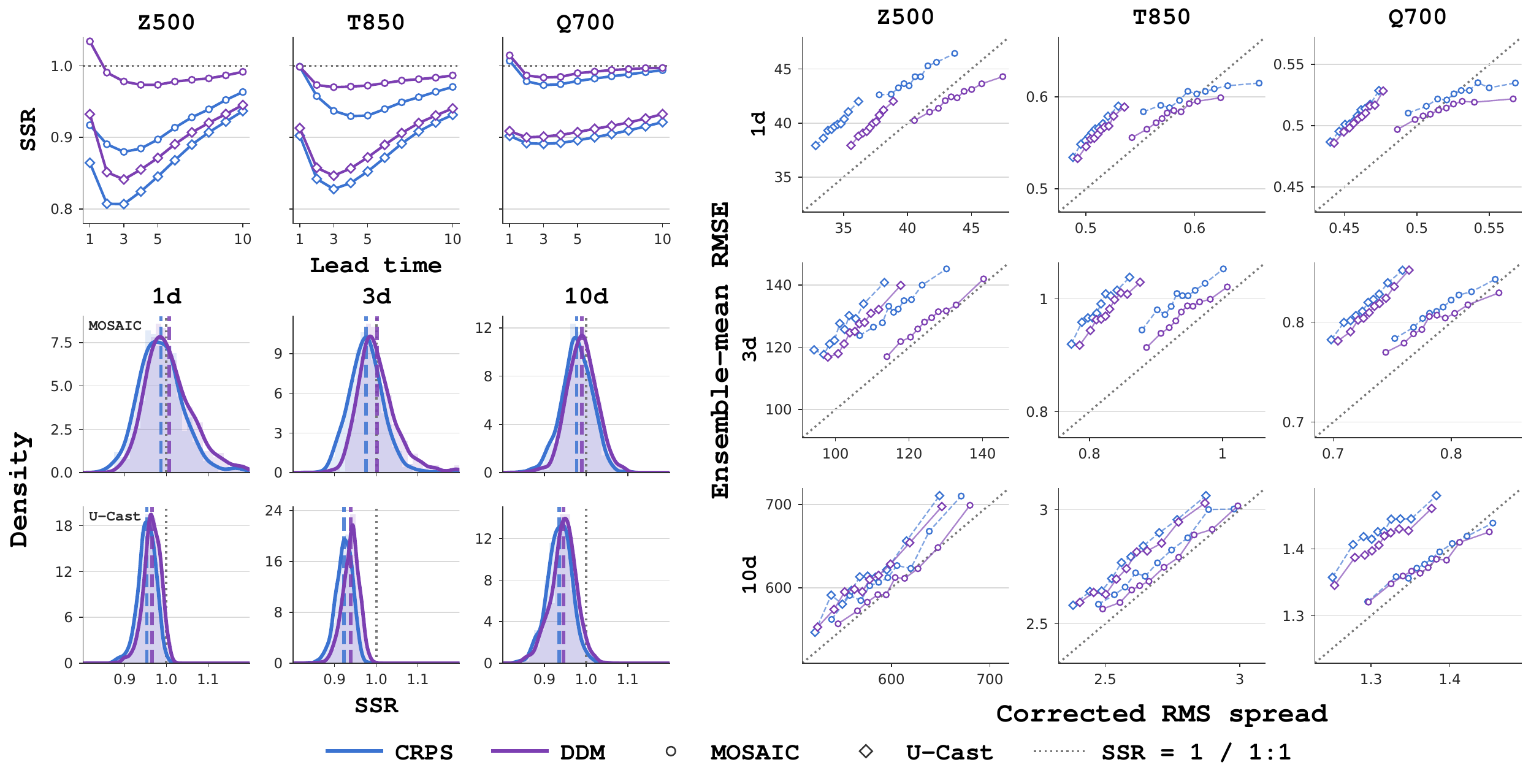}
    \caption{Ensemble calibration analysis. (Top left) SSR versus forecast lead time for Z500, T850, and Q700. (Bottom left) SSR distributions across all 84 forecast channels at 1-, 3-, and 10-day lead times. (Right) Binned spread--skill plots for the same variables at 1-, 3-, and 10-day lead times.
    }
    \label{fig:spread_skill}
    
\end{figure*}

\paragraph{Calibration Analysis }

We next examine whether the forecast-skill gains from DDM are accompanied by improved ensemble calibration. The bottom-left panels of~\Cref{fig:spread_skill} show SSR distributions across all 84 forecast channels at 1-, 3-, and 10-day lead times. DDM shifts these distributions toward $1$ across both architectures and the evaluated lead times, suggesting improved ensemble calibration.

Binned spread--skill plots assess whether forecasts with larger ensemble spread have larger errors~\citep{wang2003comparison,delle2013probabilistic}. Forecasts are binned by ensemble spread. Within each bin, we compare finite-ensemble-corrected RMS spread with the ensemble-mean RMSE. Proximity to the $1{:}1$ line indicates agreement between ensemble spread and forecast error across levels of predicted uncertainty, and DDM generally brings the curves closer to this line (\Cref{fig:spread_skill}).

\subsection{Generalization under Distribution Shift and Extreme Weather}
\label{sec:res:regularze}

In~\Cref{sec:observation}, we show that introducing auxiliary conditional denoising mitigates the deterioration in validation performance late in training (\Cref{fig:tfloor_figure}). Motivated by this observation and the potential regularizing effect of shared learning briefly discussed in~\Cref{sec:observation}, we assess the generalization ability of DDM-trained models under distribution shift and in tropical cyclone case studies.

\paragraph{Robustness to Distribution Shift}

\begin{wraptable}{r}{0.41\textwidth}
    \centering
    
    \caption{Changes in performance under the ERA5-to-HRES distribution shift.}
    \label{tab:hres_shift}
    
    \small
    \resizebox{\linewidth}{!}{%
        \setlength{\tabcolsep}{0pt}
        \setlength{\aboverulesep}{0pt}
        \setlength{\belowrulesep}{0pt}
        \renewcommand{\arraystretch}{1.5}
        \begin{tabular}{
            >{\centering\arraybackslash}p{2.0cm}
            *{2}{>{\centering\arraybackslash}p{1.8cm}}
            >{\centering\arraybackslash}p{2.2cm}
        }
            \toprule
            & \textbf{\texttt{nCRPS}} $\downarrow$
            & \textbf{\texttt{nRMSE}} $\downarrow$
            & \textbf{\texttt{SSR dev.}} $\downarrow$ \\
            \cmidrule(lr){2-2}
            \cmidrule(lr){3-3}
            \cmidrule(lr){4-4}
            Objective & Relative (\%) & Relative (\%) & Absolute \\
            \midrule

            \textbf{\texttt{CRPS}}
            & $+3.10$ & $+2.68$ & $+0.017$ \\

            \textbf{\texttt{DDM}}
            & \cellbg $+2.60$
            & \cellbg $+2.24$
            & \cellbg $+0.007$ \\

            \bottomrule
        \end{tabular}%
    }
\end{wraptable}

We first evaluate ERA5-trained MOSAIC models on HRES-fc0 without fine-tuning, using the same 732 initialization times in \texttt{2020}. Across all the channels and 1--10-day lead times, both models deteriorate relative to ERA5 evaluation, but DDM shows smaller average increases in nCRPS, nRMSE, and mean SSR deviation, as shown in~\Cref{tab:hres_shift}.
To assess the impact of this degradation, we further compare both models with IFS-ENS across 19 shared channels at the same lead times. On ERA5, CRPS- and DDM-trained MOSAIC achieve lower CRPS than IFS-ENS in 99.5\% and 100\% of these 190 channel--lead pairs, respectively. On HRES-fc0, this fraction drops to 58.4\% for CRPS-trained MOSAIC, while DDM-trained MOSAIC still achieves lower CRPS than IFS-ENS in 86.8\% of pairs. Full details are provided in~\Cref{sec:app:robustness}.

\paragraph{Tropical Cyclone Tracking}

\begin{figure*}[t]
    \centering
    \includegraphics[width=\textwidth]{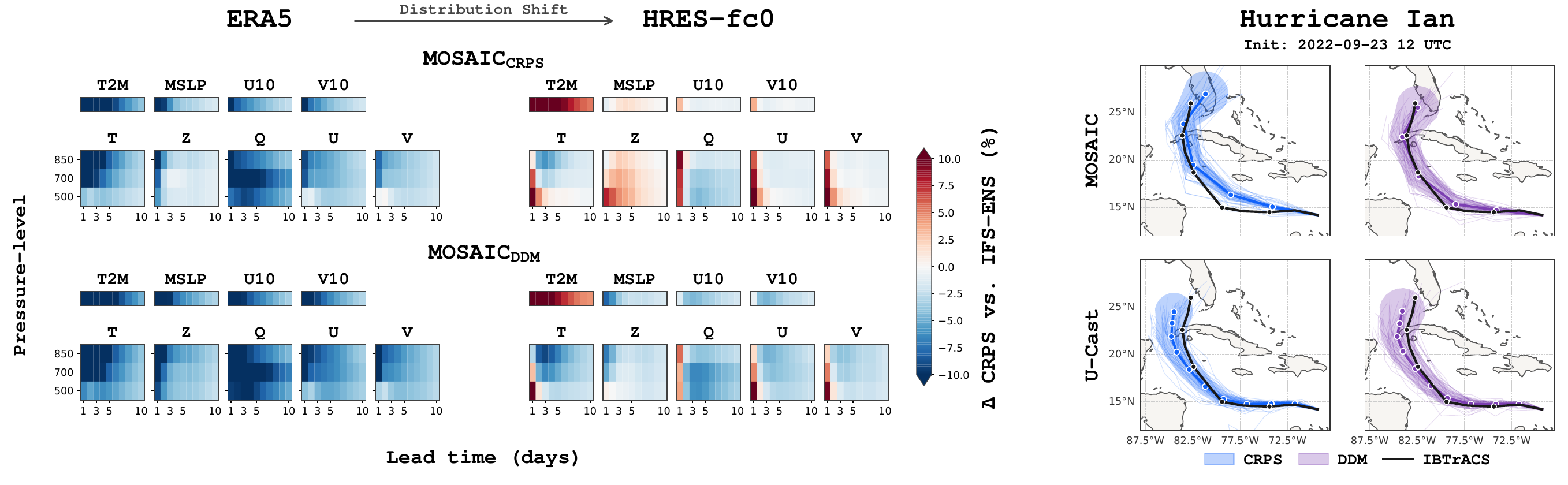}
    
    \caption{(Left) Relative CRPS differences from IFS-ENS for ERA5-trained MOSAIC evaluated on ERA5 and HRES-fc0. (Right) Ensemble track forecasts and IBTrACS best track for Hurricane Ian.
    }
    \label{fig:cyclone_cone}
    
\end{figure*}

We also evaluate 50-member ensemble forecasts for Hurricane Ian (\texttt{2022}) as an extreme-weather case study. We track cyclone centers following~\citet{zhdanov2026sparse} and compare the predicted tracks with the IBTrACS best track~\citep{knapp2010international}. \Cref{fig:cyclone_cone} shows predicted tracks and uncertainty cones enclosing 66\% of the ensemble track distribution. In this case, DDM produces track distributions closer to the IBTrACS best track for both architectures. Additional results including Hurricane Michael and quantitative tracking metrics are provided in~\Cref{sec:app:cyclone}.

\section{Conclusion}
\label{sec:conclusion}

We introduced a proper scoring rule-based diffusion framework that incorporates auxiliary conditional denoising tasks, with standard CRPS training recovered as an endpoint-only special case. Through controlled experiments, we found that jointly learning the auxiliary tasks improves one-step forecasting performance and mitigates overfitting, particularly under high uncertainty. These benefits also extend to high-dimensional global weather forecasting across different architectures and training regimes, while evaluations under distribution shift and in tropical cyclone cases suggest potential generalization benefits. Our framework achieves this with only minor conditioning changes to existing CRPS-based forecasters while retaining one-step ensemble generation. However, our work does not establish theoretical guarantees for these benefits, leaving an interesting direction for future work.

\clearpage

\bibliographystyle{plainnat}
\bibliography{paper}

@book{atkinson2012spherical,
  title={Spherical harmonics and approximations on the unit sphere: an introduction},
  author={Atkinson, Kendall and Han, Weimin},
  series={Lecture Notes in Mathematics},
  volume={2044},
  year={2012},
  publisher={Springer Heidelberg}
}

@article{vincent2010stacked,
  title={Stacked Denoising Autoencoders: Learning Useful Representations in a Deep Network with a Local Denoising Criterion},
  author={Pascal Vincent and Hugo Larochelle and Isabelle Lajoie and Yoshua Bengio and Pierre-Antoine Manzagol},
  journal={Journal of Machine Learning Research (JMLR)},
  volume={11},
  number={110},
  pages={3371--3408},
  year={2010}
}

@article{maurer2016benefit,
  title={The benefit of multitask representation learning},
  author={Maurer, Andreas and Pontil, Massimiliano and Romera-Paredes, Bernardino},
  journal={Journal of Machine Learning Research (JMLR)},
  volume={17},
  number={81},
  pages={1--32},
  year={2016}
}

@article{rasp2024weatherbench,
  title={{WeatherBench 2}: A benchmark for the next generation of data-driven global weather models},
  author={Stephan Rasp and Stephan Hoyer and Alexander Merose and Ian Langmore and Peter Battaglia and Tyler Russell and Alvaro Sanchez-Gonzalez and Vivian Yang and Rob Carver and Shreya Agrawal and Matthew Chantry and Zied Ben Bouallegue and Peter Dueben and Carla Bromberg and Jared Sisk and Luke Barrington and Aaron Bell and Fei Sha},
  journal={Journal of Advances in Modeling Earth Systems (JAMES)},
  volume={16},
  number={6},
  pages={e2023MS004019},
  year={2024},
  publisher={Wiley Online Library}
}

@article{lam2023learning,
  title={Learning skillful medium-range global weather forecasting},
  author={Remi Lam  and Alvaro Sanchez-Gonzalez  and Matthew Willson  and Peter Wirnsberger  and Meire Fortunato  and Ferran Alet  and Suman Ravuri  and Timo Ewalds  and Zach Eaton-Rosen  and Weihua Hu  and Alexander Merose  and Stephan Hoyer  and George Holland  and Oriol Vinyals  and Jacklynn Stott  and Alexander Pritzel  and Shakir Mohamed  and Peter Battaglia },
  journal={Science},
  volume={382},
  number={6677},
  pages={1416--1421},
  year={2023},
  publisher={American Association for the Advancement of Science}
}

@article{couairon2026archesweathergen,
  title={{ArchesWeatherGen}: Skillful and compute-efficient probabilistic weather forecasting with machine learning},
  author={Couairon, Guillaume and Singh, Renu and Charantonis, Anastase and Lessig, Christian and Monteleoni, Claire},
  journal={Science Advances},
  volume={12},
  number={17},
  pages={eadx2372},
  year={2026},
  publisher={American Association for the Advancement of Science}
}

@article{kochkov2024neural,
  title={Neural general circulation models for weather and climate},
  author={Kochkov, Dmitrii and Yuval, Janni and Langmore, Ian and Norgaard, Peter and Smith, Jamie and Mooers, Griffin and Klöwer, Milan and Lottes, James and Rasp, Stephan and Düben, Peter and Hatfield, Sam and Battaglia, Peter and Sanchez-Gonzalez, Alvaro and Willson, Matthew and Brenner, Michael P. and Hoyer, Stephan},
  journal={Nature},
  volume={632},
  number={8027},
  pages={1060--1066},
  year={2024},
  publisher={Nature Publishing Group UK London}
}

@article{price2025probabilistic,
  title={Probabilistic weather forecasting with machine learning},
  author={Price, Ilan and Sanchez-Gonzalez, Alvaro and Alet, Ferran and Andersson, Tom R and El-Kadi, Andrew and Masters, Dominic and Ewalds, Timo and Stott, Jacklynn and Mohamed, Shakir and Battaglia, Peter and Lam, Remi and Willson, Matthew},
  journal={Nature},
  volume={637},
  number={8044},
  pages={84--90},
  year={2025},
  publisher={Nature Publishing Group UK London}
}

@article{bi2023accurate,
  title={Accurate medium-range global weather forecasting with 3D neural networks},
  author={Bi, Kaifeng and Xie, Lingxi and Zhang, Hengheng and Chen, Xin and Gu, Xiaotao and Tian, Qi},
  journal={Nature},
  volume={619},
  number={7970},
  pages={533--538},
  year={2023},
  publisher={Nature Publishing Group UK London}
}

@article{bauer2015quiet,
  title={The quiet revolution of numerical weather prediction},
  author={Bauer, Peter and Thorpe, Alan and Brunet, Gilbert},
  journal={Nature},
  volume={525},
  number={7567},
  pages={47--55},
  year={2015},
  publisher={Nature Publishing Group UK London}
}

@article{lang2026aifs,
  title={{AIFS-CRPS}: ensemble forecasting using a model trained with a loss function based on the continuous ranked probability score},
  author={Simon Lang and Mihai Alexe and Mariana C. A. Clare and Christopher Roberts and Rilwan Adewoyin and Zied Ben Bouallègue and Matthew Chantry and Jesper Dramsch and Peter D. Dueben and Sara Hahner and Pedro Maciel and Ana Prieto-Nemesio and Cathal O'Brien and Florian Pinault and Jan Polster and Baudouin Raoult and Steffen Tietsche and Martin Leutbecher},
  journal={npj Artificial Intelligence},
  volume={2},
  number={1},
  pages={18},
  year={2026},
  publisher={Nature Publishing Group UK London}
}

@article{gneiting2007strictly,
  title={Strictly proper scoring rules, prediction, and estimation},
  author={Gneiting, Tilmann and Raftery, Adrian E},
  journal={Journal of the American statistical Association (JASA)},
  volume={102},
  number={477},
  pages={359--378},
  year={2007},
  publisher={Taylor \& Francis}
}

@article{slingo2011uncertainty,
  title={Uncertainty in weather and climate prediction},
  author={Slingo, Julia and Palmer, Tim},
  journal={Philosophical transactions. Series A, Mathematical, physical, and engineering sciences},
  volume={369},
  number={1956},
  pages={4751--4767},
  year={2011}
}

@article{gneiting2014probabilistic,
  title={Probabilistic forecasting},
  author={Gneiting, Tilmann and Katzfuss, Matthias},
  journal={Annual Review of Statistics and Its Application},
  volume={1},
  number={1},
  pages={125--151},
  year={2014},
  publisher={Annual Reviews}
}

@article{leutbecher2008ensemble,
  title={Ensemble forecasting},
  author={Leutbecher, Martin and Palmer, Tim N},
  journal={Journal of computational physics},
  volume={227},
  number={7},
  pages={3515--3539},
  year={2008},
  publisher={Elsevier}
}

@article{ferro2014fair,
  title={Fair scores for ensemble forecasts},
  author={Ferro, Christopher AT},
  journal={Quarterly Journal of the Royal Meteorological Society},
  volume={140},
  number={683},
  pages={1917--1923},
  year={2014},
  publisher={Wiley Online Library}
}

@article{matheson1976scoring,
  title={Scoring rules for continuous probability distributions},
  author={Matheson, James E and Winkler, Robert L},
  journal={Management science},
  volume={22},
  number={10},
  pages={1087--1096},
  year={1976},
  publisher={INFORMS}
}

@article{knapp2010international,
  title={The International Best Track Archive for Climate Stewardship ({IBTrACS}): Unifying Tropical Cyclone Data},
  author={Knapp, Kenneth R and Kruk, Michael C and Levinson, David H and Diamond, Howard J and Neumann, Charles J},
  journal={Bulletin of the American Meteorological Society},
  volume={91},
  number={3},
  pages={363--376},
  year={2010},
  publisher={American Meteorological Society}
}

@article{wang2003comparison,
  title={A comparison of breeding and ensemble transform Kalman filter ensemble forecast schemes},
  author={Wang, Xuguang and Bishop, Craig H},
  journal={Journal of the atmospheric sciences},
  volume={60},
  number={9},
  pages={1140--1158},
  year={2003}
}

@article{delle2013probabilistic,
  title={Probabilistic weather prediction with an analog ensemble},
  author={Delle Monache, Luca and Eckel, F Anthony and Rife, Daran L and Nagarajan, Badrinath and Searight, Keith},
  journal={Monthly Weather Review},
  volume={141},
  number={10},
  pages={3498--3516},
  year={2013}
}

@article{hersbach2020era5,
  title={The {ERA5} global reanalysis},
  author = {{Hersbach}, Hans and {Bell}, Bill and {Berrisford}, Paul and {Hirahara}, Shoji and {Hor{\'a}nyi}, Andr{\'a}s and {Mu{\~n}oz-Sabater}, Joaqu{\'\i}n and {Nicolas}, Julien and {Peubey}, Carole and {Radu}, Raluca and {Schepers}, Dinand and {Simmons}, Adrian and {Soci}, Cornel and {Abdalla}, Saleh and {Abellan}, Xavier and {Balsamo}, Gianpaolo and {Bechtold}, Peter and {Biavati}, Gionata and {Bidlot}, Jean and {Bonavita}, Massimo and {Chiara}, Giovanna and {Dahlgren}, Per and {Dee}, Dick and {Diamantakis}, Michail and {Dragani}, Rossana and {Flemming}, Johannes and {Forbes}, Richard and {Fuentes}, Manuel and {Geer}, Alan and {Haimberger}, Leo and {Healy}, Sean and {Hogan}, Robin J. and {H{\'o}lm}, El{\'\i}as and {Janiskov{\'a}}, Marta and {Keeley}, Sarah and {Laloyaux}, Patrick and {Lopez}, Philippe and {Lupu}, Cristina and {Radnoti}, Gabor and {Rosnay}, Patricia and {Rozum}, Iryna and {Vamborg}, Freja and {Villaume}, Sebastien and {Th{\'e}paut}, Jean-No{\"e}l},
  journal={Quarterly Journal of the Royal Meteorological Society},
  volume={146},
  number={730},
  pages={1999--2049},
  year={2020},
  publisher={Wiley Online Library}
}

@article{caruana1997multitask,
  title={Multitask learning},
  author={Caruana, Rich},
  journal={Machine learning},
  volume={28},
  number={1},
  pages={41--75},
  year={1997},
  publisher={Springer}
}

@article{chen2023fuxi,
  title={{FuXi}: A cascade machine learning forecasting system for 15-day global weather forecast},
  author={Chen, Lei and Zhong, Xiaohui and Zhang, Feng and Cheng, Yuan and Xu, Yinghui and Qi, Yuan and Li, Hao},
  journal={npj climate and atmospheric science},
  volume={6},
  number={1},
  pages={190},
  year={2023},
  publisher={Nature Publishing Group UK London}
}

@article{stock2026swift,
  title={{Swift}: An autoregressive consistency model for efficient weather forecasting},
  author={Stock, Jason and Arcomano, Troy and Kotamarthi, Rao},
  journal={Machine Learning: Earth},
  volume={2},
  number={2},
  pages={025004},
  year={2026},
  publisher={IOP Publishing}
}

@inproceedings{kurth2023fourcastnet,
  title={{FourCastNet}: Accelerating global high-resolution weather forecasting using adaptive fourier neural operators},
  author={Kurth, Thorsten and Subramanian, Shashank and Harrington, Peter and Pathak, Jaideep and Mardani, Morteza and Hall, David and Miele, Andrea and Kashinath, Karthik and Anandkumar, Anima},
  booktitle={Platform for Advanced Scientific Computing Conference (PASC)},
  year={2023}
}

@inproceedings{zhdanov2026sparse,
  title={(Sparse) Attention to the Details: Preserving Spectral Fidelity in {ML}-based Weather Forecasting Models},
  author={Maksim Zhdanov and Ana Lucic and Max Welling and Jan-Willem van de Meent},
  booktitle={International Conference on Machine Learning (ICML)},
  year={2026}
}

@inproceedings{cachay2026u,
  title={{U-Cast}: A Surprisingly Simple and Efficient Frontier Probabilistic {AI} Weather Forecaster},
  author={Salva R{\"u}hling Cachay and Duncan Watson-Parris and Rose Yu},
  booktitle={International Conference on Machine Learning (ICML)},
  year={2026}
}

@inproceedings{pmlr-v267-de-bortoli25b,
  title={Distributional Diffusion Models with Scoring Rules},
  author={De Bortoli, Valentin and Galashov, Alexandre and Guntupalli, J Swaroop and Zhou, Guangyao and Murphy, Kevin Patrick and Gretton, Arthur and Doucet, Arnaud},
  booktitle={International Conference on Machine Learning (ICML)},
  year={2025}
}

@inproceedings{ho2020denoising,
  title={Denoising diffusion probabilistic models},
  author={Ho, Jonathan and Jain, Ajay and Abbeel, Pieter},
  booktitle={Conference on Neural Information Processing Systems (NeurIPS)},
  year={2020}
}

@inproceedings{liu2021swin,
  title={Swin transformer: Hierarchical vision transformer using shifted windows},
  author={Liu, Ze and Lin, Yutong and Cao, Yue and Hu, Han and Wei, Yixuan and Zhang, Zheng and Lin, Stephen and Guo, Baining},
  booktitle={International Conference on Computer Vision (ICCV)},
  year={2021}
}

@inproceedings{ronneberger2015u,
  title={U-net: Convolutional networks for biomedical image segmentation},
  author={Ronneberger, Olaf and Fischer, Philipp and Brox, Thomas},
  booktitle={International Conference on Medical Image Computing and Computer Assisted Intervention (MICCAI)},
  year={2015}
}

@inproceedings{hoogeboom2023blurring,
  title={Blurring Diffusion Models},
  author={Emiel Hoogeboom and Tim Salimans},
  booktitle={International Conference on Learning Representations (ICLR)},
  year={2023}
}

@inproceedings{vincent2008extracting,
  title={Extracting and composing robust features with denoising autoencoders},
  author={Vincent, Pascal and Larochelle, Hugo and Bengio, Yoshua and Manzagol, Pierre-Antoine},
  booktitle={International Conference on Machine Learning (ICML)},
  year={2008}
}

@inproceedings{goodfellow2013multi,
  title={Multi-prediction deep Boltzmann machines},
  author={Goodfellow, Ian and Mirza, Mehdi and Courville, Aaron and Bengio, Yoshua},
  booktitle={Conference on Neural Information Processing Systems (NIPS)},
  year={2013}
}

@inproceedings{lipman2023flow,
  title={Flow Matching for Generative Modeling},
  author={Yaron Lipman and Ricky T. Q. Chen and Heli Ben-Hamu and Maximilian Nickel and Matthew Le},
  booktitle={International Conference on Learning Representations (ICLR)},
  year={2023}
}

@inproceedings{song2021denoising,
  title={Denoising Diffusion Implicit Models},
  author={Jiaming Song and Chenlin Meng and Stefano Ermon},
  booktitle={International Conference on Learning Representations (ICLR)},
  year={2021}
}

@inproceedings{andrae2025continuous,
  title={Continuous ensemble weather forecasting with diffusion models},
  author={Andrae, Martin and Landelius, Tomas and Oskarsson, Joel and Lindsten, Fredrik},
  booktitle={International Conference on Learning Representations (ICLR)},
  year={2025}
}

@inproceedings{kingma2021variational,
  title={Variational diffusion models},
  author={Kingma, Diederik and Salimans, Tim and Poole, Ben and Ho, Jonathan},
  booktitle={Conference on Neural Information Processing Systems (NeurIPS)},
  year={2021}
}

@inproceedings{sohl2015deep,
  title={Deep unsupervised learning using nonequilibrium thermodynamics},
  author={Sohl-Dickstein, Jascha and Weiss, Eric and Maheswaranathan, Niru and Ganguli, Surya},
  booktitle={International Conference on Machine Learning (ICML)},
  year={2015}
}

@misc{bonev2025fourcastnet,
  title={{FourCastNet 3}: A geometric approach to probabilistic machine-learning weather forecasting at scale},
  author={Bonev, Boris and Kurth, Thorsten and Mahesh, Ankur and Bisson, Mauro and Kossaifi, Jean and Kashinath, Karthik and Anandkumar, Anima and Collins, William D and Pritchard, Michael S and Keller, Alexander},
  url={https://arxiv.org/abs/2507.12144v2},
  year={2025}
}

@misc{alet2025skillful,
  title={Skillful joint probabilistic weather forecasting from marginals},
  author={Ferran Alet and Ilan Price and Andrew El-Kadi and Dominic Masters and Stratis Markou and Tom R. Andersson and Jacklynn Stott and Remi Lam and Matthew Willson and Alvaro Sanchez-Gonzalez and Peter Battaglia},
  url={https://arxiv.org/abs/2506.10772v1},
  year={2025}
}

@misc{gomez2026tcbench,
  title={{TCBench}: A Benchmark for Tropical Cyclone Track and Intensity Forecasting at the Global Scale},
  author={Milton Gomez and Marie McGraw and Saranya Ganesh S. and Frederick Iat-Hin Tam and Ilia Azizi and Samuel Darmon and Monika Feldmann and Stella Bourdin and Louis Poulain--Auzéau and Suzana J. Camargo and Jonathan Lin and Dan Chavas and Chia-Ying Lee and Ritwik Gupta and Andrea Jenney and Tom Beucler},
  url={https://arxiv.org/abs/2601.23268v2},
  year={2026}
}

@misc{rasp2026weathernext,
  title={{WeatherNext 3}: Increasing resolution and performance of global weather models with raw observations},
  author={Stephan Rasp and Boris Babenko and Dominic Masters and Andrew El-Kadi and Samier Merchant and Guy Shalev and Ilan Price and Fred Zyda and Remi Lam and Sasha Shysheya and Matthew Willson and Stratis Markou and Shreya Agrawal and Suhani Vora and Mohammed Alewi Hassen and Sunny Mak and Tom R. Andersson and Megan Bela and Akib Uddin and Nofar Peled Levi and Ben Gaiarin and Ferran Alet and Aaron Bell and Peter Battaglia and Alvaro Sanchez-Gonzalez},
  url={https://arxiv.org/abs/2609.03582v1},
  year={2026}
}

@misc{xu2026tyche,
  title={Tyche: One Step Flow for Efficient Probabilistic Weather Forecasting},
  author={Xu, Fan and Gao, Yuan and Wang, Kun and Su, Rui and Ling, Fenghua and Wu, Hao and Ouyang, Wanli},
  url={https://arxiv.org/abs/2605.06916v1},
  year={2026}
}

\clearpage
\appendix
\section{Related Work}
\label{sec:app:related_work}

\paragraph{Scoring Rules for Probabilistic Weather Forecasting}

Proper scoring rules serve as both evaluation metrics and training objectives for probabilistic forecasting \citep{gneiting2007strictly}. For global weather forecasting, AIFS-CRPS, FGN, and FourCastNet3 (FCN3) train stochastic predictors with CRPS-based objectives and generate samples without iterative denoising~\citep{lang2026aifs,alet2025skillful,bonev2025fourcastnet}. MOSAIC addresses spectral fidelity through its architecture and output parameterization~\citep{zhdanov2026sparse}, while U-Cast combines deterministic pretraining with CRPS-based fine-tuning and Monte Carlo dropout~\citep{cachay2026u}. WeatherNext 3 extends FGN to incorporate observations and predict multiple data modalities~\citep{rasp2026weathernext}.

\paragraph{Diffusion-based Weather Forecasting}

GenCast and ArchesWeatherGen generate ensemble members with sharp spatial features through iterative diffusion and flow-based sampling, respectively~\citep{price2025probabilistic,couairon2026archesweathergen}. Repeated model evaluations at each forecast step make generating large ensembles over long rollouts computationally expensive. Efficiency can therefore be improved by parallelizing forecast lead times or reducing the sampling cost of each forecast step. For the former, \citet{andrae2025continuous} combine lead-time conditioning with correlated noise to generate temporally coherent ensemble trajectories in parallel. For the latter, Swift~\citep{stock2026swift} and Tyche~\citep{xu2026tyche} enable one-step generation through consistency and MeanFlow-based training, respectively, followed by autoregressive CRPS fine-tuning. These methods learn a deterministic one-step transport from noise and introduce CRPS in a separate fine-tuning stage. In contrast, our formulation applies the scoring rule at every noise level throughout training and requires only additional conditioning inputs to existing CRPS-based forecasters.

\paragraph{Auxiliary Denoising for One-step Forecasting}
Distributional diffusion models learn conditional distributions of clean data given corrupted observations across noise levels~\citep{pmlr-v267-de-bortoli25b}.
In our forecasting formulation, these prediction tasks retain the same future-state target while varying the information supplied by its corrupted version.
Multi-step sampling uses intermediate denoising predictions explicitly, whereas our primary one-step setting evaluates the predictor only at the fully corrupted endpoint.
In the latter setting, intermediate denoising tasks therefore provide auxiliary supervision through shared parameters rather than additional inference steps.
This interpretation connects to denoising autoencoders, which demonstrate that reconstruction from corrupted inputs can support useful representation learning~\citep{vincent2008extracting,vincent2010stacked}, and to multitask learning, which exploits shared representations across related tasks~\citep{caruana1997multitask,goodfellow2013multi,maurer2016benefit}.
Together, these perspectives motivate the hypothesis that learning intermediate denoising conditionals can benefit endpoint forecasting through shared learning, potentially acting as an implicit regularizer even when intermediate predictions are not used at inference.

\section{Mathematical Details}

\subsection{Sampling Procedures}
\label{sec:app:sampling}

In this work, we primarily use one-step sampling ($\text{NFE}=1$). To explore the potential benefits of additional sampling steps, we follow the procedure described in~\citet{pmlr-v267-de-bortoli25b}.

\paragraph{One-step Sampling}
Given forecast context $c$, we generate $M$ ensemble members using the stochastic predictor in \eqref{eq:predictive_samples}. For each member, we independently draw corruption noise $\varepsilon^{(m)}$ from the endpoint distribution of the forward process and stochastic source $\zeta_0^{(m)}\sim p_\zeta$. The forecast is obtained through
\begin{equation}
    \hat X^{(m)}
    =
    \hat X_\theta
    \left(c,\varepsilon^{(m)},1,\zeta_0^{(m)}\right),
    \qquad m=1,\ldots,M.
    \label{eq:ddm_sampling_one_step}
\end{equation}
All members share the same context $c$ but use independent draws of $\varepsilon$ and $\zeta$.

\paragraph{Multi-step Sampling}
For $K$-step sampling, we use the uniform grid $t_k=1-k/K$ and initialize $X_{t_0}^{(m)}=\varepsilon^{(m)}$. At each step, we independently draw $\zeta_k^{(m)}\sim p_\zeta$ and predict a clean future
\begin{equation}
    \hat X_k^{(m)}
    =
    \hat X_\theta
    \left(c,X_{t_k}^{(m)},t_k,\zeta_k^{(m)}\right).
    \label{eq:ddm_sampling_prediction}
\end{equation}
For the linear noise schedule in~\Cref{sec:method}, the coupled update is
\begin{equation}
    X_{t_{k+1}}^{(m)}
    =
    \hat X_k^{(m)}
    +
    \frac{t_{k+1}}{t_k}
    \left(X_{t_k}^{(m)}-\hat X_k^{(m)}\right),
    \qquad k=0,\ldots,K-1.
    \label{eq:ddm_sampling_update}
\end{equation}
The context $c$ remains fixed throughout sampling and the stochastic source is resampled independently across members and steps. No additional corruption noise is injected between steps, but sampling remains stochastic through $\zeta_k^{(m)}$. The final forecast equals the last clean prediction, $X_{t_K}^{(m)}=\hat X_{K-1}^{(m)}$.

For the heat kernel schedule in~\Cref{sec:app:nose_ablation}, the update in~\eqref{eq:ddm_sampling_update} is applied to each spherical-harmonic coefficient with its degree dependent noise level $t_\ell(\cdot)$ from~\eqref{eq:heat_kernel_schedule}. Omitting the member index,
\begin{equation}
    (X_{t_{k+1}})_{\ell m}
    = (\hat{X}_k)_{\ell m}
    + \frac{t_\ell(t_{k+1})}{t_\ell(t_k)}
      \big( (X_{t_k})_{\ell m} - (\hat{X}_k)_{\ell m} \big).
\end{equation}

\paragraph{Exactness under Conditional Distribution Recovery}
Let $q_t(\cdot\mid c)$ denote the conditional law of $X_t=(1-t)X+t\varepsilon$, with $\varepsilon$ independent of $(c,X)$. For any finite grid $1=t_0>t_1>\cdots>t_K=0$, assume that the stochastic predictor recovers the true joint conditional distributions,
\begin{equation}
    p_\theta(\cdot\mid c,x,t_k)
    =
    p(\cdot\mid c,x,t_k),
    \qquad k=0,\ldots,K-1,
    \label{eq:app_sampling_exact_conditionals}
\end{equation}
where equality holds for almost every context $c$
and $q_{t_k}(\cdot\mid c)$-almost every $x$.

For consecutive grid points $t=t_k>s=t_{k+1}$, let $Y_t\sim q_t(\cdot\mid c)$ denote the current sampler state and $\hat X\sim p_\theta(\cdot\mid c,Y_t,t)$ its clean prediction. The update is $Y_s=(1-s/t)\hat X+(s/t)Y_t$ as written above. Exact conditional sampling restores the joint law of $(X,X_t)$, and applying this update gives
\begin{equation}
    \begin{aligned}
        \operatorname{Law}(\hat X,Y_t\mid c)
        &=
        \operatorname{Law}(X,X_t\mid c),
        \\
        \operatorname{Law}(Y_s\mid c)
        &=
        \operatorname{Law}
        \left((1-s/t)X+(s/t)X_t\mid c\right)
        \\
        &=
        \operatorname{Law}
        \left((1-s)X+s\varepsilon\mid c\right)
        =q_s(\cdot\mid c).
    \end{aligned}
    \label{eq:app_sampling_distribution_preservation}
\end{equation}
Starting from $q_1(\cdot\mid c)$, induction therefore yields the correct state distribution at every grid point and the final distribution $q_0(\cdot\mid c)=p(\cdot\mid c)$. This holds for any finite $K\geq1$, including one-step sampling, and any corruption-noise prior independent of $(c,X)$. The same argument applies coefficient-wise to the HK schedule, with $s/t$ replaced by $t_\ell(s)/t_\ell(t)$. Therefore, under exact joint conditional recovery, DDM sampling recovers the target predictive distribution regardless of NFE.

\subsection{CRPS at the Distributional Diffusion Endpoint}
\label{sec:app:crps_endpoint}

We consider component-wise CRPS with $\lambda=\beta=1$ and assume finite first moments and finite expected scores. At the endpoint, $X_1=\varepsilon$ is independent of $(c,X)$, so $p(\cdot\mid c,X_1,1)=p(\cdot\mid c)$.

Endpoint-only training places all probability mass at $t=1$. If the predictive distribution satisfies $p_\theta(\cdot\mid c,\varepsilon,1)=p_\theta(\cdot\mid c)$ for almost every $(c,\varepsilon)$, its objective reduces to standard CRPS training,
\begin{equation}
    \begin{aligned}
        \mathcal{L}_{\mathrm{end}}(\theta)
        &=
        \mathbb{E}_{c,X,\varepsilon}
        \left[
            S_{\texttt{CRPS}}
            \left(p_\theta(\cdot\mid c,\varepsilon,1),X\right)
        \right]
        \\
        &=
        \mathbb{E}_{c,X}
        \left[
            S_{\texttt{CRPS}}
            \left(p_\theta(\cdot\mid c),X\right)
        \right].
    \end{aligned}
    \label{eq:app_endpoint_crps}
\end{equation}
The first equality defines the endpoint-only objective, and the second uses independence from corruption noise. This restricts the predictive distribution and allows stochastic predictions through $\zeta$.

If the predictive distribution depends on the corruption noise $\varepsilon$, the objectives need not coincide. To show that they nevertheless share the same population-optimal forecast marginals when optimized over all predictive distributions, let $F_{\theta,j}(u\mid c,\varepsilon)$ and $F_j(u\mid c)$ denote the endpoint predictive and target CDFs evaluated at $u\in\mathbb{R}$ for each component $j\in\{1,\ldots,d\}$, respectively. The CDF representation of CRPS~\citep{gneiting2007strictly} yields the population excess risk
\begin{equation}
    \mathcal{L}_{\mathrm{end}}(\theta)-\mathcal{L}^{*}
    =
    \mathbb{E}_{c,\varepsilon}
    \left[
        \sum_{j=1}^{d}
        \int_{\mathbb{R}}
        \left(
            F_{\theta,j}(u\mid c,\varepsilon)-F_j(u\mid c)
        \right)^2
        \,\mathrm{d}u
    \right],
    \label{eq:app_endpoint_excess_risk}
\end{equation}
where $\mathcal{L}^{*}=\mathbb{E}_{c,X} [S_{\texttt{CRPS}}(p(\cdot\mid c),X)]$ is independent of $\theta$. The nonnegative excess risk vanishes exactly when all predictive marginals match their targets for almost every $(c,\varepsilon)$. Therefore, the true conditional forecast marginals minimize both the endpoint-only DDM and standard CRPS population objectives, even when fixed nonnegative channel and area weights are applied in~\Cref{sec:app:preprocessing}.

\section{Experimental Details}
\label{sec:app:exp}

\subsection{Data and Preprocessing}
\label{sec:app:preprocessing}

\paragraph{Data and Splits}

We conduct experiments at two spatial resolutions: $2.8125^\circ$ on a $64\times128$ grid for the pilot experiments, and $1.5^\circ$ on a $121\times240$ grid for the high-dimensional experiments. Both settings use 6-hourly ERA5 data, with \texttt{1979--2017} for training, \texttt{2018--2019} for validation, and \texttt{2020} for testing. For checkpoint and configuration selection, we compute validation nCRPS on a fixed set of 240 initializations evenly spaced over the validation period and shared across all runs.

\paragraph{Input and Target Normalization}

Let $k$ index channels, $X$ denote the future state, and $X^{\mathrm{cur}}$ denote the current state in the forecast context $c$. Dynamic input states are standardized channel-wise using the training-period mean $\mu_k$ and standard deviation $\sigma_k$. Target normalization depends on the prediction parameterization. FCN3~\citep{bonev2025fourcastnet} and MOSAIC~\citep{zhdanov2026sparse} predict the future state, while FGN~\citep{alet2025skillful} and U-Cast~\citep{cachay2026u} predict its residual relative to the current state. For each forecast lead, the normalized target is
\begin{equation}
    T_k =
    \begin{cases}
        (X_k-\mu_k)/\sigma_k,
        & \text{state prediction}, \\[2pt]
        (X_k-X^{\mathrm{cur}}_k)/s_k,
        & \text{residual prediction},
    \end{cases}
    \label{eq:target_normalization}
\end{equation}
where $s_k$ is the training-period standard deviation of $X_k-X^{\mathrm{cur}}_k$ for the corresponding forecast lead.

\paragraph{Spatial and Channel Weighting}
All spatially aggregated metrics use cosine-latitude weights to account for unequal grid-cell areas. Let $H$ and $W$ denote the numbers of latitude and longitude grid points, respectively, and $\phi_h$ the latitude of row $h$. We normalize the area weights to have unit mean:
\begin{equation}
    a_h = \frac{\cos(\phi_h)}{H^{-1}\sum_{h'=1}^{H}\cos(\phi_{h'})}.
\end{equation}
Then, the spatially and channel-averaged training loss is
\begin{equation}
\mathcal{L} = \frac{1}{CHW}\sum_{k=1}^{C}\sum_{h=1}^{H}\sum_{w=1}^{W}\gamma_k\, a_h\, \ell_{khw},
\end{equation}
where $\ell_{khw}$ is the grid-point loss and $\gamma_k$ is the channel weight specified below.

\begin{itemize}[leftmargin=2em]
    \item In the $2.8125^\circ$ setting, we use nine representative channels: \texttt{Z500}, \texttt{T850}, \texttt{Q700}, \texttt{U850}, \texttt{V850}, \texttt{T2M}, \texttt{U10}, \texttt{V10}, and \texttt{MSLP}. All channels are weighted uniformly ($\gamma_k = 1$).
    
    \item In the $1.5^\circ$ setting, we use the 84 dynamic state channels listed in Table~\ref{tab:state_variables}, comprising 6 upper-air variables at 13 pressure levels and 6 surface or near-surface fields. Following the pressure-level weighting convention of \citet{lam2023learning}, each upper-air channel is assigned $\gamma_k=p/\overline{p}$, where $\overline{p}\approx463.46$ hPa is the mean of the 13 pressure levels. For surface and near-surface fields, we use $\gamma_k=1.0$ for $T2m$ and $\gamma_k=0.1$ for all remaining channels, as summarized in Table~\ref{tab:loss_weights}.
\end{itemize}

\begin{table*}[t]
\centering

\begin{minipage}[t]{0.48\textwidth}
    \centering
    \footnotesize
    \captionof{table}{
        Channels used in the $1.5^\circ$ experiments.
    }
    \label{tab:state_variables}
    \vspace{0.3em}

    \setlength{\tabcolsep}{4pt}
    \begin{tabular}{@{}llr@{}}
        \toprule
        Variable & Unit & Channels \\
        \midrule
        \multicolumn{3}{l}{\textit{Surface and near-surface fields}} \\
        2-m temperature
            & K                       & 1 \\
        Mean sea-level pressure
            & Pa                      & 1 \\
        10-m zonal wind
            & $\mathrm{m\,s^{-1}}$    & 1 \\
        10-m meridional wind
            & $\mathrm{m\,s^{-1}}$    & 1 \\
        100-m zonal wind
            & $\mathrm{m\,s^{-1}}$    & 1 \\
        100-m meridional wind
            & $\mathrm{m\,s^{-1}}$    & 1 \\
        \midrule
        \multicolumn{3}{l}{\textit{Upper-air fields}} \\
        Geopotential
            & $\mathrm{m^2\,s^{-2}}$  & 13 \\
        Temperature
            & K                       & 13 \\
        Specific humidity
            & $\mathrm{kg\,kg^{-1}}$  & 13 \\
        Zonal wind
            & $\mathrm{m\,s^{-1}}$    & 13 \\
        Meridional wind
            & $\mathrm{m\,s^{-1}}$    & 13 \\
        Vertical velocity
            & $\mathrm{Pa\,s^{-1}}$   & 13 \\
        \midrule
        \textbf{Total forecast channels}
            &                         & \textbf{84} \\
        \midrule
        \multicolumn{3}{l}{\textit{Static and forcing inputs (not predicted)}} \\
        Surface geopotential
            & $\mathrm{m^2\,s^{-2}}$  & 1 \\
        Land--sea mask
            & --                      & 1 \\
        Latitude ($\sin$)
            & --                      & 1 \\
        Longitude ($\sin$, $\cos$)
            & --                      & 2 \\
        Local time of day ($\sin$, $\cos$)
            & --                      & 6 \\
        Time of year ($\sin$, $\cos$)
            & --                      & 6 \\
        \midrule
        \textbf{Total static and forcing channels}
            &                         & \textbf{17} \\
        \bottomrule
    \end{tabular}
\end{minipage}
\hfill
\begin{minipage}[t]{0.48\textwidth}
    \centering
    \footnotesize
    \captionof{table}{
        Channel weights in $1.5^\circ$ experiments.
    }
    \label{tab:loss_weights}
    \vspace{0.3em}

    \setlength{\tabcolsep}{4pt}
    \begin{tabular}{@{}lr@{}}
        \toprule
        Variable / Pressure level & Weight $\gamma_k$ \\
        \midrule
        \multicolumn{2}{l}{\textit{Surface and near-surface variables}} \\
        2-meter temperature       & 1.0 \\
        10-meter zonal wind       & 0.1 \\
        10-meter meridional wind  & 0.1 \\
        100-meter zonal wind      & 0.1 \\
        100-meter meridional wind & 0.1 \\
        Mean sea-level pressure   & 0.1 \\
        \midrule
        \multicolumn{2}{l}{\textit{Pressure levels (all upper-air variables)}} \\
        1000 hPa & 2.158 \\
         925 hPa & 1.996 \\
         850 hPa & 1.834 \\
         700 hPa & 1.510 \\
         600 hPa & 1.295 \\
         500 hPa & 1.079 \\
         400 hPa & 0.863 \\
         300 hPa & 0.647 \\
         250 hPa & 0.539 \\
         200 hPa & 0.432 \\
         150 hPa & 0.324 \\
         100 hPa & 0.216 \\
          50 hPa & 0.108 \\
        \bottomrule
    \end{tabular}
\end{minipage}

\end{table*}

\subsection{Evaluation Metrics}

We evaluate $N$ initializations indexed by $i$. For initialization $i$, a forecast lead, and channel $k$, let $\{\hat{X}_i^{(m)}\}_{m=1}^{M}$ denote the ensemble forecasts with $M\geq2$, $\bar{X}_i=M^{-1}\sum_{m=1}^{M}\hat{X}_i^{(m)}$ their mean, and $X_i$ the verifying state. We omit the lead index throughout and the channel index $k$ when unambiguous. Predictions are evaluated in state space after reconstructing states, and we use the normalized cosine-latitude weights $a_h$ defined in~\Cref{sec:app:preprocessing}.

\paragraph{CRPS}
We average the fair CRPS estimates over initializations and grid points with area weights,
\begin{equation}
\begin{aligned}
    \mathrm{CRPS}
    =
    \frac{1}{N}\sum_{i=1}^{N}
    \frac{1}{HW}
    \sum_{h=1}^{H}\sum_{w=1}^{W}
    a_h
    \Bigg[&
    \frac{1}{M}\sum_{m=1}^{M}
    \left|\hat{X}_{i,h,w}^{(m)}-X_{i,h,w}\right|
    \\
    &{}-
    \frac{1}{2M(M-1)}
    \sum_{m\neq m'}
    \left|
        \hat{X}_{i,h,w}^{(m)}-\hat{X}_{i,h,w}^{(m')}
    \right|
    \Bigg].
\end{aligned}
\end{equation}

\paragraph{Ensemble-mean RMSE}
Following WeatherBench~2~\citep{rasp2024weatherbench}, we average the area-weighted squared error of the ensemble mean over initializations before taking the square root,
\begin{equation}
    \mathrm{RMSE}
    =
    \sqrt{\frac{1}{N}\sum_{i=1}^{N}\frac{1}{HW}\sum_{h=1}^{H}\sum_{w=1}^{W} a_h\bigl(\bar{X}_{i,h,w}-X_{i,h,w}\bigr)^2 }.
\end{equation}

\paragraph{Spread--skill Ratio (SSR)}
For each initialization, we compute the unbiased ensemble variance
\begin{equation}
    v_{i,h,w}
    =
    \frac{1}{M-1}
    \sum_{m=1}^{M}
    \left(\hat{X}_{i,h,w}^{(m)}-\bar{X}_{i,h,w}\right)^2
\end{equation}
and the corresponding spatial RMS spread
\begin{equation}
    \mathrm{Spread}_i
    =
    \sqrt{
        \frac{1}{HW}
        \sum_{h=1}^{H}\sum_{w=1}^{W}
        a_h v_{i,h,w}
    }.
\end{equation}
As for RMSE, we pool squared spread over initializations before taking the square root. Following~\citet{bonev2025fourcastnet}, the finite-ensemble-corrected SSR for channel $k$ is
\begin{equation}
    \mathrm{SSR}_k
    =
    \sqrt{\frac{M+1}{M}}\,
    \frac{
        \sqrt{N^{-1}\sum_{i=1}^{N}
        \mathrm{Spread}_{i,k}^{\,2}}
    }{
        \mathrm{RMSE}_k
    }.
\end{equation}
The correction accounts for the sampling variability of the finite ensemble mean. When reporting an aggregate across channels, we average the channel-wise SSRs.

\paragraph{Normalized CRPS and Ensemble-mean RMSE}
Let $\mathrm{CRPS}_k$ and $\mathrm{RMSE}_k$ denote the metrics above for channel $k$. Their normalized aggregates are
\begin{equation}
    \mathrm{nCRPS}
    =
    \frac{1}{C}\sum_{k=1}^{C}
    \frac{\mathrm{CRPS}_k}{\sigma_k},
    \qquad
    \mathrm{nRMSE}
    =
    \frac{1}{C}\sum_{k=1}^{C}
    \frac{\mathrm{RMSE}_k}{\sigma_k}.
\end{equation}
Here $\sigma_k$ is the training-period state normalization standard deviation. These summaries weight channels equally without the training loss weights $\gamma_k$. In the case of training curves in~\Cref{fig:crps_training}, nCRPS uses the target normalization of each model, which may differ from the state normalization.

\subsection{Architectural Modifications}
\label{sec:app:architecture_modification}

Throughout this work, we adapt the forecasting backbones to DDM by adding conditioning on the corrupted target $X_t$ and noise level $t$. These modifications retain the prediction parameterization and ensemble-generation mechanism of the corresponding CRPS-based weather forecasting models. Thus, the architectural changes are limited to expanding the input layers and adding time-conditioning modules, resulting in the modest parameter increases reported in~\Cref{tab:pilot_archs,tab:1p5_backbones}.

\subsection{Experimental Details at 2.8125 degrees}
\label{sec:app:2p8}

\paragraph{Architectures}

The pilot study spans four backbones. Two are general-purpose architectures: Swin Transformer (\textbf{\texttt{Swin}})~\citep{liu2021swin}, a hierarchical window-attention transformer, and \textbf{\texttt{U-Net}}~\citep{ronneberger2015u}, a convolutional encoder--decoder with skip connections. The other two are CRPS-based probabilistic weather forecasting models: \textbf{\texttt{FGN}}~\citep{alet2025skillful}, a graph-based model that injects noise through a low-dimensional latent vector, and \textbf{\texttt{FCN3}}~\citep{bonev2025fourcastnet}, a spherical neural operator that injects noise internally at multiple spatial scales. Each backbone is trained under both objectives with matched optimization settings.

\begin{table}[h]
\centering
\footnotesize
\caption{Backbones used in the $2.8125^\circ$ pilot study.}
\label{tab:pilot_archs}
\setlength{\tabcolsep}{5.5pt}
\begin{tabular}{@{}c c c c c c@{}}
\toprule
& & & \multicolumn{3}{c}{\# Parameters} \\
\cmidrule(lr){4-6}
Model & Target & Ensemble randomness
& \textbf{\texttt{CRPS}} & \textbf{\texttt{DDM}} & \textbf{\texttt{$\Delta$}} \\
\midrule
\textbf{\texttt{Swin}}
& residual & latent vector ($d{=}32$)
& 26.82M & 26.84M & $+0.09\%$ \\
\textbf{\texttt{U-Net}}
& residual & latent vector ($d{=}32$)
& 22.75M & 22.76M & $+0.05\%$ \\
\textbf{\texttt{FGN}}
& residual & latent vector ($d{=}32$)
& 40.71M & 40.80M & $+0.21\%$ \\
\textbf{\texttt{FCN3}}
& state & internal multi-scale noise
& 17.84M & 17.85M & $+0.03\%$ \\
\bottomrule
\end{tabular}
\end{table}

\Cref{tab:pilot_archs} summarizes the model backbones, their prediction targets and noise-entry points, and the parameter counts. Switching from CRPS to DDM only expands the input layers to accept the corrupted target $X_t$ and adds time-conditioning modules for $t$.

\paragraph{Training Configuration}

The main pilot experiments use the optimization settings in~\Cref{tab:pilot_optim}. Each training step draws $M=2$ ensemble members and scores them with the fair CRPS estimator. Evaluation uses EMA weights. Each run targets a single forecast lead ($\textcolor{step12h}{\textbf{\texttt{12h}}}, \textcolor{step1d}{\textbf{\texttt{1d}}}, \textcolor{step3d}{\textbf{\texttt{3d}}}, \textcolor{step5d}{\textbf{\texttt{5d}}}$), taking its two input frames separated by the lead (FCN3 uses a single frame, following its native configuration). Widths and depths are set per backbone to a comparable scale and are held fixed across the objectives.

\begin{table}[h]
\centering
\footnotesize
\caption{Optimization settings, shared by every $2.8125^\circ$ pilot run.}
\label{tab:pilot_optim}
\setlength{\tabcolsep}{6pt}
\begin{tabular}{@{}c c@{}}
\toprule
Setting & Value \\
\midrule
Optimizer                 & AdamW \\
Learning rate             & $1\times10^{-4}$ \\
Weight decay              & $1\times10^{-4}$ \\
Epochs                    & 50 \\
Batch size                & 16 \\
Gradient clipping         & 1.0 \\
EMA decay                 & 0.999 \\
Members (M) & 2 \\
\bottomrule
\end{tabular}
\end{table}

\subsection{Experimental Details at 1.5 degrees}
\label{sec:app:1p5}

\paragraph{Architectures}

At $1.5^\circ$, we use the official implementations of MOSAIC~\citep{zhdanov2026sparse} and U-Cast~\citep{cachay2026u} to compare CRPS training with DDM. \Cref{tab:1p5_backbones} summarizes the prediction targets, ensemble-generation mechanisms, and parameter counts. DDM retains each backbone's original prediction parameterization and ensemble-generation mechanism, while additionally conditioning on the corrupted prediction target and its noise level. This conditioning increases the parameter count by $3.23\%$ for MOSAIC and $1.06\%$ for U-Cast.

\begin{table}[h]
\centering
\footnotesize
\caption{Backbones used in the $1.5^\circ$ experiments.}
\label{tab:1p5_backbones}
\setlength{\tabcolsep}{5.5pt}
\begin{tabular}{@{}c c c c c c@{}}
\toprule
& & & \multicolumn{3}{c}{\# Parameters} \\
\cmidrule(lr){4-6}
Model & Target & Ensemble randomness
& \textbf{\texttt{CRPS}}
& \textbf{\texttt{DDM}}
& $\boldsymbol{\Delta}$ \\
\midrule
\textbf{\texttt{MOSAIC}}
& state & latent vector ($d{=}32$)
& 214.19M & 221.11M & $+3.23\%$ \\
\textbf{\texttt{U-Cast}}
& residual & Monte Carlo dropout
& 895.43M & 904.96M & $+1.06\%$ \\
\bottomrule
\end{tabular}
\end{table}

\paragraph{Training Configuration}

Under our common experimental setup, we follow the published training recipes of MOSAIC and U-Cast, using only the first training stage for MOSAIC and omitting deep ensembling for U-Cast. The training settings summarized in~\Cref{tab:1p5_training} are matched between the CRPS and DDM variants of each backbone. Based on the pilot results in~\Cref{sec:app:nose_ablation}, both DDM variants use spherical Gaussian corruption noise (\texttt{SpG}) with the heat-kernel schedule ($\kappa_0=0.01$).

\begin{table}[t]
\centering
\caption{Training configurations at $1.5^\circ$.}
\label{tab:1p5_training}
\renewcommand{\arraystretch}{1.2}
\setlength{\tabcolsep}{8pt}
\begin{tabular}{@{}ccc@{}}
\toprule
& \textbf{\texttt{MOSAIC}} & \textbf{\texttt{U-Cast}} \\
\midrule
Training budget
& $250{,}000$ steps & $100$ pretraining / $8$ fine-tuning epochs \\

Effective batch size
& $16$ & $48$ \\

Optimizer
& Muon & Muon / AdamW \\

Peak learning rate
& $10^{-3}$ & $7\times10^{-3}\,/\,7\times10^{-5}$ \\

Learning-rate schedule
& cosine & linear warmup + cosine \\

Final learning rate
& $10^{-6}$ & $10^{-8}$ \\

Warmup steps
& $0$ & $1{,}500$ \\

Weight decay
& $10^{-2}$ & $0\,/\,0.1$ \\

Training ensemble size
& $2$ & $2$ \\

Gradient-norm clipping
& $1.0$ & $1.0$ \\
\bottomrule
\end{tabular}%
\end{table}

\section{Additional Results}
\label{sec:app:additional}

\subsection{Training Dynamics of Context-Only CRPS Training}
\label{sec:app:overfit}

\begin{figure}[t]
    \centering
    \includegraphics[width=\linewidth]{
        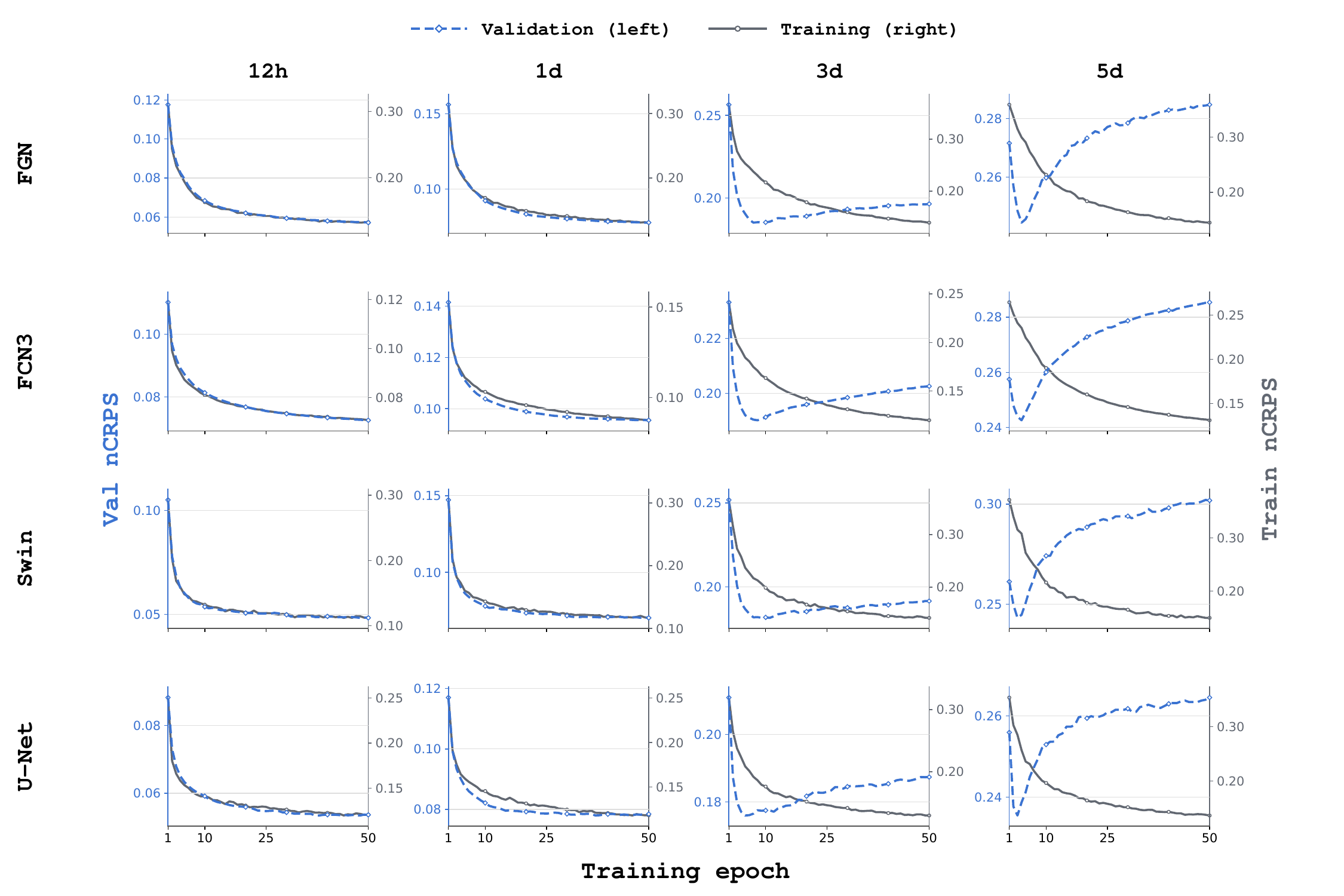
    }
    \caption{
        Training loss (solid, right axis) and validation nCRPS (dashed, left axis) of CRPS-trained FGN, FCN3, Swin, and U-Net across forecast horizons at $2.8125^\circ$. Training loss uses residual normalization for FGN, Swin, and U-Net and state normalization for FCN3, whereas validation nCRPS uses state normalization for all models. Training and validation losses therefore share the same normalization only for FCN3. Training configurations are provided in~\Cref{sec:app:2p8}.
    }
    \label{fig:crps_training}
\end{figure}

We extend the analysis in~\Cref{sec:observation} by examining training and validation trajectories across all four pilot architectures. As shown in~\Cref{fig:crps_training}, validation nCRPS decreases and gradually levels off at the 12-hour and 1-day horizons. At the 3- and 5-day horizons, it reaches an early minimum and then increases while training loss continues to decrease. This increase is larger at 5 days than at 3 days across all four architectures, indicating more pronounced overfitting at longer forecast horizons.

Training loss uses residual normalization for FGN, Swin, and U-Net and state normalization for FCN3, consistent with the prediction targets in~\Cref{tab:pilot_archs}. Validation nCRPS uses state normalization for all models. The two losses are therefore on different scales for all models except FCN3, and we assess overfitting from their trends during training. Training configurations are provided in~\Cref{sec:app:2p8}.

\subsection{Proof-of-Concept Experiment in~\texorpdfstring{\Cref{sec:observation}}{Section \ref*{sec:observation}}}
\label{sec:app:poc_experiment}

We provide additional results for the proof-of-concept experiment in~\Cref{sec:observation}, including complete test metrics for FCN3 and an extended experiment with FGN. Both models perform direct 3-day-ahead forecasting at $2.8125^\circ$ resolution. We compare the CRPS baseline with DDM trained using $t\sim\mathcal{U}[t_{\min},1]$ for $t_{\min}\in\{0,0.25,0.50,0.75\}$, together with endpoint-only training at exactly $t=1$. We train FCN3 for 50 epochs and extend FGN training from 50 to 100 epochs to better observe later training behavior. For each run, we select the checkpoint with the lowest validation nCRPS, evaluated using 240 initializations from \texttt{2018--2019} with $M=50$ ensemble members.

\Cref{tab:t_floor_fcn3,tab:t_floor_fgn} report test nCRPS and ensemble-mean nRMSE, with percentage changes relative to the corresponding CRPS baseline. Endpoint-only training ($t_{\min}=1$) closely matches CRPS training, reducing nCRPS by $0.47\%$ for FCN3 and $0.41\%$ for FGN. As discussed in~\Cref{sec:app:crps_endpoint}, endpoint-only and standard CRPS training share the same population-optimal forecast marginals. Although the learned predictive distribution need not be independent of corruption noise, the similar performance observed here is consistent with this theoretical relationship. Lowering $t_{\min}$ progressively improves both metrics for both architectures. With the full auxiliary conditional denoising tasks ($t_{\min}=0$), nCRPS and nRMSE decrease by $8.10\%$ and $6.55\%$ for FCN3, and by $9.96\%$ and $8.55\%$ for FGN.

\begin{figure*}[t]
    \centering
    \includegraphics[width=\textwidth]{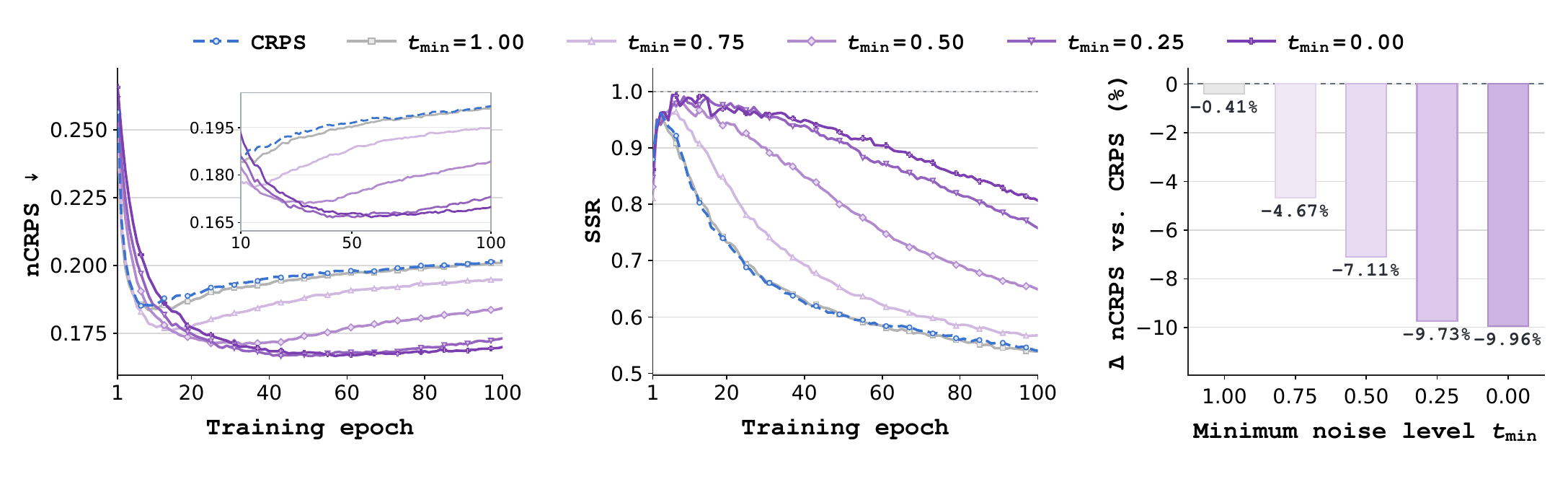}
    \caption{
    \textbf{Effect of conditional denoising tasks on FGN one-step forecasting.} FGN is trained for 100 epochs for 3-day-ahead forecasting at $2.8125^\circ$. Panels show validation nCRPS (left), validation SSR (middle), and test nCRPS changes (\%) relative to the CRPS baseline (right). Test results use checkpoints selected by validation nCRPS, with $\mathrm{NFE}=1$ for all DDM configurations.
    }
    
\label{fig:poc_fgn}
\end{figure*}

The validation trajectories in \Cref{fig:tfloor_figure,fig:poc_fgn} provide a complementary view. For CRPS and endpoint-only training, validation nCRPS increases after an initial improvement, while SSR declines substantially below $1$. Broader training noise ranges mitigate this degradation and maintain SSR closer to $1$, although underdispersion remains. This qualitative pattern is observed for both FCN3 and FGN. Together, these results are consistent with a regularizing effect of auxiliary conditional denoising tasks learned through shared parameters, supporting the hypothesis that these tasks can improve one-step forecasting and provide better generalization.

\begin{table}[t]
\centering
\footnotesize
\caption{
Effect of $t_{\min}$ on FCN3-\textcolor{step3d}{\textbf{\texttt{3d}}} one-step forecasting performance at $2.8125^\circ$ used in~\Cref{fig:tfloor_figure}.
}
\label{tab:t_floor_fcn3}
\renewcommand{\arraystretch}{1.2}
\setlength{\tabcolsep}{6pt}
\begin{tabular}{@{} l c c c @{}}
\toprule
Objective
& Training $t$
& \textbf{\texttt{nCRPS}} $\downarrow$
& \textbf{\texttt{nRMSE}} $\downarrow$
\\
\midrule

\textbf{\texttt{CRPS}}
& --
& 0.1915
& 0.3920
\\

\midrule

\multirow{5}{*}{\textbf{\texttt{DDM}}}
& $t=1$
& 0.1906 $(-0.47\%)$
& 0.3906 $(-0.36\%)$
\\

& $t\sim\mathcal{U}[0.75,1]$
& 0.1853 $(-3.21\%)$
& 0.3822 $(-2.52\%)$
\\

& $t\sim\mathcal{U}[0.50,1]$
& 0.1809 $(-5.53\%)$
& 0.3749 $(-4.37\%)$
\\

& $t\sim\mathcal{U}[0.25,1]$
& 0.1780 $(-7.03\%)$
& 0.3699 $(-5.65\%)$
\\

& $t\sim\mathcal{U}[0,1]$
& 0.1759 $(-8.10\%)$
& 0.3664 $(-6.55\%)$
\\

\bottomrule
\end{tabular}
\end{table}

\begin{table}[t]
\centering
\footnotesize
\caption{
Effect of $t_{\min}$ on FGN-\textcolor{step3d}{\textbf{\texttt{3d}}}-100ep one-step forecasting at $2.8125^\circ$
used in~\Cref{fig:poc_fgn}. Note that the best checkpoint of the CRPS baseline is the same as FGN-\textcolor{step3d}{\textbf{\texttt{3d}}}-50ep due to overfitting.
}
\label{tab:t_floor_fgn}
\renewcommand{\arraystretch}{1.2}
\setlength{\tabcolsep}{6pt}
\begin{tabular}{@{} l c c c @{}}
\toprule
Objective
& Training $t$
& \textbf{\texttt{nCRPS}} $\downarrow$
& \textbf{\texttt{nRMSE}} $\downarrow$
\\
\midrule

\textbf{\texttt{CRPS}}
& --
& 0.1868
& 0.3847
\\

\midrule

\multirow{5}{*}{\textbf{\texttt{DDM}}}
& $t=1$
& 0.1861 $(-0.41\%)$
& 0.3828 $(-0.49\%)$
\\

& $t\sim\mathcal{U}[0.75,1]$
& 0.1781 $(-4.67\%)$
& 0.3685 $(-4.20\%)$
\\

& $t\sim\mathcal{U}[0.50,1]$
& 0.1736 $(-7.11\%)$
& 0.3611 $(-6.13\%)$
\\

& $t\sim\mathcal{U}[0.25,1]$
& 0.1687 $(-9.73\%)$
& 0.3528 $(-8.27\%)$
\\

& $t\sim\mathcal{U}[0,1]$
& 0.1682 $(-9.96\%)$
& 0.3517 $(-8.55\%)$
\\

\bottomrule
\end{tabular}
\end{table}

\subsection{Additional Results for the \texorpdfstring{$2.8125^\circ$}{2.8125°} Pilot Experiments}
\label{sec:app:additional_2.8}

\subsubsection{Direct Forecasting and Autoregressive Rollout Performance}
\label{sec:app:pilot_direct_ar}

\Cref{table:pilot_single_comprehensive} reports one-step forecasting performance for all four backbones, and \Cref{table:pilot_ar_crps,table:pilot_ar_rmse} extend the rollout evaluation of FGN and FCN3. As in~\Cref{table:pilot_direct,table:pilot_ar}, DDM generally improves nCRPS and nRMSE, with larger gains at longer rollout intervals that partly persist during rollouts.

\begin{table}[t]
    \centering
    \caption{Direct forecasting performance at $2.8125^\circ$ across four architectures.}
    \label{table:pilot_single_comprehensive}
    \resizebox{\linewidth}{!}{%
    \renewcommand{\arraystretch}{1.2}
    \setlength{\tabcolsep}{8pt}
    \begin{tabular}{@{} l l c c c c c c c c @{}}
        \toprule
        & &
        \multicolumn{4}{c}{\textbf{\texttt{nCRPS}} $\downarrow$}
        & \multicolumn{4}{c}{\textbf{\texttt{nRMSE}} $\downarrow$}
        \\
        \cmidrule(lr){3-6}
        \cmidrule(lr){7-10}
        Model & Objective
        & \textcolor{step12h}{\textbf{\texttt{12h}}}
        & \textcolor{step1d}{\textbf{\texttt{1d}}}
        & \textcolor{step3d}{\textbf{\texttt{3d}}}
        & \textcolor{step5d}{\textbf{\texttt{5d}}}
        & \textcolor{step12h}{\textbf{\texttt{12h}}}
        & \textcolor{step1d}{\textbf{\texttt{1d}}}
        & \textcolor{step3d}{\textbf{\texttt{3d}}}
        & \textcolor{step5d}{\textbf{\texttt{5d}}}
        \\
        \midrule

        \multirow{3}{*}{\textbf{\texttt{Swin}}}
        & \textbf{\texttt{CRPS}}
        & 0.0491 & 0.0712 & 0.1839 & 0.2498
        & 0.0966 & 0.1429 & 0.3789 & 0.5151
        \\
        & \textbf{\texttt{DDM}}
        & \cellbg 0.0480 & \cellbg 0.0698
        & \cellbg 0.1628 & \cellbg 0.2264
        & \cellbg 0.0947 & \cellbg 0.1406
        & \cellbg 0.3409 & \cellbg 0.4739
        \\
        \cmidrule(lr){2-10}
        & \textbf{\texttt{$\Delta$(\%)}}
        & -2.1 & -2.0 & -11.5 & -9.4
        & -1.9 & -1.6 & -10.0 & -8.0
        \\
        \midrule

        \multirow{3}{*}{\textbf{\texttt{U-Net}}}
        & \textbf{\texttt{CRPS}}
        & 0.0547 & 0.0799 & 0.1759 & 0.2333
        & 0.1072 & 0.1581 & 0.3608 & 0.4782
        \\
        & \textbf{\texttt{DDM}}
        & \cellbg 0.0543 & \cellbg 0.0789
        & \cellbg 0.1687 & \cellbg 0.2226
        & \cellbg 0.1064 & \cellbg 0.1579
        & \cellbg 0.3507 & \cellbg 0.4632
        \\
        \cmidrule(lr){2-10}
        & \textbf{\texttt{$\Delta$(\%)}}
        & -0.7 & -1.3 & -4.1 & -4.6
        & -0.7 & -0.1 & -2.8 & -3.1
        \\
        \midrule

        \multirow{3}{*}{\textbf{\texttt{FGN}}}
        & \textbf{\texttt{CRPS}}
        & 0.0581 & 0.0792 & 0.1868 & 0.2463
        & 0.1138 & 0.1586 & 0.3847 & 0.5034
        \\
        & \textbf{\texttt{DDM}}
        & \cellbg 0.0577 & \cellbg 0.0778
        & \cellbg 0.1692 & \cellbg 0.2290
        & \cellbg 0.1130 & \cellbg 0.1559
        & \cellbg 0.3542 & \cellbg 0.4760
        \\
        \cmidrule(lr){2-10}
        & \textbf{\texttt{$\Delta$(\%)}}
        & -0.8 & -1.8 & -9.4 & -7.0
        & -0.7 & -1.7 & -7.9 & -5.4
        \\
        \midrule

        \multirow{3}{*}{\textbf{\texttt{FCN3}}}
        & \textbf{\texttt{CRPS}}
        & 0.0736 & 0.0967 & 0.1915 & 0.2452
        & 0.1461 & 0.1937 & 0.3920 & 0.5029
        \\
        & \textbf{\texttt{DDM}}
        & \cellbg 0.0735 & \cellbg 0.0942
        & \cellbg 0.1759 & \cellbg 0.2283
        & \cellbg 0.1461 & \cellbg 0.1894
        & \cellbg 0.3664 & \cellbg 0.4755
        \\
        \cmidrule(lr){2-10}
        & \textbf{\texttt{$\Delta$(\%)}}
        & -0.1 & -2.6 & -8.1 & -6.9
        & -0.0 & -2.2 & -6.6 & -5.5
        \\
        \bottomrule
    \end{tabular}%
    }
\end{table}

\begin{table}[t]
    \centering
    \caption{Autoregressive rollout nCRPS at $2.8125^\circ$.}
    \label{table:pilot_ar_crps}
    \renewcommand{\arraystretch}{1.2}
    \setlength{\tabcolsep}{8pt}
    \begin{tabular}{@{} c c l c c c c c c @{}}
        \toprule
        Model & RI & Objective
        & {{\textbf{\texttt{24h}}}}
        & {{\textbf{\texttt{72h}}}}
        & {{\textbf{\texttt{120h}}}}
        & {{\textbf{\texttt{240h}}}}
        & {{\textbf{\texttt{288h}}}}
        & {{\textbf{\texttt{360h}}}} \\
        \midrule
        \multirow{14}{*}[0pt]{\textbf{\texttt{FGN}}}
        & \multirow{3}{*}{\textcolor{step12h}{\textbf{\texttt{12h}}}}
        & \textbf{\texttt{CRPS}} & 0.0747 & 0.1420 & 0.1928 & 0.2442 & 0.2514 & 0.2571 \\
        & & \textbf{\texttt{DDM}} & \cellbg 0.0743 & \cellbg 0.1419 & \cellbg 0.1926 & 0.2443 & 0.2514 & \cellbg 0.2570 \\
        \cmidrule(lr){3-9}
        & & \textbf{\texttt{$\Delta$(\%)}} & -0.5 & -0.1 & -0.1 & +0.0 & +0.0 & -0.0 \\
        \cmidrule(lr){2-9}
        & \multirow{3}{*}{\textcolor{step1d}{\textbf{\texttt{1d}}}}
        & \textbf{\texttt{CRPS}} & 0.0792 & 0.1434 & 0.1934 & 0.2444 & 0.2509 & 0.2566 \\
        & & \textbf{\texttt{DDM}} & \cellbg 0.0778 & \cellbg 0.1412 & \cellbg 0.1910 & \cellbg 0.2428 & \cellbg 0.2499 & \cellbg 0.2559 \\
        \cmidrule(lr){3-9}
        & & \textbf{\texttt{$\Delta$(\%)}} & -1.8 & -1.6 & -1.2 & -0.7 & -0.4 & -0.3 \\
        \cmidrule(lr){2-9}
        & \multirow{3}{*}{\textcolor{step3d}{\textbf{\texttt{3d}}}}
        & \textbf{\texttt{CRPS}} & $-$ & 0.1868 & $-$ & $-$ & 0.2628 & 0.2659 \\
        & & \textbf{\texttt{DDM}} & $-$ & \cellbg 0.1692 & $-$ & $-$ & \cellbg 0.2538 & \cellbg 0.2582 \\
        \cmidrule(lr){3-9}
        & & \textbf{\texttt{$\Delta$(\%)}} & $-$ & -9.4 & $-$ & $-$ & -3.4 & -2.9 \\
        \cmidrule(lr){2-9}
        & \multirow{3}{*}{\textcolor{step5d}{\textbf{\texttt{5d}}}}
        & \textbf{\texttt{CRPS}} & $-$ & $-$ & 0.2463 & 0.2652 & $-$ & 0.2701 \\
        & & \textbf{\texttt{DDM}} & $-$ & $-$ & \cellbg 0.2290 & \cellbg 0.2557 & $-$ & \cellbg 0.2621 \\
        \cmidrule(lr){3-9}
        & & \textbf{\texttt{$\Delta$(\%)}} & $-$ & $-$ & -7.0 & -3.6 & $-$ & -3.0 \\
        \midrule
        \multirow{14}{*}[0pt]{\textbf{\texttt{FCN3}}}
        & \multirow{3}{*}{\textcolor{step12h}{\textbf{\texttt{12h}}}}
        & \textbf{\texttt{CRPS}} & 0.0922 & 0.1634 & 0.2105 & 0.2512 & 0.2560 & 0.2602 \\
        & & \textbf{\texttt{DDM}} & 0.0923 & \cellbg 0.1632 & \cellbg 0.2100 & \cellbg 0.2507 & \cellbg 0.2556 & \cellbg 0.2600 \\
        \cmidrule(lr){3-9}
        & & \textbf{\texttt{$\Delta$(\%)}} & +0.1 & -0.1 & -0.2 & -0.2 & -0.2 & -0.1 \\
        \cmidrule(lr){2-9}
        & \multirow{3}{*}{\textcolor{step1d}{\textbf{\texttt{1d}}}}
        & \textbf{\texttt{CRPS}} & 0.0967 & 0.1641 & 0.2107 & 0.2513 & 0.2561 & 0.2608 \\
        & & \textbf{\texttt{DDM}} & \cellbg 0.0942 & \cellbg 0.1597 & \cellbg 0.2058 & \cellbg 0.2487 & \cellbg 0.2541 & \cellbg 0.2588 \\
        \cmidrule(lr){3-9}
        & & \textbf{\texttt{$\Delta$(\%)}} & -2.6 & -2.7 & -2.3 & -1.0 & -0.8 & -0.8 \\
        \cmidrule(lr){2-9}
        & \multirow{3}{*}{\textcolor{step3d}{\textbf{\texttt{3d}}}}
        & \textbf{\texttt{CRPS}} & $-$ & 0.1915 & $-$ & $-$ & 0.2620 & 0.2656 \\
        & & \textbf{\texttt{DDM}} & $-$ & \cellbg 0.1759 & $-$ & $-$ & \cellbg 0.2544 & \cellbg 0.2585 \\
        \cmidrule(lr){3-9}
        & & \textbf{\texttt{$\Delta$(\%)}} & $-$ & -8.1 & $-$ & $-$ & -2.9 & -2.7 \\
        \cmidrule(lr){2-9}
        & \multirow{3}{*}{\textcolor{step5d}{\textbf{\texttt{5d}}}}
        & \textbf{\texttt{CRPS}} & $-$ & $-$ & 0.2452 & 0.2612 & $-$ & 0.2662 \\
        & & \textbf{\texttt{DDM}} & $-$ & $-$ & \cellbg 0.2283 & \cellbg 0.2527 & $-$ & \cellbg 0.2593 \\
        \cmidrule(lr){3-9}
        & & \textbf{\texttt{$\Delta$(\%)}} & $-$ & $-$ & -6.9 & -3.2 & $-$ & -2.6 \\
        \bottomrule
    \end{tabular}%
\end{table}

\begin{table}[t]
    \centering
    \caption{Autoregressive rollout nRMSE at $2.8125^\circ$. The same runs and
    conventions as \Cref{table:pilot_ar_crps}.}
    \label{table:pilot_ar_rmse}
    \renewcommand{\arraystretch}{1.2}
    \setlength{\tabcolsep}{8pt}
    \begin{tabular}{@{} c c l c c c c c c @{}}
        \toprule
        Model & RI & Objective
        & {{\textbf{\texttt{24h}}}}
        & {{\textbf{\texttt{72h}}}}
        & {{\textbf{\texttt{120h}}}}
        & {{\textbf{\texttt{240h}}}}
        & {{\textbf{\texttt{288h}}}}
        & {{\textbf{\texttt{360h}}}} \\
        \midrule
        \multirow{14}{*}[0pt]{\textbf{\texttt{FGN}}}
        & \multirow{3}{*}{\textcolor{step12h}{\textbf{\texttt{12h}}}}
        & \textbf{\texttt{CRPS}} & 0.1497 & 0.2986 & 0.4060 & 0.5082 & 0.5220 & 0.5332 \\
        & & \textbf{\texttt{DDM}} & \cellbg 0.1490 & \cellbg 0.2985 & \cellbg 0.4056 & 0.5085 & 0.5223 & \cellbg 0.5329 \\
        \cmidrule(lr){3-9}
        & & \textbf{\texttt{$\Delta$(\%)}} & -0.4 & -0.0 & -0.1 & +0.1 & +0.1 & -0.0 \\
        \cmidrule(lr){2-9}
        & \multirow{3}{*}{\textcolor{step1d}{\textbf{\texttt{1d}}}}
        & \textbf{\texttt{CRPS}} & 0.1586 & 0.3012 & 0.4066 & 0.5083 & 0.5210 & 0.5321 \\
        & & \textbf{\texttt{DDM}} & \cellbg 0.1559 & \cellbg 0.2978 & \cellbg 0.4031 & \cellbg 0.5062 & \cellbg 0.5199 & \cellbg 0.5314 \\
        \cmidrule(lr){3-9}
        & & \textbf{\texttt{$\Delta$(\%)}} & -1.7 & -1.1 & -0.9 & -0.4 & -0.2 & -0.1 \\
        \cmidrule(lr){2-9}
        & \multirow{3}{*}{\textcolor{step3d}{\textbf{\texttt{3d}}}}
        & \textbf{\texttt{CRPS}} & $-$ & 0.3847 & $-$ & $-$ & 0.5380 & 0.5441 \\
        & & \textbf{\texttt{DDM}} & $-$ & \cellbg 0.3542 & $-$ & $-$ & \cellbg 0.5258 & \cellbg 0.5345 \\
        \cmidrule(lr){3-9}
        & & \textbf{\texttt{$\Delta$(\%)}} & $-$ & -7.9 & $-$ & $-$ & -2.3 & -1.8 \\
        \cmidrule(lr){2-9}
        & \multirow{3}{*}{\textcolor{step5d}{\textbf{\texttt{5d}}}}
        & \textbf{\texttt{CRPS}} & $-$ & $-$ & 0.5034 & 0.5408 & $-$ & 0.5501 \\
        & & \textbf{\texttt{DDM}} & $-$ & $-$ & \cellbg 0.4760 & \cellbg 0.5277 & $-$ & \cellbg 0.5402 \\
        \cmidrule(lr){3-9}
        & & \textbf{\texttt{$\Delta$(\%)}} & $-$ & $-$ & -5.4 & -2.4 & $-$ & -1.8 \\
        \midrule
        \multirow{14}{*}[0pt]{\textbf{\texttt{FCN3}}}
        & \multirow{3}{*}{\textcolor{step12h}{\textbf{\texttt{12h}}}}
        & \textbf{\texttt{CRPS}} & 0.1853 & 0.3399 & 0.4391 & 0.5209 & 0.5303 & 0.5384 \\
        & & \textbf{\texttt{DDM}} & 0.1855 & \cellbg 0.3394 & \cellbg 0.4378 & \cellbg 0.5197 & \cellbg 0.5293 & \cellbg 0.5379 \\
        \cmidrule(lr){3-9}
        & & \textbf{\texttt{$\Delta$(\%)}} & +0.1 & -0.1 & -0.3 & -0.2 & -0.2 & -0.1 \\
        \cmidrule(lr){2-9}
        & \multirow{3}{*}{\textcolor{step1d}{\textbf{\texttt{1d}}}}
        & \textbf{\texttt{CRPS}} & 0.1937 & 0.3401 & 0.4381 & 0.5207 & 0.5301 & 0.5394 \\
        & & \textbf{\texttt{DDM}} & \cellbg 0.1894 & \cellbg 0.3327 & \cellbg 0.4297 & \cellbg 0.5164 & \cellbg 0.5270 & \cellbg 0.5361 \\
        \cmidrule(lr){3-9}
        & & \textbf{\texttt{$\Delta$(\%)}} & -2.2 & -2.2 & -1.9 & -0.8 & -0.6 & -0.6 \\
        \cmidrule(lr){2-9}
        & \multirow{3}{*}{\textcolor{step3d}{\textbf{\texttt{3d}}}}
        & \textbf{\texttt{CRPS}} & $-$ & 0.3920 & $-$ & $-$ & 0.5368 & 0.5437 \\
        & & \textbf{\texttt{DDM}} & $-$ & \cellbg 0.3664 & $-$ & $-$ & \cellbg 0.5273 & \cellbg 0.5354 \\
        \cmidrule(lr){3-9}
        & & \textbf{\texttt{$\Delta$(\%)}} & $-$ & -6.6 & $-$ & $-$ & -1.8 & -1.5 \\
        \cmidrule(lr){2-9}
        & \multirow{3}{*}{\textcolor{step5d}{\textbf{\texttt{5d}}}}
        & \textbf{\texttt{CRPS}} & $-$ & $-$ & 0.5029 & 0.5357 & $-$ & 0.5445 \\
        & & \textbf{\texttt{DDM}} & $-$ & $-$ & \cellbg 0.4755 & \cellbg 0.5243 & $-$ & \cellbg 0.5373 \\
        \cmidrule(lr){3-9}
        & & \textbf{\texttt{$\Delta$(\%)}} & $-$ & $-$ & -5.5 & -2.1 & $-$ & -1.3 \\
        \bottomrule
    \end{tabular}%
\end{table}

\subsubsection{Noise Prior and Schedule Ablation}
\label{sec:app:nose_ablation}

To assess the sensitivity of one-step forecasting to the corruption process, we compare spatial noise priors and corruption schedules while keeping the remaining DDM settings fixed (\Cref{table:pilot_noise}). Here, $x$ is the normalized prediction target and $\varepsilon$ is corruption noise, distinct from the randomness source $\zeta$.

Standard diffusion models typically use independent Gaussian noise at each grid point (\texttt{Grid})~\citep{ho2020denoising}. GenCast reports small but consistent improvements from sampling isotropic Gaussian noise in the spherical-harmonic domain~\citep{price2025probabilistic}.
Motivated by this finding, we evaluate spherical Gaussian (\texttt{SpG}) noise with equal variance across retained harmonic modes, generated as
\begin{equation}
\varepsilon_{\mathrm{SpG}}(\omega)
=
g\sum_{\ell=0}^{L_{\mathrm{noise}}}
\sum_{m=-\ell}^{\ell}
a_{\ell m}\,\psi_{\ell m}(\omega),
\qquad
a_{\ell m}\overset{\mathrm{i.i.d.}}{\sim}\mathcal{N}(0,1),
\end{equation}
where $\omega\in\mathbb{S}^{2}$, $\psi_{\ell m}$ are real orthonormal spherical harmonics, $L_{\mathrm{noise}}$ is the maximum degree, and $g$ is a normalization constant. We sample this isotropic Gaussian field independently across channels.

\begin{wrapfigure}[12]{r}{0.35\linewidth}
    \centering
    
    \includegraphics[width=\linewidth]{
        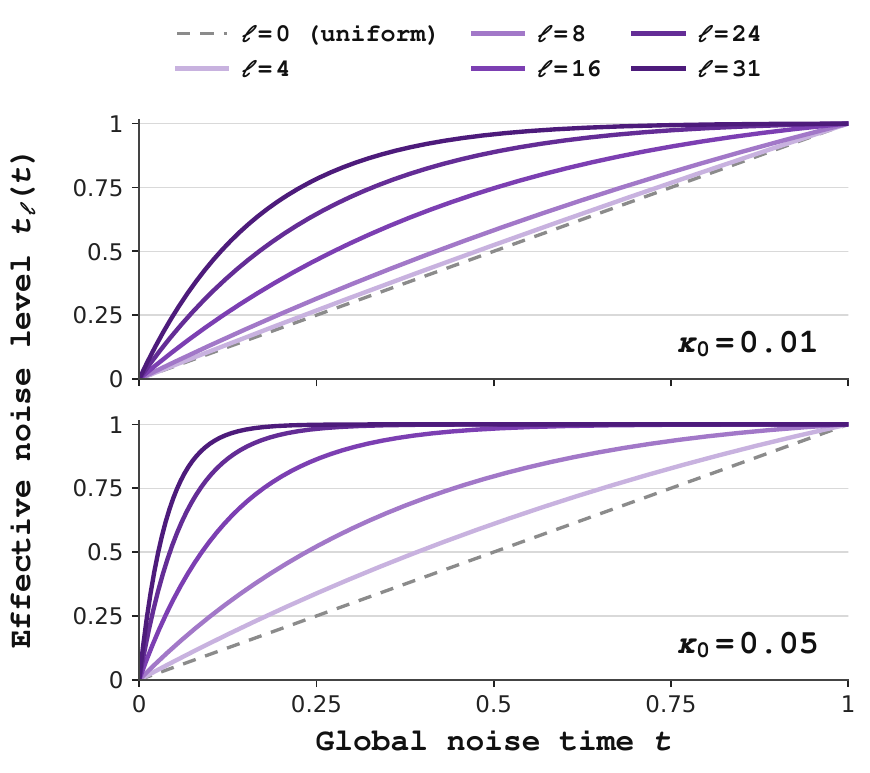
    }
    \caption{
        Effective noise levels by harmonic degree
        for $\kappa_0=0.01$ (top) and $\kappa_0=0.05$ (bottom).
    }
    \label{fig:heat_kernel_schedule}
\end{wrapfigure}

Linear interpolation attenuates all target coefficients by the same factor. We therefore consider a spherical heat-kernel (\texttt{HK}) schedule, inspired by~\citet{hoogeboom2023blurring}, that attenuates fine-scale target information more rapidly.

Since spherical harmonics of degree $\ell$ have Laplace--Beltrami eigenvalue $-\ell(\ell+1)$ on the unit sphere~\citep{atkinson2012spherical}, we define the interpolation for degrees $0\leq\ell\leq L_{\mathrm{HK}}$ as
\begin{equation}
\begin{aligned}
t_\ell(t)
&=
1-(1-t)\exp\!\left[
-\frac{\kappa_0 t}{2}\ell(\ell+1)
\right],\\
(x_t)_{\ell m}
&=
\bigl[1-t_\ell(t)\bigr]x_{\ell m}
+t_\ell(t)\varepsilon_{\ell m}.
\end{aligned}
\label{eq:heat_kernel_schedule}
\end{equation}

Here, $x_{\ell m}$ and $\varepsilon_{\ell m}$ are the spherical-harmonic coefficients of the target and noise, $L_{\mathrm{HK}}$ is the maximum degree, and $\kappa_0\geq0$ controls the attenuation strength. As shown in~\Cref{fig:heat_kernel_schedule}, higher degrees have larger $t_\ell(t)$ and smaller target weights $1-t_\ell(t)$ at the same $t$. Larger $\kappa_0$ strengthens this scale dependence. 

In practice, $g$ is computed analytically to ensure unit grid-averaged marginal variance. We use $(L_{\mathrm{noise}},L_{\mathrm{HK}})=(63,31)$ at $2.8125^\circ$ and $(120,60)$ at $1.5^\circ$. The residual outside the retained representation is preserved and interpolated using $t_{L_{\mathrm{HK}}}(t)$. The \texttt{Grid} and \texttt{SpG} variants use linear interpolation, with independent standard Gaussian grid noise and spherical Gaussian noise, respectively. The \texttt{HK} variants use the same spherical Gaussian prior with $\kappa_0 \in \{0.01, 0.05\}$.

\Cref{table:pilot_noise} shows that \texttt{HK} generally achieves the lowest nCRPS at \textcolor{step12h}{\textbf{\texttt{12h}}} and \textcolor{step1d}{\textbf{\texttt{1d}}}, whereas \texttt{Grid} or \texttt{SpG} performs best at \textcolor{step3d}{\textbf{\texttt{3d}}} and \textcolor{step5d}{\textbf{\texttt{5d}}}.
Thus, the preferred noise distribution and schedule depend on the rollout interval and backbone. Nevertheless, all DDM configurations outperform the CRPS baseline at the two longer intervals across all four backbones. For the main pilot experiments, we select the noise prior and schedule with the lowest validation nCRPS for each backbone and forecast horizon.

\begin{table}[t]
    \centering
    \caption{Noise scheduling for each backbone and RI at $2.8125^\circ$, with the single-step nCRPS. Values are test nCRPS. Shaded cells denote the configuration selected by the lowest validation nCRPS.}
    \label{table:pilot_noise}
    \renewcommand{\arraystretch}{1.2}
    \setlength{\tabcolsep}{6pt}
    \begin{tabular}{@{} c c c c c c c c @{}}
        \toprule
        & & \textbf{\texttt{CRPS}}
          & \multicolumn{4}{c}{\textbf{\texttt{DDM}}}
          & \\
        \cmidrule(lr){3-3} \cmidrule(lr){4-7}
        Model & RI & baseline
        & \texttt{grid} & \texttt{SpG}
        & \texttt{HK} $\kappa_0{=}0.01$ & \texttt{HK} $\kappa_0{=}0.05$
        & $\Delta$(\%) \\
        \midrule
        \multirow{4}{*}{\textbf{\texttt{Swin}}}
        & \textcolor{step12h}{\textbf{\texttt{12h}}} & 0.0491 & 0.0489 & 0.0496 & 0.0484 & \cellbg 0.0480 & -2.1 \\
        & \textcolor{step1d}{\textbf{\texttt{1d}}}   & 0.0712 & 0.0712 & 0.0708 & \cellbg 0.0698 & 0.0704 & -2.0 \\
        & \textcolor{step3d}{\textbf{\texttt{3d}}}   & 0.1839 & 0.1644 & \cellbg 0.1628 & 0.1645 & 0.1705 & -11.5 \\
        & \textcolor{step5d}{\textbf{\texttt{5d}}}   & 0.2498 & \cellbg 0.2264 & 0.2268 & 0.2297 & 0.2358 & -9.4 \\
        \midrule
        \multirow{4}{*}{\textbf{\texttt{U-Net}}}
        & \textcolor{step12h}{\textbf{\texttt{12h}}} & 0.0547 & 0.0553 & 0.0561 & 0.0548 & \cellbg 0.0543 & -0.7 \\
        & \textcolor{step1d}{\textbf{\texttt{1d}}}   & 0.0799 & \cellbg 0.0789 & 0.0797 & 0.0800 & 0.0801 & -1.3 \\
        & \textcolor{step3d}{\textbf{\texttt{3d}}}   & 0.1759 & \cellbg 0.1687 & 0.1692 & 0.1692 & 0.1718 & -4.1 \\
        & \textcolor{step5d}{\textbf{\texttt{5d}}}   & 0.2333 & \cellbg 0.2226 & 0.2237 & 0.2268 & 0.2280 & -4.6 \\
        \midrule
        \multirow{4}{*}{\textbf{\texttt{FGN}}}
        & \textcolor{step12h}{\textbf{\texttt{12h}}} & 0.0581 & 0.0592 & 0.0592 & \cellbg 0.0577 & 0.0578 & -0.8 \\
        & \textcolor{step1d}{\textbf{\texttt{1d}}}   & 0.0792 & 0.0797 & 0.0793 & \cellbg 0.0778 & 0.0782 & -1.8 \\
        & \textcolor{step3d}{\textbf{\texttt{3d}}}   & 0.1868 & 0.1697 & \cellbg 0.1692 & 0.1704 & 0.1775 & -9.4 \\
        & \textcolor{step5d}{\textbf{\texttt{5d}}}   & 0.2463 & 0.2298 & \cellbg 0.2290 & 0.2303 & 0.2356 & -7.0 \\
        \midrule
        \multirow{4}{*}{\textbf{\texttt{FCN3}}}
        & \textcolor{step12h}{\textbf{\texttt{12h}}} & 0.0736 & 0.0743 & 0.0740 & \cellbg 0.0735 & 0.0736 & -0.1 \\
        & \textcolor{step1d}{\textbf{\texttt{1d}}}   & 0.0967 & 0.0956 & 0.0950 & \cellbg 0.0942 & 0.0945 & -2.6 \\
        & \textcolor{step3d}{\textbf{\texttt{3d}}}   & 0.1915 & 0.1761 & \cellbg 0.1759 & 0.1784 & 0.1845 & -8.1 \\
        & \textcolor{step5d}{\textbf{\texttt{5d}}}   & 0.2452 & \cellbg 0.2283 & 0.2295 & 0.2314 & 0.2372 & -6.9 \\
        \bottomrule
    \end{tabular}%
\end{table}

\subsection{Additional Results at \texorpdfstring{$1.5^\circ$}{1.5°} High-dimensional Global Weather Forecasting}

\subsubsection{Forecast Performance}
\label{sec:app:forecast_performance}

\begin{figure*}[t]
    \centering
    \includegraphics[width=\textwidth]{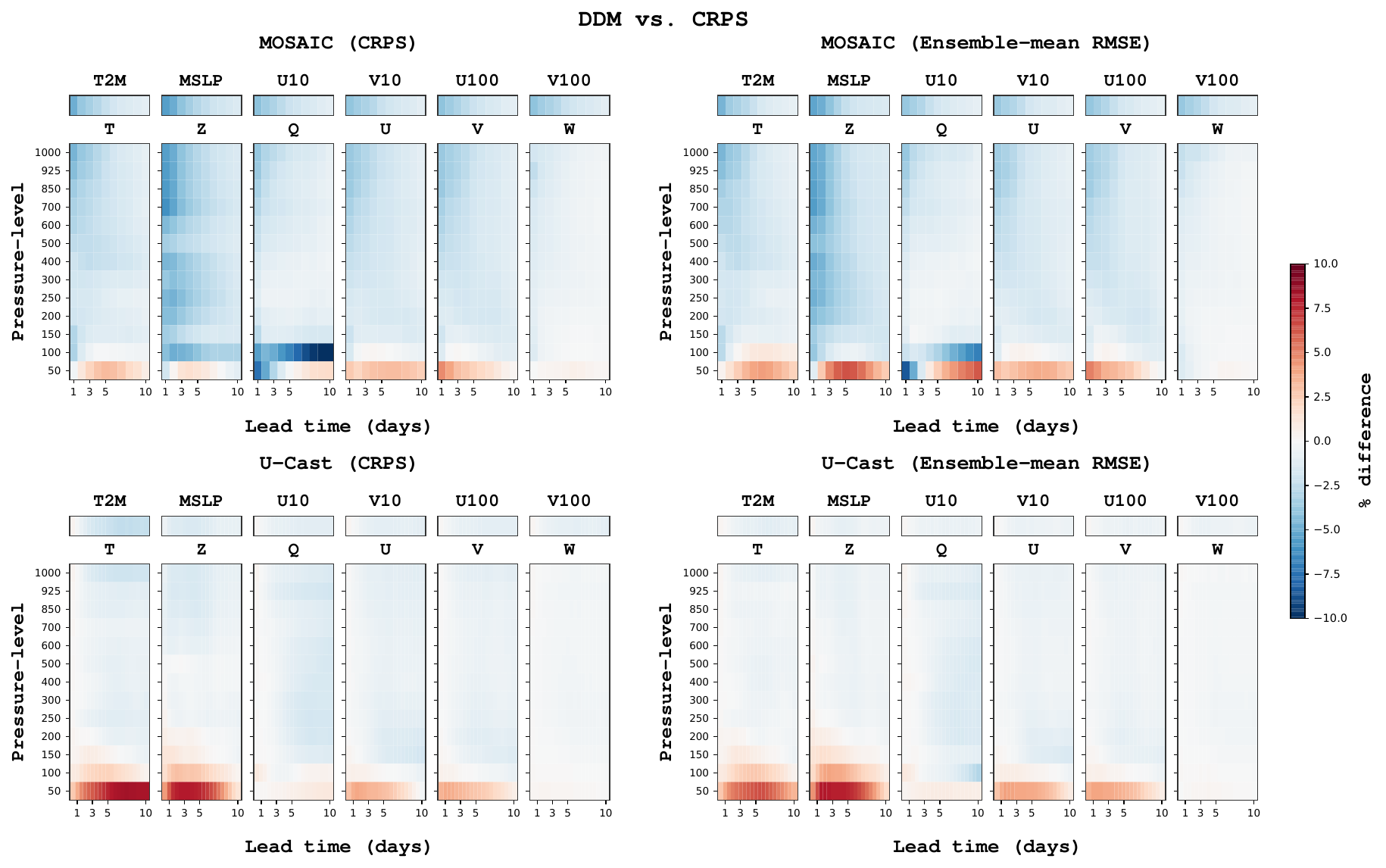}
    \caption{Relative performance differences in CRPS (left) and ensemble-mean RMSE (right) between DDM and CRPS training for MOSAIC (top) and U-Cast (bottom). Negative values (blue) indicate that DDM shows better performance than CRPS.
    }
    \label{fig:scorecard_2x2}
\end{figure*}

We provide channel-wise comparisons and rollout curves using all 732 initializations at 00 and 12 UTC in \texttt{2020}, with $M=50$ ensemble members and $\text{NFE}=1$ for DDM. Figure~\ref{fig:scorecard_2x2} compares DDM and CRPS training across all 84 forecast channels over 10-day rollouts. In addition to IFS-ENS, \Cref{tab:1p5deg_analysis} reports results for GenCast~\citep{price2025probabilistic} and NeuralGCM~\citep{kochkov2024neural}. These baselines are included for reference, as their configurations, including spatial resolution, differ from ours. Our primary focus is comparing DDM and CRPS training under matched experimental settings.

Comparisons use native forecast intervals of 24\,h for MOSAIC and 12\,h for U-Cast, yielding 840 and 1,680 channel--lead pairs, respectively. DDM achieves lower CRPS in $93.7\%$ (787/840) of these pairs for MOSAIC and $82.0\%$ (1,377/1,680) for U-Cast. The corresponding fractions for ensemble-mean RMSE are $92.0\%$ (773/840) and $81.0\%$ (1,360/1,680). Improvements extend across surface variables and most tropospheric levels, whereas degradations are concentrated at the lowest-pressure levels.

\begin{figure*}[t]
    \centering
    \includegraphics[width=\textwidth]{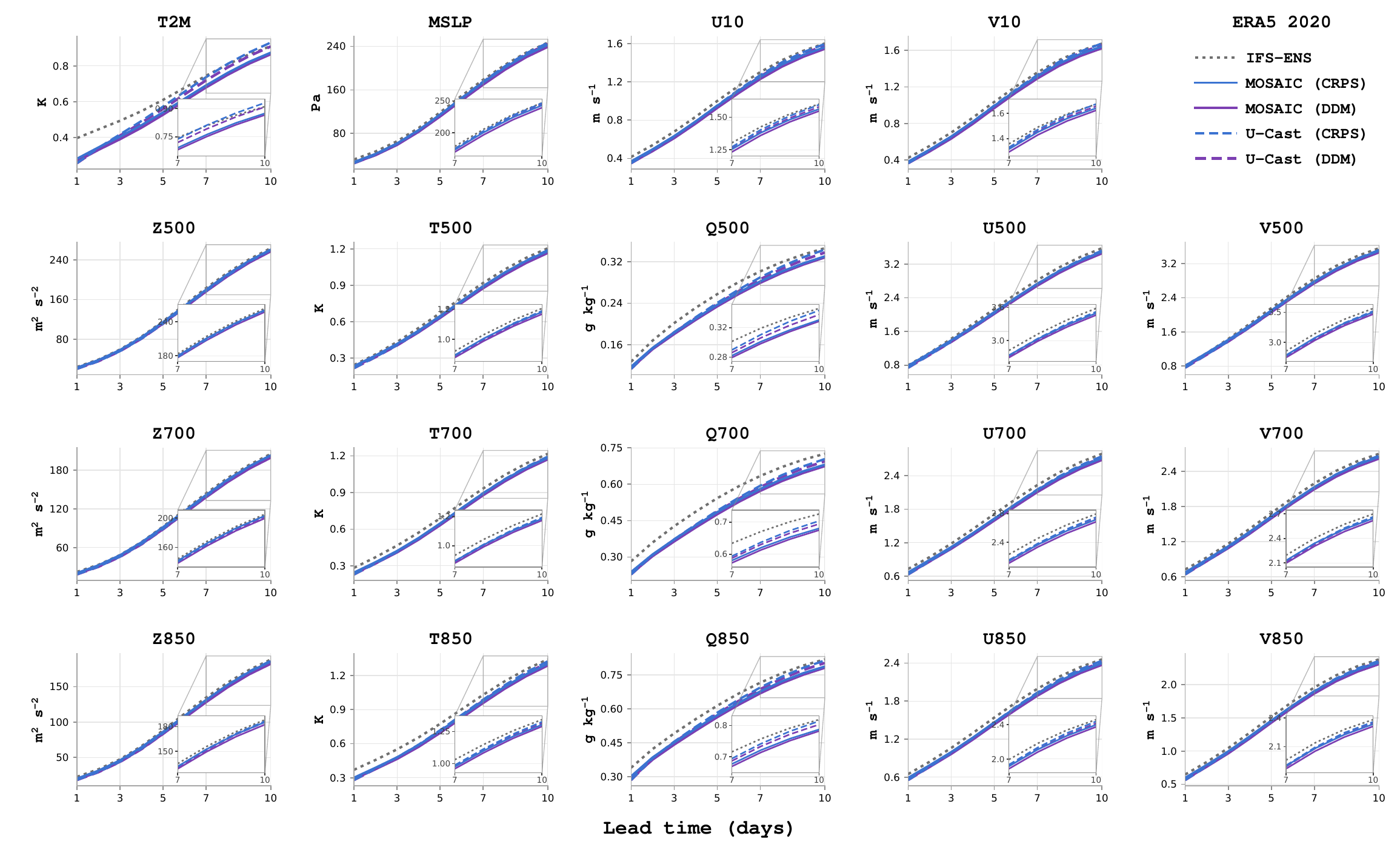}
    \caption{CRPS over 10-day autoregressive rollouts in ERA5 \texttt{2020}.
    }
    \label{fig:rollout_crps}
\end{figure*}

\begin{figure*}[t]
    \centering
    \includegraphics[width=\textwidth]{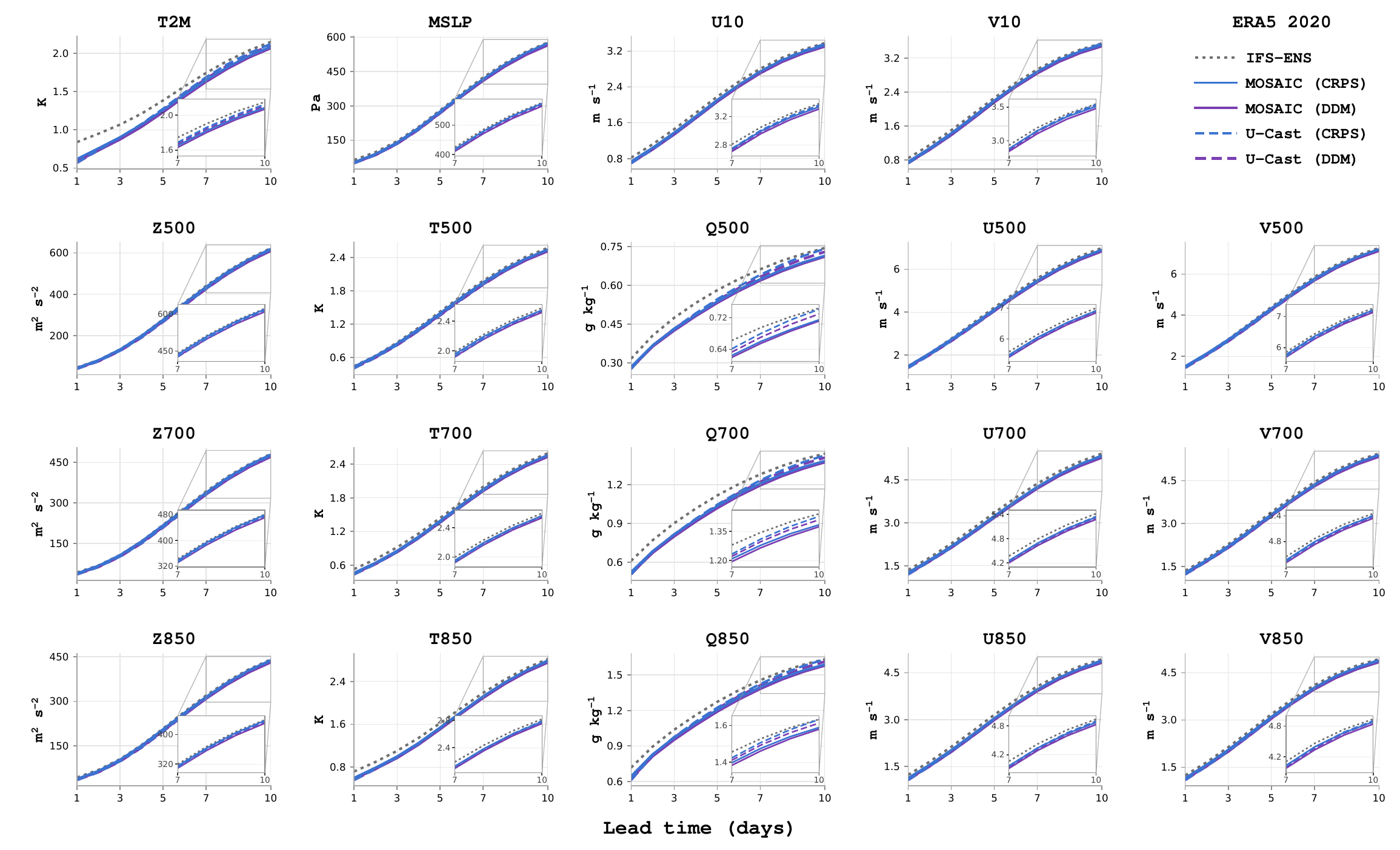}
    \caption{Ensemble-mean RMSE over 10-day autoregressive rollouts in ERA5 \texttt{2020}.
    }
    \label{fig:rollout_rmse}
\end{figure*}

\begin{figure*}[t]
    \centering
    \includegraphics[width=\textwidth]{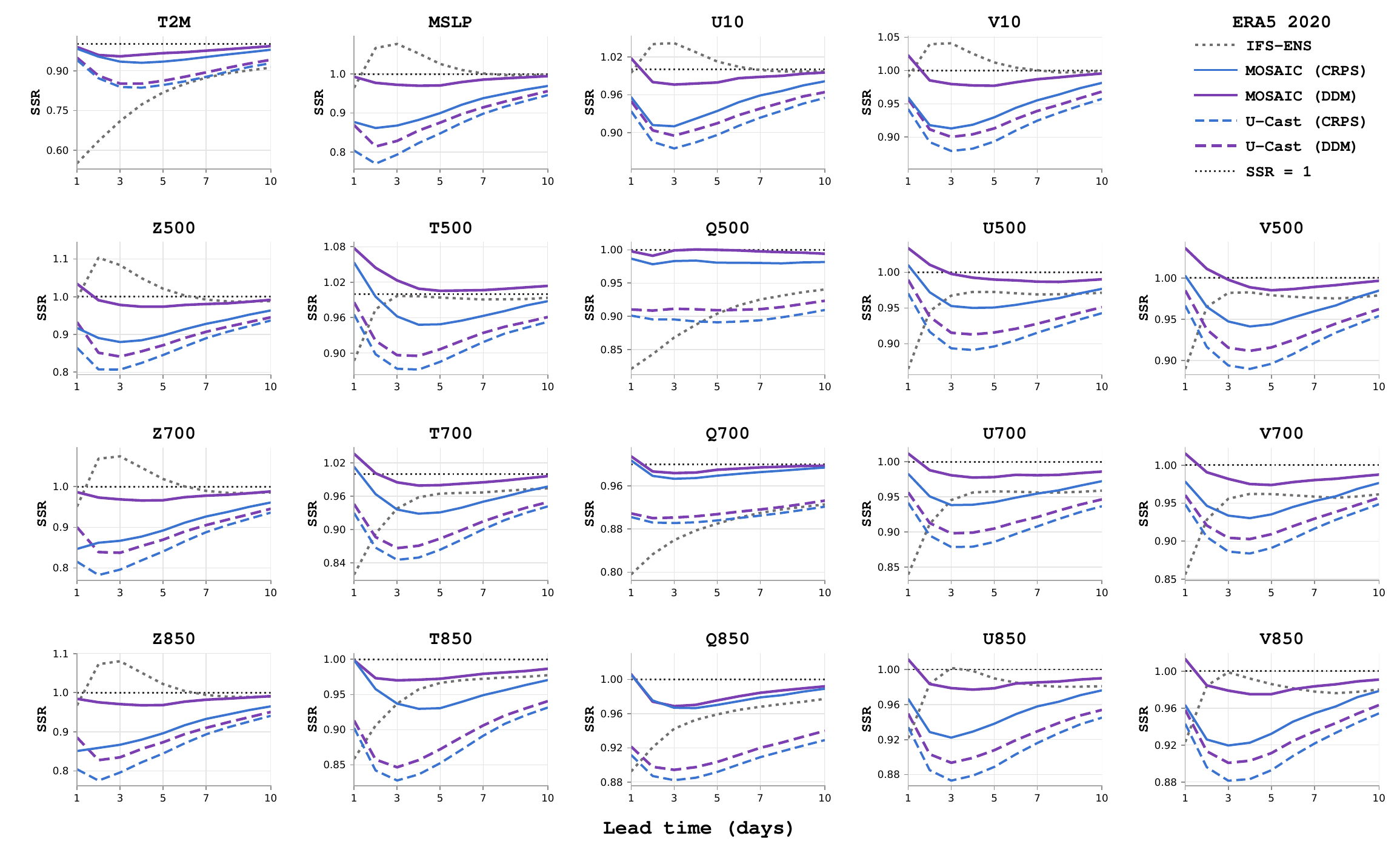}
    \caption{SSR over 10-day autoregressive rollouts in ERA5 \texttt{2020}. 
    }
    \label{fig:rollout_ssr}
\end{figure*}

We next visualize autoregressive rollout curves on 19 forecast channels common to our models and the publicly available WeatherBench~2 IFS-ENS archive~\citep{rasp2024weatherbench} (Figures~\ref{fig:rollout_crps}--\ref{fig:rollout_ssr}). For both backbones, DDM achieves lower CRPS than the corresponding CRPS-trained model at every evaluated channel--lead pair in this comparison, although several U-Cast differences fall below the precision displayed in~\Cref{tab:1p5deg_analysis}. Ensemble-mean RMSE improves across all pairs for MOSAIC and nearly all pairs for U-Cast. The SSR curves show that DDM generally moves the SSR closer to $1$.

\begin{figure*}[t]
    \centering
    \includegraphics[width=\textwidth]{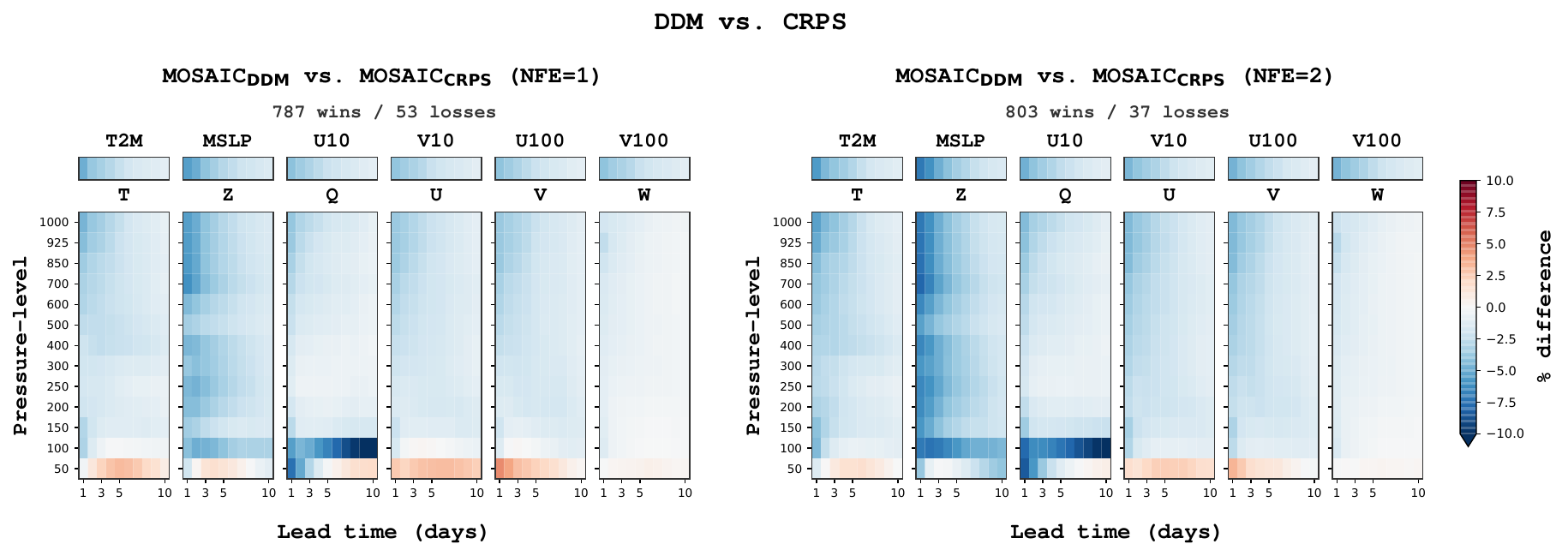}
    \caption{Effect of NFE on MOSAIC forecasting performance. Relative CRPS differences between DDM and CRPS-trained MOSAIC across 84 forecast channels over 10-day autoregressive rollouts. The same DDM checkpoint is evaluated with $\text{NFE}=1$ (left) and $\text{NFE}=2$ (right) in ERA5 \texttt{2020}.
    }
    \label{fig:mosaic_nfe}
\end{figure*}

We additionally evaluate MOSAIC with $\mathrm{NFE}=2$ under the same evaluation protocol. As shown in~\Cref{fig:mosaic_nfe}, two-step sampling retains lower CRPS in all 787 channel--lead pairs where one-step sampling outperforms CRPS-trained MOSAIC and achieves lower CRPS in 16 additional pairs, bringing the total to 803 out of 840. These results illustrate the potential benefits of refinement without retraining, at the cost of one additional model evaluation per member and rollout step.

\subsubsection{Robustness to Distribution Shift}
\label{sec:app:robustness}

In the distribution shift experiment, we initialize and verify forecasts with HRES-fc0. As in the ERA5 evaluation, we use all 732 initializations at 00 and 12 UTC in \texttt{2020}, with $M=50$ ensemble members and $\text{NFE}=1$ for DDM. Note that the SSR deviation reported in~\Cref{tab:hres_shift} refers to $|\mathrm{SSR}-1|$ averaged over all channels and lead times, with smaller values indicating better spread--skill balance.

\begin{figure*}[t]
    \centering
    \includegraphics[width=\textwidth]{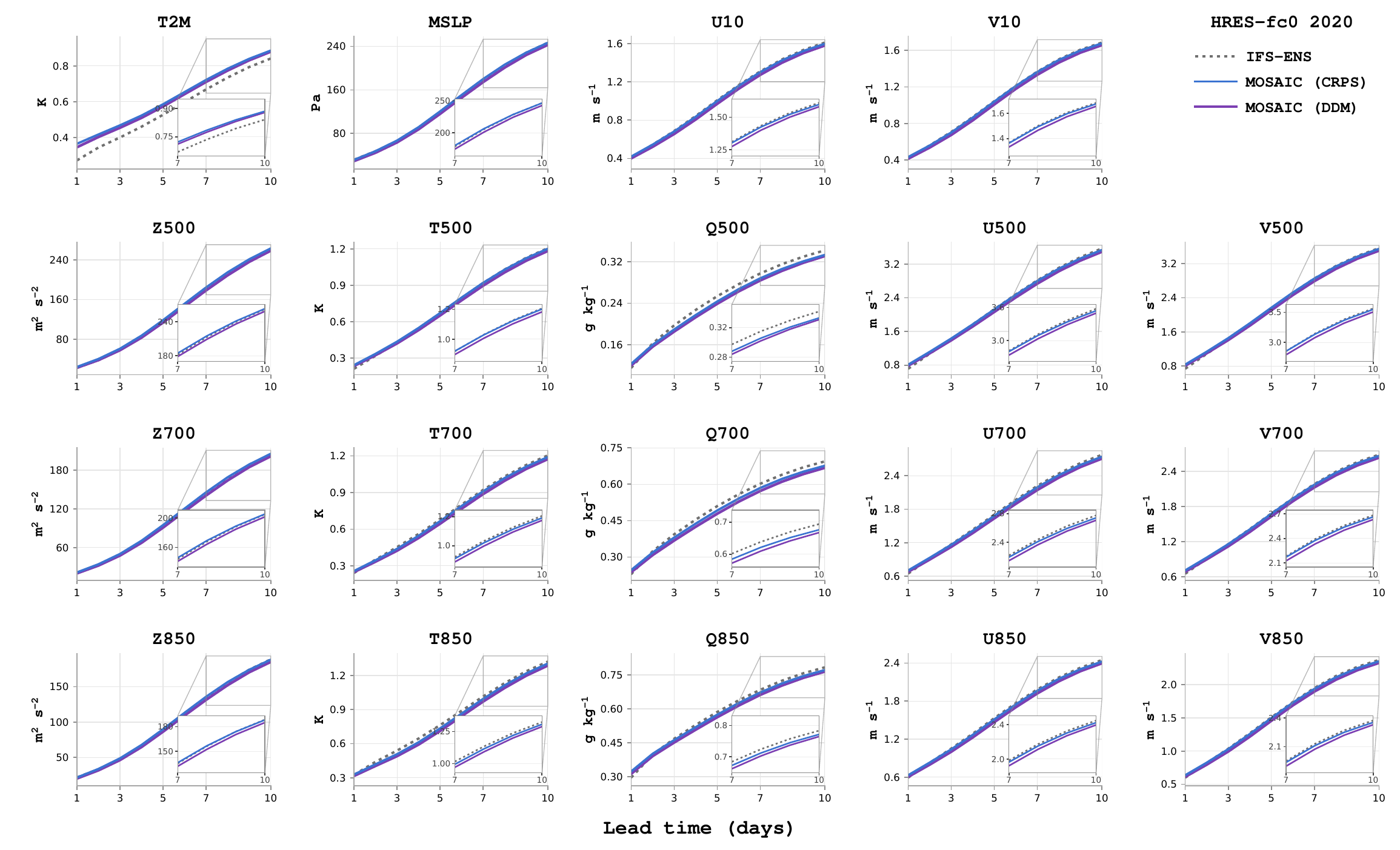}
    \caption{CRPS over 10-day autoregressive rollouts in HRES-fc0 \texttt{2020}.
    }
    \label{fig:rollout_crps_hres}
\end{figure*}

\begin{figure*}[t]
    \centering
    \includegraphics[width=\textwidth]{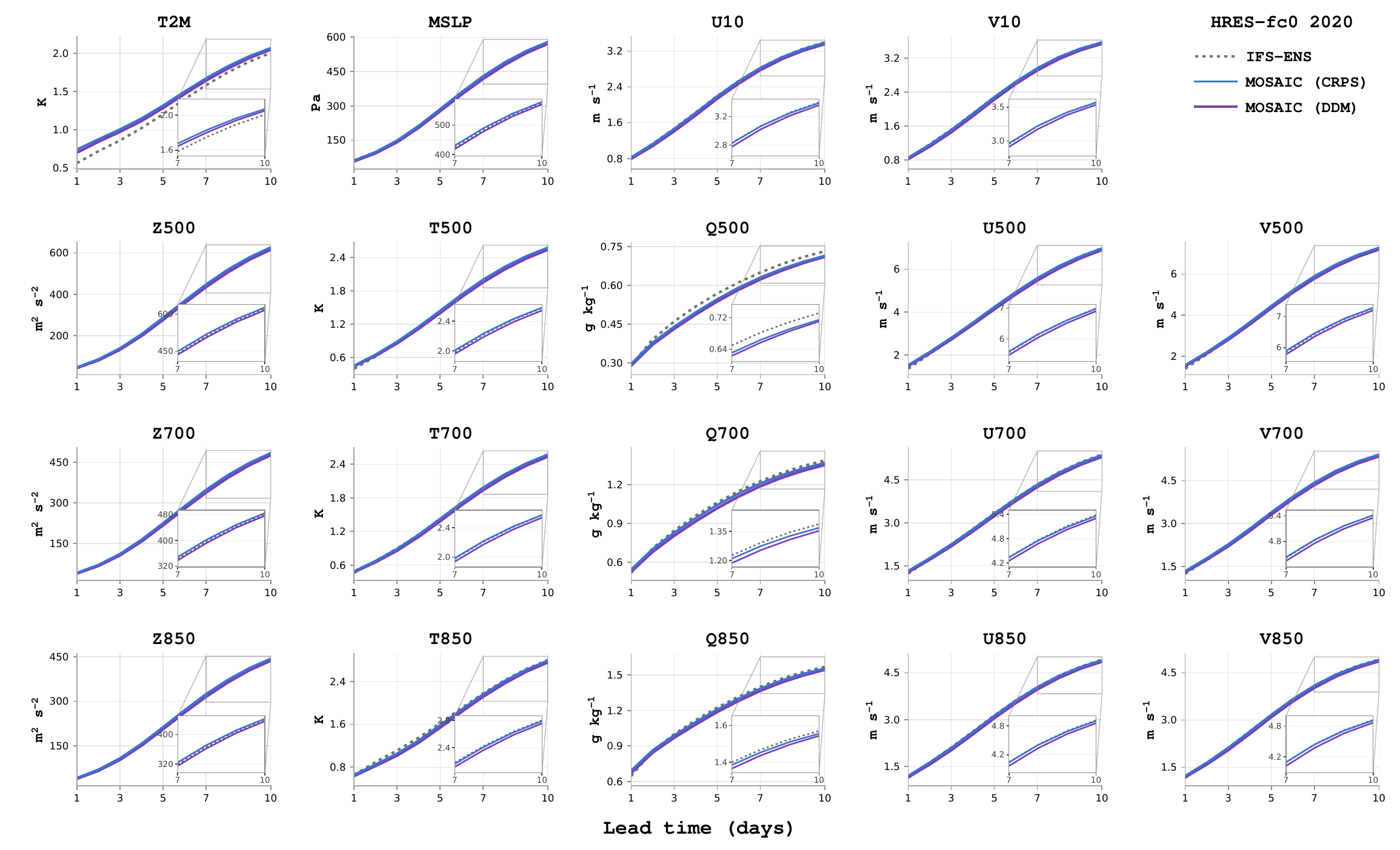}
    \caption{Ensemble-mean RMSE over 10-day autoregressive rollouts in HRES-fc0 \texttt{2020}.
    }
    \label{fig:rollout_rmse_hres}
\end{figure*}

\begin{figure*}[t]
    \centering
    \includegraphics[width=\textwidth]{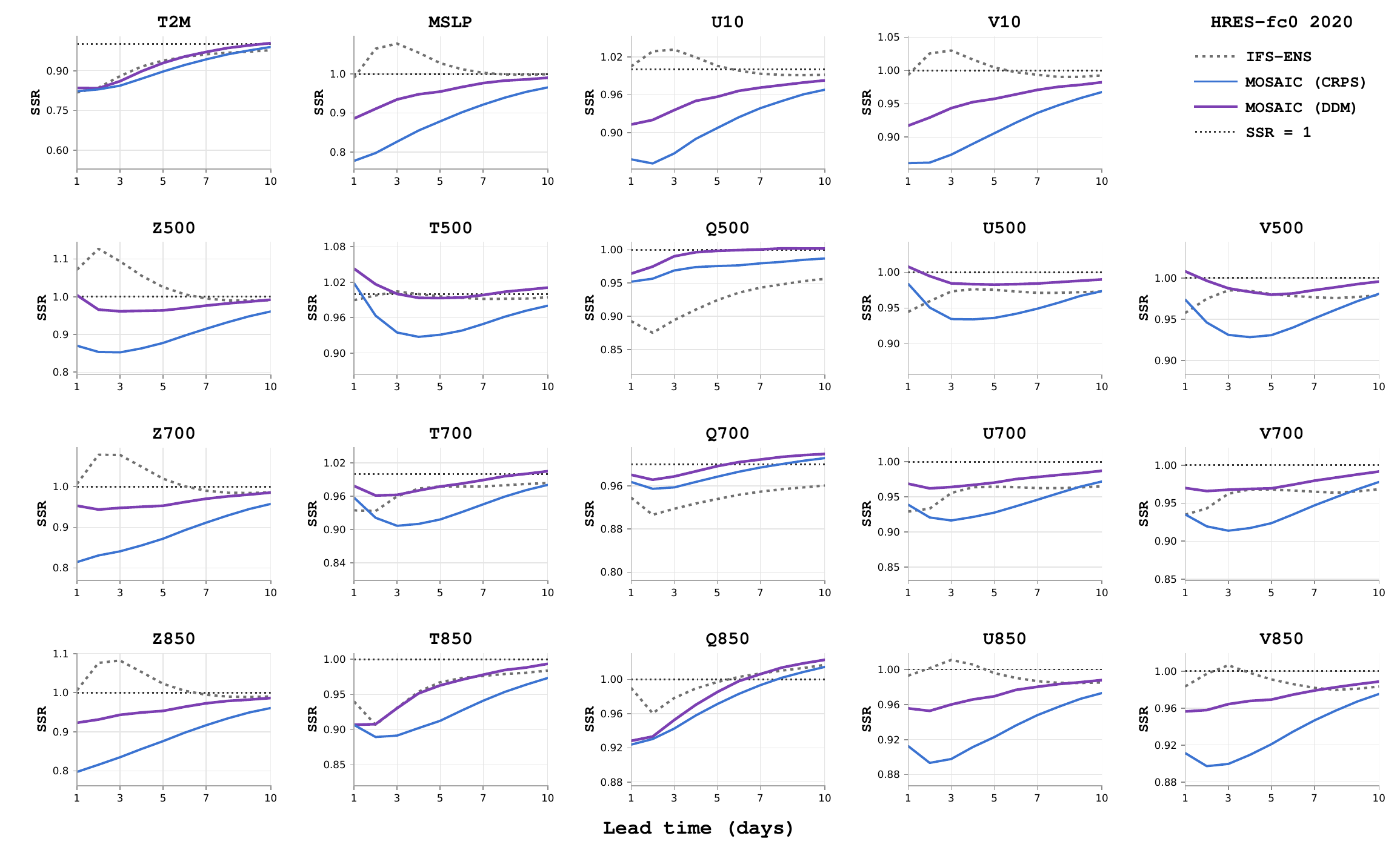}
    \caption{SSR over 10-day autoregressive rollouts in HRES-fc0 \texttt{2020}. 
    }
    \label{fig:rollout_ssr_hres}
\end{figure*}

To compare with the IFS-ENS baseline, Figures~\ref{fig:rollout_crps_hres}--\ref{fig:rollout_ssr_hres} show rollout curves for CRPS, ensemble-mean RMSE and SSR for the 19 channels shared with IFS-ENS. DDM achieves lower CRPS and ensemble-mean RMSE than the CRPS-trained model across all 190 channel--lead pairs and the spread--skill ratio is also generally closer to $1$, indicating better agreement between ensemble spread and forecast error. These results show that the advantages of DDM persist under the ERA5--HRES shift.

\subsubsection{Tropical Cyclone Tracking}
\label{sec:app:cyclone}

\begin{figure}[!t]
    \centering
    \includegraphics[width=0.9\linewidth]{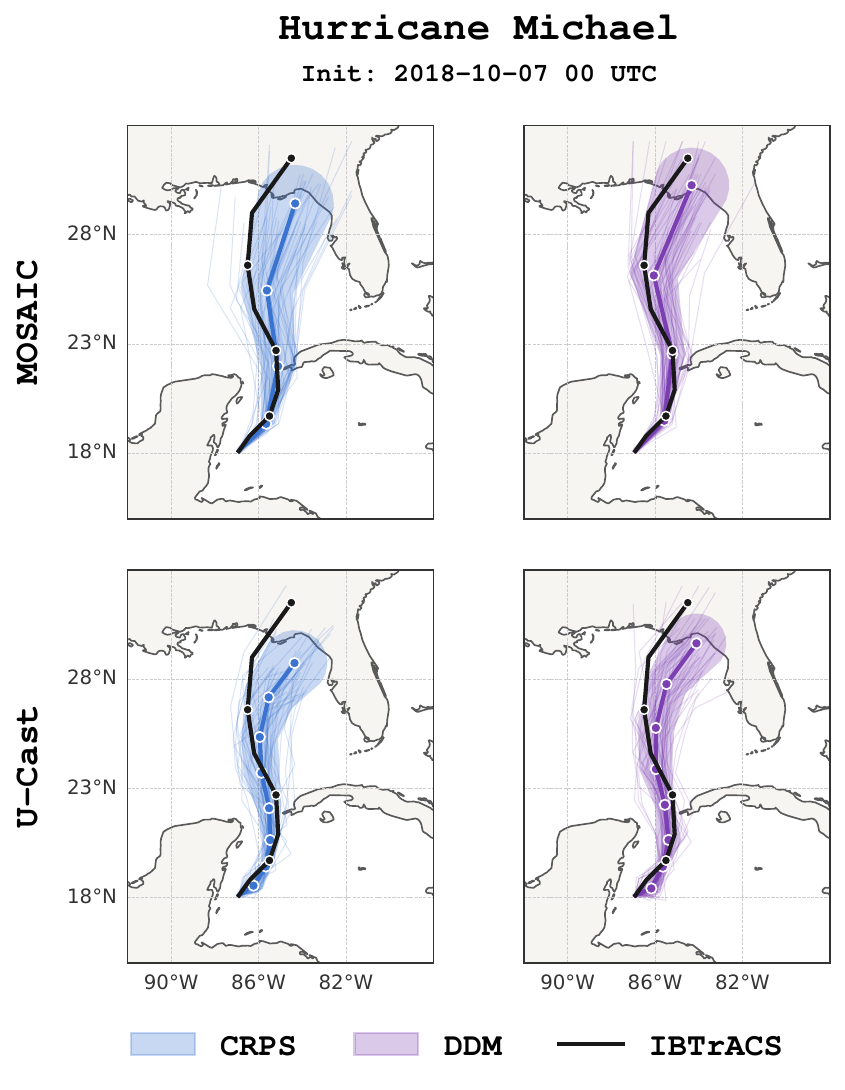}
    
    \caption{Hurricane Michael track forecasts from 50-member MOSAIC (top) and U-Cast (bottom) ensembles, comparing CRPS (blue, left) and DDM (purple, right). Michael occurred in 2018, within the validation period, and is included as an illustrative case study.
    }
    \label{fig:hurricane_michael}
    
\end{figure}

We provide additional tropical-cyclone tracking results for Hurricanes Ian and Michael using 50-member ensembles from MOSAIC and U-Cast, with IBTrACS best tracks as the reference. Figure~\ref{fig:hurricane_michael} shows the ensemble track forecasts for Hurricane Michael, and Table~\ref{tab:tc_tracks} reports ensemble-mean track error and track CRPS for both storms. DDM improves both metrics over CRPS training.

\begin{figure}[!t]
    \centering
    \includegraphics[width=0.9\linewidth]{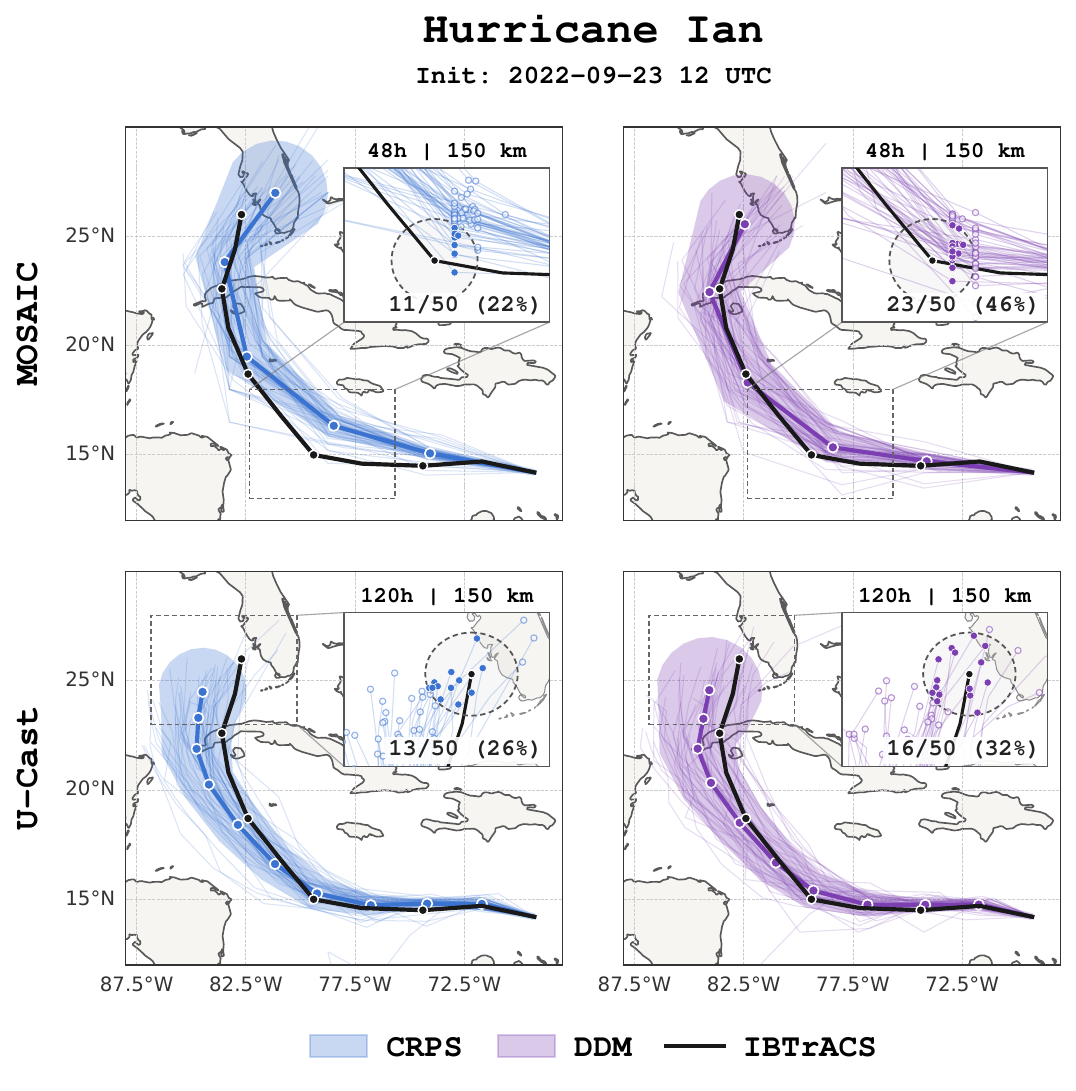}
    
    \caption{50-member ensemble track forecasts for Hurricane Ian initialized at 12 UTC on 23 September 2022, from MOSAIC (top) and U-Cast (bottom), comparing CRPS (blue, left) and DDM (purple, right). Insets show the 48\,h MOSAIC and 120\,h U-Cast forecast positions. Dashed circles mark a 150\,km radius around the time-matched IBTrACS center.
    }
    \label{fig:ian_inset_150}
    
\end{figure}

\begin{figure}[!t]
    \centering
    \includegraphics[width=0.9\linewidth]{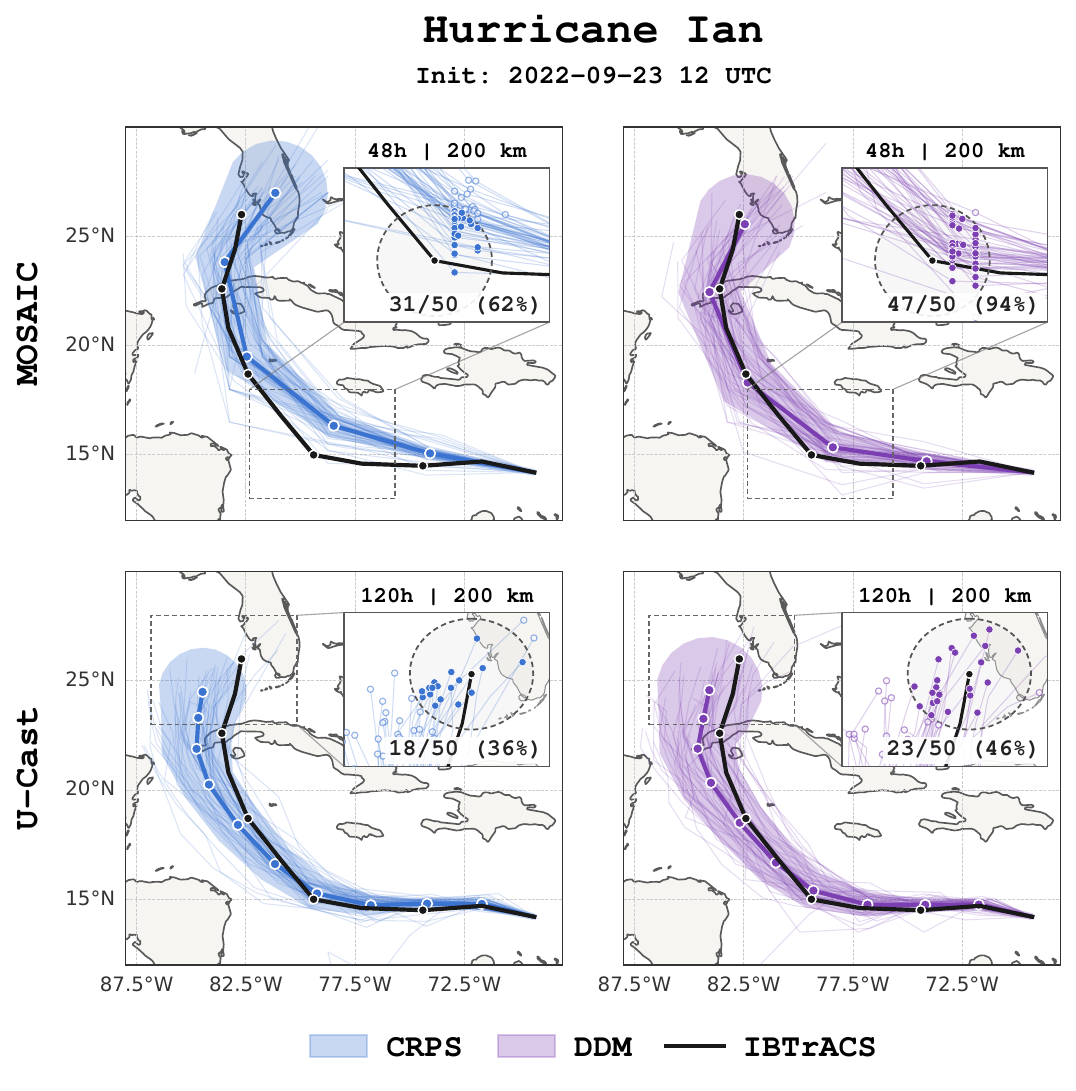}
    
    \caption{Same as Figure~\ref{fig:ian_inset_150}, but with a 200\,km radius.
    }
    \label{fig:ian_inset_200}
    
\end{figure}

\begin{figure}[!t]
    \centering
    \includegraphics[width=0.9\linewidth]{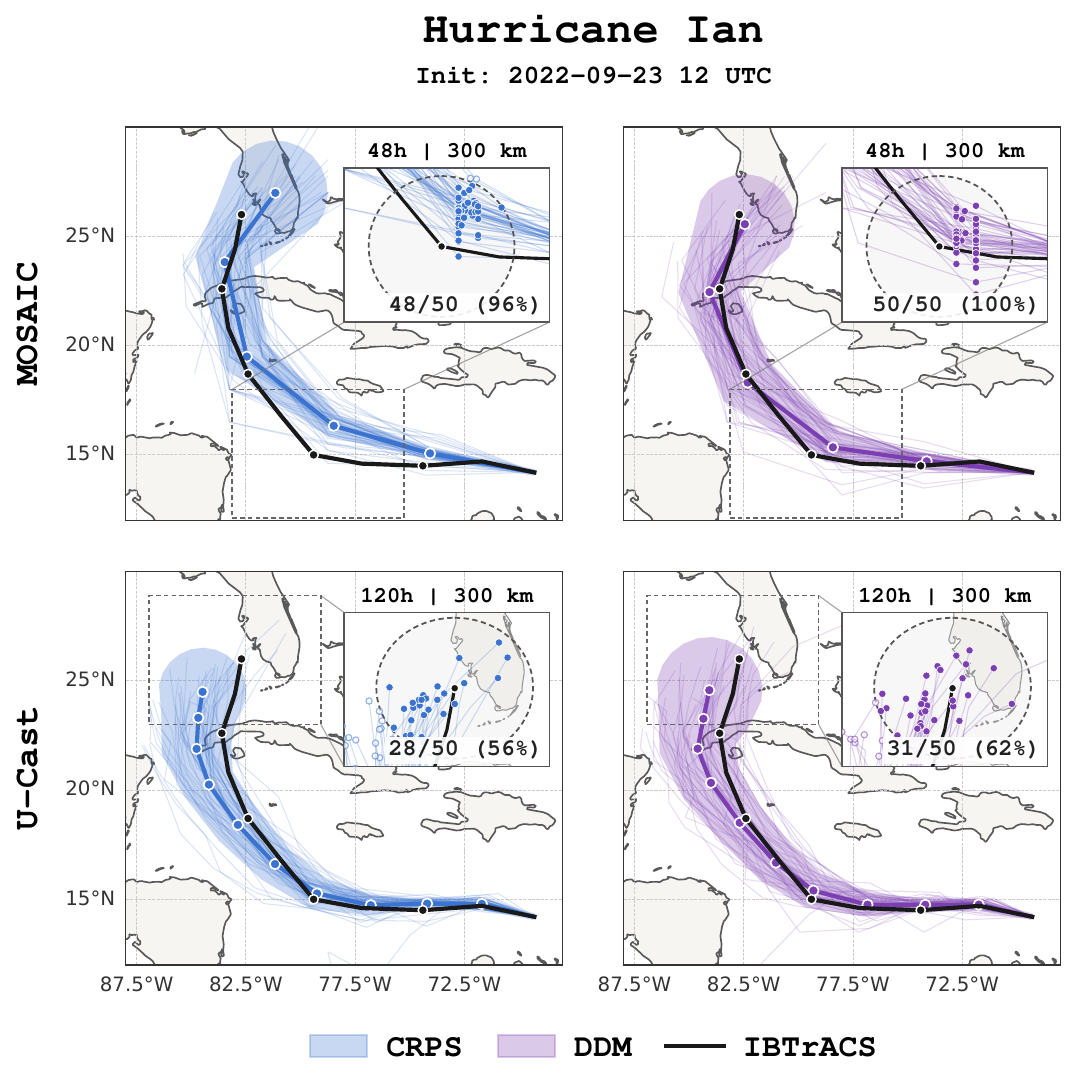}
    
    \caption{Same as Figure~\ref{fig:ian_inset_150}, but with a 300\,km radius.
    }
    \label{fig:ian_inset_300}
    
\end{figure}

To examine the ensemble forecast positions beyond the ensemble mean, Figures~\ref{fig:ian_inset_150}--\ref{fig:ian_inset_300} show enlarged views of Hurricane Ian forecasts with circles of radius 150, 200, and 300\,km around the time-matched IBTrACS center. The highlighted lead times are 48\,h for MOSAIC and 120\,h for U-Cast. Table~\ref{tab:ian_radius_counts} summarizes the number and percentage of members within each radius, using all 50 members regardless of whether individual positions fall inside the displayed crop. Across these radii, DDM places more members near the reference center than CRPS training for both backbones. At 200\,km, the fraction increases from $62\%$ to $94\%$ for MOSAIC and from $36\%$ to $46\%$ for U-Cast.

\paragraph{Tracking and Metrics}

We track storm centers using MSLP minima following~\citet{zhdanov2026sparse}. Each member track is initialized at the IBTrACS~\citep{knapp2010international} position and subsequently follows the minimum MSLP within a search radius around its previous center. Forecasts are evaluated against the time-matched IBTrACS best track during the tropical phase.

At each lead time, let $\{\mathbf{x}_m\}_{m=1}^M$ denote the predicted storm positions and $\mathbf{y}$ the observed position. We measure great-circle distance $d$ using the haversine formula with Earth radius $R=6371$\,km. The ensemble-mean track error is $d(\bar{\mathbf{x}},\mathbf{y})$, where $\bar{\mathbf{x}}$ is the arithmetic mean of the member latitude--longitude coordinates. Following TCBench~\citep{gomez2026tcbench}, we compute track CRPS as
\begin{equation}
    \mathrm{TrackCRPS}
    =
    \frac{1}{M}\sum_{m=1}^{M} d(\mathbf{x}_m,\mathbf{y})
    -
    \frac{1}{2M(M-1)}
    \sum_{i=1}^{M}
    \sum_{\substack{j=1\\j\neq i}}^{M}
    d(\mathbf{x}_i,\mathbf{x}_j).
\end{equation}

\begin{table*}[t]
    \centering
    \caption{
        Tropical-cyclone ensemble-mean track error and track CRPS (km).
    }
    \label{tab:tc_tracks}

    \resizebox{0.92\textwidth}{!}{%
    \renewcommand{\arraystretch}{1.2}
    \setlength{\tabcolsep}{6.0pt}

    \begin{tabular}{@{} l l  ccc ccc | cccc cccc @{}}
        \toprule

        &
        
        & 
        \multicolumn{6}{c}{
            {\texttt{\textbf{Hurricane Ian}}}
        }
        &
        \multicolumn{8}{c}{
            {\texttt{\textbf{Hurricane Michael}}}
        }
        \\

        \cmidrule(lr){3-8}
        \cmidrule(lr){9-16}

        &

        & 
        \multicolumn{3}{c}{
            {\textbf{\texttt{Track Error}} $\downarrow$}
        }
        &
        \multicolumn{3}{c}{
            {\textbf{\texttt{Track CRPS}} $\downarrow$}
        }
        &
        \multicolumn{4}{c}{
            {\textbf{\texttt{Track Error}} $\downarrow$}
        }
        &
        \multicolumn{4}{c}{
            {\textbf{\texttt{Track CRPS}} $\downarrow$}
        }
        \\

        \cmidrule(lr){3-5}
        \cmidrule(lr){6-8}
        \cmidrule(lr){9-12}
        \cmidrule(lr){13-16}

        Method

        & Objective

        & {\color{gray}\textbf{\texttt{1d}}}
        & {\color{gray}\textbf{\texttt{3d}}}
        & {\color{gray}\textbf{\texttt{5d}}}

        & {\color{gray}\textbf{\texttt{1d}}}
        & {\color{gray}\textbf{\texttt{3d}}}
        & {\color{gray}\textbf{\texttt{5d}}}

        & {\color{gray}\textbf{\texttt{1d}}}
        & {\color{gray}\textbf{\texttt{2d}}}
        & {\color{gray}\textbf{\texttt{3d}}}
        & {\color{gray}\textbf{\texttt{4d}}}

        & {\color{gray}\textbf{\texttt{1d}}}
        & {\color{gray}\textbf{\texttt{2d}}}
        & {\color{gray}\textbf{\texttt{3d}}}
        & {\color{gray}\textbf{\texttt{4d}}}
        \\

        \midrule

        \textbf{\texttt{MOSAIC}}
        & \textbf{\texttt{CRPS}}
        
        & 73
        & 89
        & 189

        & 51
        & 68
        & 125

        & 42
        & 81
        & 156
        & 231

        & 30
        & 54
        & 110
        & 153
        \\

        \textbf{\texttt{MOSAIC}}
        & \textbf{\texttt{DDM}}
        & \cellbg 38
        & \cellbg 44
        & \cellbg 55

        & \cellbg 26
        & \cellbg 39
        & \cellbg 63

        & \cellbg 25
        & \cellbg 17
        & \cellbg 68
        & \cellbg 138

        & \cellbg 17
        & \cellbg 17
        & \cellbg 47
        & \cellbg 86
        \\

        \midrule

        \textbf{\texttt{U-Cast}}
        & \textbf{\texttt{CRPS}}
        & 39
        & 59
        & 245

        & 30
        & 43
        & 148

        & 36
        & 75
        & 150
        & 308

        & 25
        & 51
        & 112
        & 231
        \\

        \textbf{\texttt{U-Cast}}
        & \textbf{\texttt{DDM}}
        & \cellbg 35
        & \cellbg 37
        & \cellbg 210

        & \cellbg 24
        & \cellbg 38
        & \cellbg 123

        & \cellbg 35
        & \cellbg 62
        & \cellbg 107
        & \cellbg 210

        & \cellbg 24
        & \cellbg 40
        & \cellbg 72
        & \cellbg 141
        \\

        \bottomrule
    \end{tabular}%
    }
\end{table*}

\begin{table}[t]
\centering
\caption{Number and percentage of ensemble members within 150, 200, and 300\,km of the time-matched IBTrACS center for Hurricane Ian. MOSAIC is evaluated at 48\,h and U-Cast at 120\,h, corresponding to the inset lead times as shown in Figures \ref{fig:ian_inset_150}--\ref{fig:ian_inset_300}.}
\label{tab:ian_radius_counts}
\renewcommand{\arraystretch}{1.2}
\setlength{\tabcolsep}{6pt}
\begin{tabular}{@{}llcccc@{}}
\toprule
Method & Objective & Lead time & 150\,km & 200\,km & 300\,km \\
\midrule
\textbf{\texttt{MOSAIC}} & \textbf{\texttt{CRPS}} & 48\,h & 11/50 (22\%) & 31/50 (62\%) & 48/50 (96\%) \\
\textbf{\texttt{MOSAIC}} & \textbf{\texttt{DDM}} & 48\,h & \cellbg 23/50 (46\%) & \cellbg 47/50 (94\%) & \cellbg 50/50 (100\%) \\
\midrule
\textbf{\texttt{U-Cast}} & \textbf{\texttt{CRPS}} & 120\,h & 13/50 (26\%) & 18/50 (36\%) & 28/50 (56\%) \\
\textbf{\texttt{U-Cast}} & \textbf{\texttt{DDM}} & 120\,h & \cellbg 16/50 (32\%) & \cellbg 23/50 (46\%) & \cellbg 31/50 (62\%) \\
\bottomrule
\end{tabular}%
\end{table}

\end{document}